%% file: DEID_manuscript.tex
\PassOptionsToPackage{unicode}{hyperref}
\PassOptionsToPackage{hyphens}{url}
\documentclass[
]{article}
\usepackage{lmodern}
\usepackage{amssymb,amsmath}
\usepackage{iftex}
\ifPDFTeX
\usepackage[T1]{fontenc}
\usepackage[utf8]{inputenc}
\usepackage{textcomp} % provide euro and other symbols
\else % if luatex or xetex
\usepackage{unicode-math}
\defaultfontfeatures{Scale=MatchLowercase}
\defaultfontfeatures[\rmfamily]{Ligatures=TeX,Scale=1}
\fi
\IfFileExists{upquote.sty}{\usepackage{upquote}}{}
\IfFileExists{microtype.sty}{% use microtype if available
  \usepackage[]{microtype}
  \UseMicrotypeSet[protrusion]{basicmath} % disable protrusion for tt fonts
}{}
\makeatletter
\@ifundefined{KOMAClassName}{% if non-KOMA class
  \IfFileExists{parskip.sty}{%
    \usepackage{parskip}
  }{% else
    \setlength{\parindent}{0pt}
  \setlength{\parskip}{6pt plus 2pt minus 1pt}}
}{% if KOMA class
\KOMAoptions{parskip=half}}
\makeatother
\usepackage{xcolor}
\IfFileExists{xurl.sty}{\usepackage{xurl}}{} % add URL line breaks if available
\IfFileExists{bookmark.sty}{\usepackage{bookmark}}{\usepackage{hyperref}}
\newcommand{\manuscripttitle}{MedDeID enables locally governed clinical-text de-identification from real or synthetic training data}
\hypersetup{
  hidelinks,
pdftitle={\manuscripttitle},
pdfcreator={LaTeX via pandoc}}
\usepackage{longtable,booktabs}
\usepackage{etoolbox}
\makeatletter
\patchcmd\longtable{\par}{\if@noskipsec\mbox{}\fi\par}{}{}
\makeatother
\IfFileExists{footnotehyper.sty}{\usepackage{footnotehyper}}{\usepackage{footnote}}
\makesavenoteenv{longtable}
\usepackage{graphicx}
\makeatletter
\def\maxwidth{\ifdim\Gin@nat@width>\linewidth\linewidth\else\Gin@nat@width\fi}
\def\maxheight{\ifdim\Gin@nat@height>\textheight\textheight\else\Gin@nat@height\fi}
\makeatother
\setkeys{Gin}{width=\maxwidth,height=\maxheight,keepaspectratio}
\makeatletter
\def\fps@figure{htbp}
\makeatother
\providecommand{\tightlist}{%
\setlength{\itemsep}{0pt}\setlength{\parskip}{0pt}}
\usepackage{array}
\usepackage{caption}
\usepackage[super,comma,sort&compress]{natbib}
\input{supplementary_identifiers.tex}

\graphicspath{{figures/main/}}

\author{}
\date{}

\begin{document}

\hypertarget{title}{%
	\section{\texorpdfstring{\raggedright\manuscripttitle}{\manuscripttitle}}\label{title}}

\textbf{Authors:} Stig Hellemans\textsuperscript{1,3,*}, Tom
Stroobants\textsuperscript{2,3}, Elyne Scheurwegs\textsuperscript{3},
Pieter Meysman\textsuperscript{1}, Philippe G. Jorens\textsuperscript{2,3}, Kris
Laukens\textsuperscript{1}

\textbf{Affiliations}

\begin{enumerate}
	\tightlist
	\item Adrem Data Lab, Department of Computer Science, University of Antwerp, Antwerp, Belgium
	\item Laboratory of Experimental Medicine and Pediatrics (LEMP), University of Antwerp, Antwerp, Belgium
	\item Antwerp University Hospital (UZA), Edegem, Belgium
\end{enumerate}

\textsuperscript{*}Correspondence: stig.hellemans@uantwerpen.be

\clearpage

\hypertarget{abstract}{%
	\subsection{Abstract}\label{abstract}}

Clinical notes contain personally identifiable information (PII), restricting reuse for research
and medical AI, especially when data cannot leave an institution. We developed MedDeID, an
on-premises framework combining in-house annotation and synthetic-note generation with model
training, inference, pseudonymisation and evaluation. On an independently annotated, adjudicated
300-note Dutch hospital benchmark, a hospital-trained compact transformer detected 98.9\% of
identifying text while redacting 0.24\% of text outside annotated identifiers; a synthetic-only
counterpart detected 96.1\%. On 100 primary-care notes, the synthetic-trained model achieved higher
recall than the hospital-trained model (90.3\% versus 87.0\%) and greater robustness to
identifier-format perturbations. An English instantiation trained without real text detected 99.7\%
and 98.9\% of annotated identifier characters on two external synthetic benchmarks. These results
demonstrate transfer of the workflow to another language, but not clinical English performance.
MedDeID provides a route to locally governed de-identification using real or synthetic training
data.

\begin{center}\rule{0.5\linewidth}{0.5pt}
\end{center}

\hypertarget{introduction}{%
	\subsection{Introduction}\label{introduction}}

The unstructured free text of the electronic health record --- admission documentation, discharge
letters, progress notes, consultation reports --- is the location where clinicians record the
reasoning, history and nuance that structured fields cannot capture. It is therefore among the most
valuable substrates for clinical research and for developing and evaluating medical artificial
intelligence \citep{kreimeyer2017nlp}. Moreover, it is saturated with personally identifiable
information (PII). Under the EU General Data Protection Regulation, processing health data for
scientific research requires a lawful basis, an applicable Article 9 condition and appropriate
technical and organisational safeguards. Article 89 specifically requires safeguards for research,
including data minimisation and, where the purposes can be fulfilled, pseudonymisation
\citep{gdpr2016, art29wp2014}. In the United States, the HIPAA Privacy Rule provides Safe Harbor
and Expert Determination routes for de-identification \citep{hhs2012deid}. De-identification is
therefore a key safeguard for secondary uses of clinical free text. The task extends beyond
removing obvious identifiers: both direct identifiers, such as a patient's name, and indirect (or
quasi-) identifiers, such as age, postal code or admission date, may reveal an individual's
identity and therefore fall within scope \citep{art29wp2014, kayaalp2018modes}.

English-language systems have reached both benchmark maturity and production scale. The
i2b2/UTHealth shared tasks established a shared vocabulary for protected health information (PHI)
\citep{stubbs2015overview, stubbs2015annotating}; neural sequence models pushed entity-level F1
into the high nineties \citep{dernoncourt2017deid, johnson2020deid, murugadoss2021ensemble}; and
Philter, an openly released rule-based system \citep{norgeot2020philter}, was subsequently
certified by external audit and scaled at the University of California, San Francisco, where it has
placed more than 130 million de-identified notes from 2.75 million patients in the hands of over
600 researchers without further ethics review \citep{radhakrishnan2023certified}. That audit found
no residual identifiers in any of the seventeen non-date HIPAA categories and estimated that fewer
than 0.025\% of patients remained at risk of re-identification through shifted dates
\citep{radhakrishnan2023certified}. Philter is the reference example of what a deployable, openly
available de-identification system makes possible \citep{norgeot2020philter}.

The situation for Dutch and Flemish clinical text, and by extension for most non-English settings,
is very different, and the deficit is not merely one of effort. De-identification is inherently
language-specific: identifiers are expressed through language-dependent patterns, including local
lexicons, name morphology, date and address conventions, and the shorthand style of clinical
documentation, all of which limit the transferability of methods developed for other languages. The
principal rule-based Dutch tool, DEDUCE, was carefully hand-built for a single institution and
achieves high recall there \citep{menger2018deduce}, but rule systems of this kind generalise
poorly: Trienes and colleagues showed that an existing Dutch rule-based method failed on new data,
whereas a state-of-the-art neural architecture generalised across institutions and domains with far
less configuration effort \citep{trienes2020comparing}, and a 2025 replication study confirmed the
variability of these tools across settings \citep{seinen2025replication}. Dutch clinical language
models --- MedRoBERTa.nl, pre-trained on real hospital notes, and the general-purpose RobBERT
family --- provide practical encoder backbones for compact Dutch de-identifiers
\citep{verkijk2021medroberta, delobelle2020robbert, delobelle2024robbert2023}. In parallel, a wave
of work applies generative large language models to de-identification with impressive recall in
favourable conditions \citep{wiest2025anonymizer, altalla2025gpt, doremus2025moderate,
	chang2024zeroshot, singh2024indian, aghakasiri2025survey}; however, the LLM evaluated here was
substantially slower than the compact local model and remained brittle to the structured formatting
of real notes.

Two further gaps compound the first. Public real-data benchmarks for Dutch clinical
de-identification remain unavailable, so groups evaluate on inaccessible local corpora and
independent reproduction is difficult \citep{libbi2021synthetic}. Synthetic pre-training,
diversity-aware data construction and teacher--student distillation are emerging as routes around
the same bottleneck in English and French \citep{yazdani2026conormdeid, posada2026shield}, but
their benefit depends on training regime and evaluation domain. Existing systems also often stop at
detection, ignoring what healthcare institutions need from pseudonymisation: dates must be shifted
so that clinical chronology survives, and ages must be reduced to the coarsest granularity that
still carries clinical meaning. Collapsing a two-month-old infant to ``0 years old'', for example,
destroys information a paediatrician needs and creates an avoidable trade-off between
re-identification risk and clinical value.

This study addresses three challenges across languages and clinical settings: poor transfer of
de-identification systems, restrictions on sharing clinical text for model development and
evaluation, and loss of useful information during pseudonymisation. MedDeID responds with a locally
deployable workflow that supports training on local annotations or synthetic notes, evaluation on
shareable synthetic benchmarks, and clinically informed date and age pseudonymisation. To compare
real-data and synthetic-only training on authentic Dutch clinical text, we trained two independent
models with the same compact architecture: one on annotated hospital notes and the other
exclusively on synthetic Dutch notes, and evaluated both on held-out hospital and primary-care
notes. We then reuse MedDeID to train a separate English model exclusively on synthetic English
data, testing whether researchers and healthcare institutions can adapt the framework to another
language rather than rebuild the entire workflow. This does not involve transfer from the Dutch
models or establish performance on real English clinical text. Across both language instantiations,
MedDeID uses a common schema, character offsets, post-processing, pseudonymisation rules, and
evaluation contract. Core PII recall measures the percentage of genuinely identifying characters
detected, while non-PII redaction measures the clinical text removed outside annotated spans.

\hypertarget{results}{%
	\subsection{Results}\label{results}}

\hypertarget{benchmark}{%
	\subsubsection{Evaluation datasets and clinical benchmarks}\label{benchmark}}

Table 1 summarises six data resources divided into nine non-overlapping training and evaluation
partitions. Clinical performance was assessed on two real Dutch benchmarks: a held-out,
independently annotated and adjudicated hospital set (n=300) and a separately annotated
primary-care set (n=100). The Dutch synthetic benchmark provides a shareable test of the complete
workflow, whereas the three entirely synthetic English benchmarks test whether the workflow can be
reproduced in another language; they do not establish performance on real English clinical text
\citep{temlm2026technetium,weatherhead2026asqphi}. No evaluation data were used for model training.
The hospital and primary-care gold standards contained 4,269 and 4,010 PII annotations,
respectively; annotation volume, density and mapped category composition across all six benchmarks
are compared in Supplementary Fig. \SuppFigDatasetSummary. Full data, annotation and evaluation
procedures are reported in Methods and Supplementary Sections S1--S3.

\hypertarget{parity}{%
	\subsubsection{Performance on the Dutch hospital benchmark}\label{parity}}

We trained \texttt{meddeid-dutch-uza} on a separate corpus of 4,470 annotated development notes
from Antwerp University Hospital (UZA). On the 300-note hospital benchmark, core PII recall was
98.9\% (98.5--99.3\%; Table 2). Non-PII redaction was 0.24\% (0.16--0.32\%). The model uses a
RobBERT-2023 Dutch encoder \citep{delobelle2024robbert2023} with dual heads for token-level span
boundaries and span labels, followed by MedDeID's post-processing layer (Fig. 1). Annotator 1
reached 98.8\% recall at 0.17\% non-PII redaction and Annotator 2 98.5\% at 0.36\%. The model's
recall was therefore 0.1--0.4 percentage points higher, while its non-PII redaction rate lay
between the two annotators. These values place the model within the range observed for the
annotators on this benchmark. Because the annotators helped create the adjudicated reference
standard, this comparison does not show that the model is better than, or equivalent to,
physicians.

Every external comparator was substantially further back. Our upgraded Belgian DEDUCE reached
88.0\% recall, the previously published Dutch de-identification model \texttt{deidentify}, 86.4\%
\citep{trienes2020comparing}, and a locally deployed large language model (Qwen3-8B)
\citep{yang2025qwen3} 84.2\%. A zero-shot NER model for personally identifiable information (PII),
GLiNER-PII, reached 76.6\% \citep{zaratiana2024gliner}, the original Dutch DEDUCE 75.9\%
\citep{menger2018deduce}, and two general-purpose neural PII detectors reached 72.1\% (OpenAI) and
49.1\% (OpenMed). These differences varied by identifier category and by the components within
annotated identifier spans (Supplementary Table \SuppTabLabelRecall{} and Figs \SuppFigLabelRange).

Higher recall did not require excessive non-PII redaction. Using the complete MedDeID pipeline,
\texttt{meddeid-dutch-uza} redacted 0.236\% of characters outside gold spans, below every other
evaluated model and between the two annotators (0.174--0.359\%); \texttt{deidentify}, for
comparison, redacted 0.558\% (Table 2). Of the 1,236 non-PII characters redacted by
\texttt{meddeid-dutch-uza}, 353 came from entirely incorrect detections. The remaining 883 were
extra characters at the edges of detected PII. Thus, most over-redaction reflected imprecise span
boundaries rather than spurious detections (Supplementary Table \SuppTabDutchHospital).

The MedDeID pipeline can also use patient and caregiver names stored in note metadata. This
increased the hospital-trained model's recall only marginally (98.8\% to 98.9\%) but raised Belgian
DEDUCE's recall from 78.5\% to 88.0\%. Because metadata availability may differ across
institutions, results with and without this step are reported in Supplementary Table
\SuppTabDutchHospital; all main-text results use the complete MedDeID pipeline \emph{with} metadata
included.

\hypertarget{synthetic}{%
	\subsubsection{Synthetic training and cross-setting performance}\label{synthetic}}

We trained \texttt{meddeid-dutch-synth} entirely on synthetic text, without using real patient data
(Table 1; Methods). On the real hospital benchmark, it reached 96.1\% core PII recall --- 2.8
percentage points below the hospital-trained model (2.2--3.5), but above every external comparator
we tested (Table 2). For a site that cannot assemble a labelled corpus, this provides an open
starting point for local deployment and fine-tuning.

We validated all systems on 100 general-practice notes from a Belgian primary-care practice,
annotated and sub-annotated by one physician-researcher against the same guideline. Every system
achieved lower core PII recall on the primary-care benchmark than on the hospital benchmark (Table
2). On these notes, the synthetic-trained model had higher recall than the hospital-trained model
(90.3\% versus 87.0\%), a difference of 3.3 percentage points (1.7--5.2). The largest
hospital-to-primary-care difference among the neural systems was observed for the local LLM (84.2\%
versus 60.8\%); the original DEDUCE reached 75.9\% and 57.2\%, respectively.

Among the identifier spans detected by the models, the hospital-trained model assigned the correct
identifier category in 98.9\% of cases on hospital notes and 89.8\% on primary-care notes. The
corresponding values for the synthetic-trained model were 92.6\% and 79.3\% (Supplementary Table
\SuppTabLabelFidelity{} and Fig. \SuppFigLabelConfusion).

Thus, after detecting a span, the synthetic-trained model was less reliable than the
hospital-trained model at assigning the correct identifier category in both care settings.

Non-PII redaction differed even more sharply on primary-care notes. The OpenAI neural PII detector
redacted 11.46\% of all characters outside gold spans and the OpenMed detector 8.25\%, against
0.29\% for the hospital-trained model and 1.22\% for \texttt{meddeid-dutch-synth} (Table 2).
Supplementary Fig. \SuppFigNonPIIRedaction{} shows how these redacted characters were distributed
across the PII categories predicted by each system. Such rates indicate substantial loss of
research-relevant text, a failure mode that recall-only comparisons do not capture.

On the openly released synthetic benchmark, \texttt{meddeid-dutch-synth} reached 99.8\% core PII
recall (99.5--100.0\%) with 0.28\% non-PII redaction (0.21--0.35\%). The hospital-trained model
reached 96.6\% recall on this set, 2.3 points below its hospital result. The published
\texttt{deidentify} baseline reached 76.7\% core PII recall with 0.37\% non-PII redaction, whereas
Qwen3-8B reached 90.9\% recall with 1.28\% non-PII redaction. The benchmark therefore provides a
public, reproducible test of the complete workflow, but its in-domain scores are neither evidence
of model superiority nor estimates of performance on real clinical text.

\hypertarget{english-portability}{%
	\subsubsection{Performance on English synthetic benchmarks}\label{english-portability}}

To test whether MedDeID's workflow was not only specific to Dutch, we instantiated it with separate
British- and American-English profiles. This produced a distinct English model, trained exclusively
on synthetic English text; it was not obtained by transferring the Dutch model. We evaluated it on
two independently sourced external synthetic benchmarks (Table 3): the Technetium-I test split,
containing 74,700 multi-section clinical notes and 1,161,437 annotations, and ASQ-PHI (Adversarial
Synthetic Queries for Protected Health Information), containing 1,051 queries and 2,973 annotations
\citep{temlm2026technetium,weatherhead2026asqphi}.

On Technetium-I, MedDeID achieved 99.730\% annotation-character recall (99.725--99.735\%), the
highest among the seven systems. GLiNER, the next-highest comparator, achieved 97.773\%
(97.754--97.791\%); the difference was 1.957 percentage points (1.940--1.975). MedDeID's non-PII
redaction rate was 1.606\% (1.605--1.608\%).

On ASQ-PHI, MedDeID achieved 98.9\% annotation-character recall (98.6--99.2\%), the highest among
the seven evaluated systems. GLiNER, the next-highest recall comparator, achieved 96.3\%
(95.8--96.8\%); the difference was 2.58 percentage points (2.06--3.12). Its raw non-PII redaction
was 6.21\% (5.99--6.43\%). ASQ-PHI leaves explicit ages below 90 unannotated, so detecting such ages
increases the reported non-PII redaction rate. We therefore recalculated this rate for every system
after excluding the same age characters, regardless of which label a system assigned. After excluding these ages,
MedDeID's rate was 0.89\% (0.75--1.03\%), second to OBI RoBERTa i2b2 at 0.55\% (0.45--0.66\%);
the other comparator estimates ranged from 1.18\% to 4.29\% (Supplementary Table~\SuppTabASQ{}).

The two benchmarks provide complementary tests: Technetium-I covers structured clinical-note
formats at scale, whereas ASQ-PHI focuses on adversarial queries and hard negatives. Consistently
high recall across both supports the feasibility of training a competitive English de-identifier
entirely from synthetic data. Because both benchmarks are themselves synthetic, however, these
results do not establish performance on authentic English clinical notes or clinical safety in an
English-language health system. Nor do they test cross-language transfer, because the English model
was trained separately. Confidence intervals for the external benchmarks in Table 3 and the
separate held-out MedDeID English benchmark are reported in Supplementary Tables
\SuppTabEnglishRange.

\hypertarget{robustness}{%
	\subsubsection{Robustness to formatting and value perturbations}\label{robustness}}

To assess how sensitive the models were to identifier values and formatting, we selected 100 notes
from each of the three test sets (300 notes in total). We then changed one feature at a time ---
the names used, name capitalisation, name format, date format, date value or age format --- and
measured the resulting change in recall relative to the unmodified notes (Fig. 2). Supplementary
Figs \SuppFigStabilityRange{} and Tables \SuppTabStabilityDetailRange{} provide the detailed
results.

Changing which names appeared did not significantly reduce recall after correction, suggesting that
neither model depended strongly on the particular names in the test notes. Changes to formatting
and date values had larger effects, especially for the hospital-trained model. Its largest recall
loss was 19.2 points when patient-name formatting was altered on the synthetic benchmark, whereas
the synthetic-trained model's largest loss in any setting was 4.6 points under primary-care
caregiver-name capitalisation (Supplementary Fig. \SuppFigStabilityCells{} and Table
\SuppTabStabilityCells).

Overall, the synthetic-trained model was more stable across the three test sets: the mean positive
recall loss, with improvements counted as zero, was 1.25 points, compared with 2.96 points for the
hospital-trained model, and fewer perturbations produced statistically significant losses (5 versus
7; Fig. 2 and Supplementary Table \SuppTabStabilityAggregate). This pattern appeared in each test
set, although the synthetic-trained model was not better under every individual perturbation.
Sensitivity to date values remains practically important because date distributions vary across
settings and evolve over time; a deployable de-identifier should remain reliable beyond the date
range represented in its training data.

\hypertarget{pseudonymisation}{%
	\subsubsection{Pseudonymisation validity and retained clinical detail}\label{pseudonymisation}}

Detection is only half of what a healthcare institution needs. After the detection and
post-processing stages shown in Fig. 1, MedDeID applies a separate substitution step designed to
retain clinical meaning. Dates are first normalised: a model that detects the day and month of a
date but misses the year leaves a residual identifier, so partially recognised dates are healed to
their full extent using the regular formats in which dates appear. Exact dates belonging to one
patient are then shifted by a single patient-specific offset, preserving every interval between
events while breaking the absolute timeline. Date-like expressions with a resolution coarser than
one day, such as ``May 2023'' or ``summer'', are represented as the complete interval of calendar
dates compatible with the text; both interval bounds are shifted, and the output is widened only
when the shifted interval crosses a boundary at the source resolution. For illustration, a
$+10$-day shift --- chosen for clarity, not as a recommended production offset --- changes ``May
2023'' (1--31 May) to ``May/June 2023'' because the shifted interval (11 May--10 June) spans two
months. This avoids imputing an arbitrary day, false precision and the resulting directional bias.
How the offset is used also affects privacy. Operational safeguards and the residual
re-identification risk of shifted dates are discussed below and specified in Methods.

Age and birth-date reduction is likewise age-dependent: whole years suffice for older patients,
whereas progressively finer units are retained for younger children and neonates. The retained
level of detail for each age group is reported in Supplementary Table \SuppTabAgeGranularity. This
avoids discarding clinically relevant paediatric age information.

We evaluated MedDeID's pseudonymisation layer separately from PII detection by applying it to every
gold date and age/birthdate span with a fixed 371-day shift and checking whether each output
satisfied the intended transformation. Gold-span transformation-layer failure rates were 0.0\%
(0/1,693) on synthetic text, 0.8\% (14/1,798) on hospital text and 3.0\% (36/1,219) in primary
care. When model-predicted spans were used instead, end-to-end failure rates were 1.5\%, 3.4\% and
6.0\%, respectively; the rates of gold spans with unredacted characters were 0.7\%, 2.2\% and 2.4\%
(Supplementary Table \SuppTabPseudonymValidity{} and Fig. \SuppFigPseudonymFailureModes). Most
gold-span transformation failures reflected unsupported or invalid source formats, identifying a
remaining portability gap in the pseudonymisation layer rather than the detector.

\hypertarget{compute}{%
	\subsubsection{Inference time on GPU and CPU hardware}\label{compute}}

The practical case for a compact model is sharpest on compute. In the measured timing runs,
de-identifying the 300-note hospital set took the local large language model 11,411.5\,s
(3\,h\,10\,min), against 18.4\,s for the hospital-trained model on the same NVIDIA T4 --- a
621-fold wall-time difference (Fig. 3). More consequential for local deployment in healthcare
institutions, the hospital-trained model completed the same set in 320.3\,s (5.3\,min) on CPU
alone, roughly 36 times faster than the GPU-based LLM run, and the CPU-only Belgian DEDUCE finished
in 17.9\,s. The CPU result shows that smaller datasets or routine local batches can be processed
without dedicated GPU infrastructure; for millions of archived notes, we recommend GPU acceleration
(Fig. 3). The larger general-purpose neural PII detectors were slower than our compact transformer
and achieved substantially lower recall across all three evaluation datasets.

\hypertarget{framework}{%
	\subsubsection{Deployment and local governance framework}\label{framework}}

The empirical comparisons establish two complementary starting points: the hospital-trained model
defines the attainable in-domain level of performance, while the synthetic-trained model can be
deployed or adapted without access to our clinical training data. MedDeID connects these routes in
one reproducible framework (Fig. 4). A shared core contract keeps the taxonomy, JSONL document
schema, character offsets and post-processing consistent across synthetic-data generation, human
annotation, model training, inference and evaluation. Healthcare institutions can therefore begin
with the open synthetic-trained model, generate additional synthetic examples, or annotate local
notes and fit a site-specific model without exporting the source text or resulting weights.

The same contract also links prediction data to a curated and sub-annotated benchmark, allowing
evaluation to guide further annotation, data generation and model refinement. Users who only want
to de-identify text need the local \texttt{meddeid} package and a compatible model. The other
MedDeID components allow healthcare institutions to build, evaluate and refine their own models
reproducibly. Belgian DEDUCE provides a directly usable CPU baseline and Belgian lookup
foundations. Together, MedDeID and Belgian DEDUCE give healthcare institutions practical tools to
deploy and adapt a de-identifier locally, without sending clinical notes or locally trained model
weights outside the institution.

\hypertarget{discussion}{%
	\subsection{Discussion}\label{discussion}}

This study contributes both an empirical result and a reusable framework for acting on it. The
central empirical finding is that synthetic-only training provides a reproducible route to a
compact Dutch de-identifier evaluated on real hospital and primary-care text. The hospital-trained
model reached the range observed for the two physician annotators on the hospital benchmark, while
its non-PII redaction rate lay between theirs. On the notes from the primary-care practice, the
synthetic-trained model had higher recall than the hospital-trained model (90.3\% versus 87.0\%;
3.3 percentage points). Both model variants exceeded the other data-driven and rule-based
approaches tested and ran locally on commodity hardware. MedDeID translates these findings into a
common workflow for annotation or synthetic-data generation, training, local inference and
benchmarking. Because these stages share versioned interfaces, language profiles, datasets and
models can be replaced without rebuilding the full workflow, supporting reuse and audit across
locally governed deployments. Together, these features make Dutch clinical-text de-identification
more practical for routine use across healthcare institutions, as Philter has demonstrated in
English-language settings \citep{norgeot2020philter, radhakrishnan2023certified}.

Our findings also temper enthusiasm for LLM-based de-identification \citep{wiest2025anonymizer,
	altalla2025gpt, doremus2025moderate, chang2024zeroshot, singh2024indian, aghakasiri2025survey}.
Large models can reach high recall in favourable conditions --- the \textit{LLM-Anonymizer}
reported 99.24\% with a 70-billion-parameter model on German letters \citep{wiest2025anonymizer}
--- but their size creates practical barriers. Using a cloud service requires clinical text to
leave local infrastructure, whereas running the model locally demands more hardware and was far
slower than the compact transformer in our tests. The locally deployed Qwen3-8B model also reached
84.2\% recall on hospital notes and 60.8\% on primary-care notes. Runtime will vary with the
deployment setup, and more efficient software could narrow the measured speed gap. It would not,
however, remove the additional hardware and governance requirements or the lower recall observed
here. These constraints may be particularly important for smaller healthcare institutions.

The result that most changes how we think about this problem is that a model trained on nothing but
synthetic text had higher recall on the primary-care notes than a model trained on thousands of
real hospital notes. Real single-institution data teaches a model the conventions of that
institution alongside the task; synthetic data generated with deliberate formatting diversity
teaches the task with fewer of those conventions attached. The perturbation experiments locate
important mechanisms: the hospital-trained model was more sensitive to name formatting and shifted
date values across evaluation scopes. This supports synthetic generation as a complementary source
of variation rather than simply a second-best substitute for real data \citep{libbi2021synthetic}.
Recent English and French studies similarly use synthetic pre-training or diversity-aware
distillation to improve robustness and local deployment \citep{yazdani2026conormdeid,
	posada2026shield}; our contribution is to compare a synthetic-only and a real-data-trained model on
the same adjudicated Dutch hospital benchmark and then show that their ranking reversed on
authentic primary-care text. This cross-setting recall advantage did not extend to label
assignment: among detected spans, the synthetic-trained model was less accurate than the
hospital-trained model in both settings. The effect of these label errors on real-world utility
remains uncertain and is likely to depend on the misclassified category and downstream task. Some
errors may not change which text is removed, whereas errors involving dates or birth dates can
alter temporal transformations during pseudonymisation and may affect subsequent data modelling.
Aggregate label accuracy alone therefore cannot establish the practical safety or utility of
pseudonymised outputs; this requires task-specific evaluation. We did not evaluate mixed training;
future work should separately test richer synthetic data for label fidelity and mixed
real--synthetic training for additional robustness.

By extending MedDeID with English-language profiles, we show that the workflow is not tied to Dutch
and can be adapted to another language. The resulting English model achieved the highest
annotation-character recall among seven systems on both Technetium-I and ASQ-PHI. When the same
unannotated explicit ages were excluded for every system, its ASQ-PHI non-PII redaction rate was
0.89\% (0.75--1.03\%), second to OBI RoBERTa i2b2 at 0.55\% (0.45--0.66\%). Much of MedDeID's
raw rate therefore reflected ASQ-PHI's treatment of explicit ages, although the same taxonomy
mismatch also affected several age-aware comparators. This demonstrates language portability rather than
clinical validity: the model was trained and tested only on synthetic English data. Its performance
on real English clinical notes, across institutions and patient populations, therefore remains
unknown.

A further distinctive element of this work is its attention to clinical utility. De-identification
that destroys medical signal is self-defeating, so MedDeID's post-processing layer preserves
temporal relationships through a per-patient date shift and preserves clinically meaningful age
information through a transformation that is coarse for adults but graded for children, where age
precision carries diagnostic weight. The sub-span annotation layer, to our knowledge not previously
applied to Dutch clinical text, separates the identifying information itself from surrounding
formatting and clinically relevant context, and should make future cross-method comparisons fairer
\citep{pilan2022tab}. Core PII recall should nevertheless be interpreted as a measure of how much
identifying information is detected, not as a stand-alone estimate of re-identification probability
\citep{ford2025reidentification}.

Preserving chronology does not make shifted dates anonymous. Weekday and holiday schedules can
narrow candidate offsets or reveal shifted calendars \citep{evans2023dateshifting}. In the
4,770-note hospital corpus used for model development and evaluation, 4,358 notes (91.4\%) were
created from Monday to Friday, illustrating the strong weekday signature available to an attacker.
We therefore recommend patient- or stay-specific offsets exceeding one year in either direction
where feasible. When weekday consistency is desired, users may choose signed multiples of seven,
but preserving weekdays is a utility choice rather than a privacy guarantee. Date shifting remains
pseudonymisation, and dense longitudinal records may require additional safeguards
\citep{radhakrishnan2023certified,alexander2022datereidentification}.

Several constraints bound these claims. The hospital benchmark comprises 300 notes from a single
hospital. The 100-note primary-care set and the 300-note Dutch synthetic benchmark were each
annotated and sub-annotated by a single physician. Some identifiers may therefore have been missed,
and decisions about where an identifier begins and ends may reflect that physician's judgement.
Differences in performance between datasets may consequently reflect differences in annotation, not
only differences between settings. The comparison with hospital annotators is descriptive and is
measured against a gold standard that the annotators themselves constructed, not against
independent ground truth. The core sub-annotation layer reduces the advantage conferred by our own
span-boundary conventions but cannot eliminate the more fundamental dependence on our guideline and
adjudication decisions.

The Dutch synthetic benchmark and all three English benchmarks contain no real patient text. They
are valuable for reproducibility, failure analysis and workflow portability, but performance on
these benchmarks cannot show how accurately the models will perform on authentic clinical text. The
English model has not been evaluated on authentic English clinical notes.

The Qwen3-8B comparison used deliberately light prompt engineering. The other external comparators
--- \texttt{deidentify}, GLiNER-PII, the original Dutch DEDUCE, OpenAI Privacy Filter and OpenMed
--- were evaluated using their existing implementations without dataset-specific tuning. These
choices reflect realistic deployment in healthcare institutions but may not show the best
performance obtainable from these external systems. A larger frontier generative model with
extensive prompting might achieve higher recall, though it would not remove the speed, cost or
governance barriers. Qwen3-8B reached 84.2\% recall on hospital notes and 60.8\% on primary-care
notes, a difference of 23.4 points.

Although 100 notes were selected from each test set, each perturbation analysis included only notes
containing the relevant type of identifier (19--98 notes; Supplementary Table
\SuppTabStabilitySamples). The results should therefore be interpreted as showing which changes
tended to cause more failures, rather than as precise estimates of the size of these effects.
Replacing names addresses only whether model performance changes when different names are used; it
does not establish whether either model memorised examples from its training data.

Although the MedDeID components were used to construct and evaluate the systems in this study, the
MedDeID framework has not yet been prospectively adopted and evaluated across multiple independent
healthcare institutions; its portability as a complete workflow therefore remains to be
established.

Accurate and locally deployable de-identification of Dutch clinical free text is achievable with
compact models. The synthetic-only route offers a reproducible starting point where real training
data cannot be shared, while the hospital-trained model defines the attainable in-domain level of
performance. To maximise performance, we recommend validating and tuning both the model and
post-processing layer for the target text domain. By connecting both routes in MedDeID --- an open
workflow covering annotation, synthetic-data construction, training, local inference,
post-processing and evaluation through a shared schema (Fig. 4) --- we provide healthcare
institutions with a testable pathway for constructing, validating and refining locally governed
systems. The English instantiation shows that this pathway can be extended to another language;
clinical validation must still be performed in every language and health system where it will be
used.

\hypertarget{methods}{%
	\subsection{Methods}\label{methods}}

\textbf{Data sources, ethics and governance.} A total of 4,770 clinical notes
were assembled from the free-text electronic health record of Antwerp University
Hospital (UZA), a tertiary university hospital: 4,470 were assigned to model
development and 300 were reserved as a held-out hospital benchmark before model
training. Its composition is reported in Supplementary Table \SuppTabDatasetDepartments. A further 100
notes from a Belgian general practice were used for \mbox{external validation}. The Ethics Committee of Antwerp
University Hospital and the University of Antwerp approved the study on
2 June 2025 (project 7654; CTMS/EDGE 004299); its chair approved an amendment
covering the primary-care validation on 4 June 2026. The Ethics Committee waived the requirement
for individual informed consent for this retrospective use of routinely collected clinical
records. The documented legal bases were GDPR Arts. 6(1)(f) and 9(2)(j), with Art. 89 safeguards
\citep{gdpr2016}. UZA was
the controller and the University of Antwerp the processor. Identifiable
data remained in UZA's secured Azure research environment, with
restricted MFA and SSH-key access and local-only model inference. No
identifiable data were exchanged with the participating general practice; only aggregated validation
results were shared.

No formal sample-size calculation was performed. The hospital development-corpus size was
pragmatic: annotation stopped after validation performance and the marginal benefit observed during
iterative annotation had plateaued, rather than at a prespecified statistical target.
Evaluation-cohort sizes were constrained by the availability and governance of manually reviewed
clinical text and by the effort required for character-level sub-annotations. The hospital sample
was selected before training to provide broad department coverage and independent dual annotation;
the primary-care sample was intended as a first cross-setting validation rather than a definitive
estimate of primary-care performance.

\textbf{Annotation and gold-standard construction.} The
development and training corpus was annotated by a single
physician-researcher (SH). To improve throughput and consistency, candidate
spans were pre-suggested by a model trained on previously annotated data
and normalised by a rule-based layer that standardised span boundaries;
all suggestions were human-reviewed. The separately reserved hospital benchmark was
annotated independently by two physician-researchers (SH and TS), both trained in
advance on the guideline, and adjudicated into a single gold standard by
sequential review and consensus resolution of disagreements. The
Dutch and English synthetic benchmarks and the primary-care set were each
annotated by one physician-researcher (SH) against the same guideline.

\textbf{Annotation guideline and label scheme.} The guideline was
adapted from the NIH/NLM Scrubber annotation guidelines, the HIPAA
18-identifier scheme and the GraSCCo/GeMTeX framework for German clinical
text \citep{kayaalp2016guidelines,kayaalp2014challenges,hhs2012deid,
	lohr2024grascco,lohr2024grascco_phi}, with further rules and examples
developed from the clinical text encountered in this study. Span labels
cover person names; addresses and locations; healthcare and other
organisations; dates; ages and birth dates; professions; contact details;
numeric or alphanumeric identifiers; and exceptional identifying content.
Name and address labels distinguish patient, caregiver and other referents;
identifier labels distinguish patients from caregivers; and organisation
labels distinguish healthcare from other organisations. The full guideline
is provided in Supplementary Appendix A and in the versioned archival release described under
Data availability.

\textbf{Sub-annotation and core PII recall.} Gold annotations were
subdivided to separate identifying information from incidental material captured inside the same
span. For example, in the fictitious phone number \texttt{+32~(0)\underline{493}~\underline{12}~\underline{34}~\underline{57}}, spaces,
punctuation and the country prefix are recorded separately from the person-specific digits (underlined).
Characters falling into the
sub-annotation categories \emph{formatting}, \emph{additional
	information}, \emph{medical information}, \emph{title} and \emph{time}
were excluded from the core PII set, because they either identify no one on
their own or carry clinical meaning. Definitions of all sub-annotation categories and their
inclusion in core PII are provided in Supplementary Table \SuppTabSubannotations. Let $G_{\mathrm{core}}$ be the set
of all remaining gold-character positions and $R_s$ the union of
character positions redacted by system $s$. We define
\begin{equation}
	\operatorname{Recall}_{\mathrm{core\,PII}}(s)=\frac{|G_{\mathrm{core}}\cap R_s|}{|G_{\mathrm{core}}|}.
\end{equation}
This label-agnostic measure gives credit whenever the relevant content is
removed, irrespective of a system's chosen span boundary or output label.
Let $G_{\mathrm{span}}$ be the union of complete gold spans and $D$ all
evaluated document-character positions. The complementary utility measure is
\begin{equation}
	\operatorname{RedactionRate}_{\mathrm{non\text{-}PII}}(s)=\frac{|R_s\setminus G_{\mathrm{span}}|}{|D\setminus G_{\mathrm{span}}|},
\end{equation}
the non-PII redaction rate. It is decomposed in the Supplementary into
characters belonging to false-positive spans, which do not overlap any
annotated PII span, and PII boundary extensions, the excess characters redacted
beyond the boundary of an overlapping annotated PII span. Characters
excluded from the core PII set but lying inside a gold span contribute to neither
measure; their coverage remains visible through overall recall.

\textbf{Synthetic data generation.} We developed the Dutch and English synthetic
datasets between May and August 2026 using the same language-profile-driven
workflow. Each structured case combined a clinical scenario derived from Synthea
\citep{walonoski2018synthea} with synthetic PII sampled from regional resources:
the Belgian Dutch (\texttt{nl-BE}) profile of \texttt{meddeid-language-nl}, or the
British and American English (\texttt{en-GB} and \texttt{en-US}) profiles of
\texttt{meddeid-language-en} (Fig. 4). The Dutch resources originated in the lookup lists
used by Belgian DEDUCE. The only difference between the two LLM-generation
workflows was the authoring model: GPT-5.4 mini for Dutch and GPT-5.6 Luna for
English. Both models converted compact case descriptions and designated PII
fields into clinical notes and enclosed the PII in explicit markers, which we
removed locally while recording exact character offsets. Deterministic, targeted
renderers supplemented the LLM-generated notes in both languages. No real patient
text or PII was provided to either generator.

We prioritised de-identification challenges over narrative realism by including difficult non-PII
examples, such as laboratory results, medication names, eponyms and device identifiers, and by
varying the clinical-note format \citep{libbi2021synthetic}. During development, we used OpenAI
Codex to inspect individual outputs and corpus-level quality reports and to refine the prompts,
generators, validation rules and post-processing. Automated checks and manual review identified and
corrected remaining errors in annotation boundaries and coverage. The held-out benchmarks were
manually reviewed and sub-annotated against the same guideline used for the clinical benchmarks.

\textbf{Model architecture and training.} All three MedDeID models used the same
dual-head RoBERTa architecture. One head performs three-way BIO span detection
over every token; the other assigns one of 14 entity categories from the first
token of each detected span. Byte-level tokens were processed in overlapping
512-token windows with 64-token overlap, and logits were averaged across
overlaps before typed spans were reconstructed at character offsets.

The two Dutch models used RobBERT-2023 \citep{delobelle2024robbert2023} as their encoder. We
selected it after comparing RobBERT-2023 with MedRoBERTa.nl \citep{verkijk2021medroberta} on the
hospital development corpus. The English model used RoBERTa-base \citep{liu2019roberta}. The
resulting models were \texttt{meddeid-dutch-uza}, trained on annotated hospital text;
\texttt{meddeid-dutch-synth}, trained only on synthetic Dutch text; and
\texttt{meddeid-english-synth}, trained only on synthetic English text from the \texttt{en-GB} and
\texttt{en-US} profiles.

All three runs used the same MedDeID training implementation, seed, optimiser, encoder and
classification-head learning rates, weight decay, effective batch size and validation metric.
Microbatching, numerical precision, warm-up and early-stopping settings were adapted to the encoder
and hardware. Model selection maximised validation entity-level F1. After selecting the epoch
count, we independently reinitialised each base encoder and refitted it for that fixed number of
epochs on its complete development corpus: 17 epochs for each Dutch model and four for the English
model. All test benchmarks were withheld during model selection and evaluated only after the final
refit. Hyperparameter tuning was deliberately moderate because our effort focused on data quality;
complete configurations and model-selection records for all three models are reported in
Supplementary Table \SuppTabModelSettings.

\textbf{External English benchmark preparation.} We evaluated the English model on two public
synthetic benchmarks.
Technetium-I contributes a held-out test split of 74,700 multi-section clinical notes and 1,161,437
PII annotations \citep{temlm2026technetium}. We used only the published test split and mapped its
seven observed test-set entity types to the MedDeID schema; its training and validation splits were
not downloaded or used. ASQ-PHI contains short clinician-style search queries
\citep{weatherhead2026asqphi}. We included all 1,051 queries, of which 219 were hard negatives.
We converted all 2,973 annotations to character offsets in the MedDeID format. Of these, 2,972
matched the source text directly, while one required apostrophe normalisation. The external
benchmarks and MedDeID group identifiers differently. We therefore measured whether the correct
text was identified, regardless of the label assigned to it (label-agnostic annotation-character
recall). Because ASQ-PHI does not annotate explicit ages below the HIPAA threshold, we identified
these expressions in the source text using the same rules for every system, regardless of output
labels. We included numeric ages in year-old or month-old phrases, \texttt{yo} or \texttt{y} shorthand,
\texttt{age N}, compact age--sex expressions, or over/under age thresholds; overlapping matches
were merged. The resulting set comprised 8,745 non-gold characters in 878 queries (numeric values 5--88, including
two infant ages expressed in months) and did not overlap any gold annotation. For every system, the
sensitivity calculation removed redacted characters inside these expressions from the numerator,
irrespective of the predicted label, while retaining the original denominator of 119,651 non-PII
characters. Qualitative descriptors such as \textit{elderly} and \textit{adolescents} were not
excluded. Full mappings, label distributions and results after excluding these ages are reported in Supplementary
Section S10.

\textbf{Computing environment.} Hospital data preparation and CPU inference used an Azure Standard
DC4as v5 virtual machine with four vCPUs. The hospital model was trained and GPU inference was run
on an Azure Standard NC4as T4 v3 virtual machine with one NVIDIA T4. Both synthetic models were
trained on an Apple M4 Pro GPU. Primary-care validation ran on an isolated four-vCPU Google Cloud
virtual machine without a GPU. Complete model and timing configuration values are reported in
Supplementary Tables \SuppTabModelTimingRange.

\textbf{Post-processing and pseudonymisation.} To compare the systems, we first used each one to
identify PII spans. We then applied MedDeID's
post-processing to every system, optionally using known patient and caregiver names from the
metadata (Fig. 1). This step joins adjacent detections, extends
incomplete detections when the surrounding text follows a known format, and
handles repeated information consistently within a document. For example, if a
system detects the day and month of a date but misses the year, MedDeID's post-processing
adds the year to the detected span. Comparative recall and non-PII redaction were calculated from
these post-processed outputs, and warm processing time included these steps. Pseudonymisation was a
separate downstream transformation and was evaluated independently.

After detection, dates and ages can be replaced with less identifying values that retain clinical
meaning. All dates for one patient or hospital stay are shifted by the same number of days. We
refer to this number as the date shift or offset. This changes the calendar dates but preserves the
order of events and the time between them. Less precise expressions, such as a month or season, are
shifted as date ranges so that the output does not imply an exact day that was not present in the
source text. A separate shift should be used for each patient or hospital stay rather than applying
one shift to the entire dataset. MedDeID warns users when they choose a shift of one year or less.
If reversibility is required, the offset should be held separately under appropriate access
controls so that authorised users can interpret shifted dates without exposing the mapping with the
released record. Ages are made less precise according to the patient's age, while finer detail is
retained for infants and young children (Table \SuppTabAgeGranularity).

\textbf{Belgian DEDUCE.} The Dutch DEDUCE tool \citep{menger2018deduce}
was upgraded for the Belgian context by replacing its lookup lists with
Belgian Dutch \emph{and} French name, place and institution lists, and
by incorporating the logic corrections identified during annotation,
which were first implemented in MedDeID's post-processing layer and
subsequently integrated into the rule base itself.

\textbf{Comparator systems.} We compared: the two human annotators; the hospital-trained
(\texttt{meddeid-dutch-\allowbreak{}uza}) and synthetic-trained
(\texttt{meddeid-dutch-\allowbreak{}synth}) versions of
our RoBERTa transformer; Belgian DEDUCE and the original Dutch DEDUCE
\citep{menger2018deduce}, which are rule-based; \texttt{deidentify}, a
published Dutch de-identification
model \citep{trienes2020comparing}; a locally deployed generative large language model
(Qwen3-8B) with deliberately light prompt engineering, reflecting realistic local use in a healthcare institution; a
zero-shot neural NER model (GLiNER-PII) \citep{zaratiana2024gliner}; and two general-purpose neural
PII detectors (OpenAI Privacy Filter and OpenMed multilingual). The Qwen prompt used the annotation
definitions in Supplementary Table \SuppTabAnnotationLabels{} within a Dutch system prompt that described the model as a
medical text-annotation assistant and required a JSON object containing a
\texttt{spans} list. Each item had to contain the exact source substring in
\texttt{annotated\_text} and one allowed \texttt{label}; overlapping or nested entities and typo
correction were prohibited, and notes without PII had to return an empty list. Two fixed few-shot
examples illustrated patient and other-person names, ages, dates and a healthcare organisation.
The original note text was appended unchanged as the user input.

Qwen3-8B was quantised and served locally through Ollama, allowing the same evaluation interface to
run without transferring clinical text to an external service. The evaluated configuration used the
Ollama tag \texttt{qwen3:8b}, Q4\_K\_M GGUF quantisation with temperature 0.6, top-$p$ 0.95, an
8,000-token output limit, two concurrent workers and Qwen thinking enabled. Thinking blocks were
removed before JSON parsing. Returned substrings were aligned to character offsets by exact
matching, with a case- and whitespace-tolerant fallback for near-verbatim copies. The complete
prompt template, label definitions, two fixed examples, and the code used to construct, parse and
align the Qwen output are available in the standalone
\href{https://github.com/stighellemans/deid-battery}{\texttt{deid-battery} reproducibility
	repository}.

\textbf{Inference-time evaluation.} Warm end-to-end inference time was measured only on the full
300-note hospital benchmark. ``Warm'' means that one-time system setup and model loading had
already been completed. The timing included processing all notes, writing predictions, applying
the shared post-processing and writing the final outputs; it excluded setup and the preliminary
warm-up pass. Supplementary Table \SuppTabTiming{} reports the complete timing results for this evaluated deployment.

\textbf{Stability analysis.} We used a deterministic coverage-selection procedure to
select 100 notes from each of the three test sets (300 notes in total). We perturbed these notes
along six dimensions: name source (original, reshuffled real names or synthetically generated
names), name capitalisation, name format (full name, first name only, initials only, first name plus
initials or title), date format, date value shifted into the past or future, and age format. Each
analysis included only the selected notes containing the relevant type of identifier. Individual
analyses therefore included 19--98 hospital notes, 23--97 synthetic notes and 28--91 primary-care
notes. The number of relevant identifiers also varied; Supplementary Table \SuppTabStabilitySamples{} reports both note
and identifier counts. Recall was computed as
the fraction of perturbed target spans still detected with the correct
category. Confidence intervals for recall and degradation were obtained from
10,000 replicates of a note-level cluster bootstrap: complete notes were resampled, with all target
spans from a selected note kept together and each baseline--perturbation pair
preserved. Degradation was tested by a one-sided permutation test at $\alpha=0.05$, with the note,
rather than each individual identifier, treated as the independent unit. False-discovery
rates were controlled separately for each model across the 27 cross-scope cells using the
Benjamini--Hochberg procedure.
Cells with fewer than five paired target spans or fewer than five contributing
notes are excluded from all reported results.

\textbf{Pseudonymisation evaluation.} We evaluated pseudonymisation in two settings. First, to
isolate transformation-layer performance from PII detection, the evaluator received every gold
Date and Age\_Birthdate span directly. Second, to evaluate the full pipeline, it used the
metadata-enabled \texttt{meddeid-dutch-synth} predictions for each test set. A gold target failed
end to end unless one predicted span covered it completely, had the correct label and produced a
protocol-valid transformation. We separately counted gold spans with unredacted characters:
targets for which at least one original character was not covered by any predicted redaction.
These cases are included among end-to-end failures; other end-to-end failures were fully redacted
but failed because the identifier was split across predictions, assigned the wrong label or
transformed incorrectly. All rates used the total number of gold Date and Age\_Birthdate spans as
the denominator. Both evaluations used a fixed document creation date of 15 January 2025, a
$+371$-day shift and birthdate-to-age replacement.

\textbf{Statistics and reproducibility.} Core PII recall and non-PII redaction rate are reported for
every system on the three Dutch test sets. Because the external English benchmarks lack the
subannotations required to calculate core PII recall, we report annotation-character recall and
non-PII redaction instead. Unless stated otherwise, numbers in parentheses in the Results and
Supplementary Information are 95\% confidence intervals. We calculated these intervals with the percentile
method from 10,000 bootstrap samples of complete documents. We used this method for every
confidence interval reported.

In each bootstrap sample, complete documents were drawn with replacement. All gold annotations and
system predictions from a selected document were kept together. We then calculated each rate from
the total character counts across the sampled documents, rather than averaging document-level
percentages. The same sampled documents were used for all systems on a benchmark, allowing paired
confidence intervals for between-system differences. Documents without PII remained in the
denominator of the non-PII redaction rate. The comparison between the hospital model and the
annotators was descriptive; we did not test whether their performance was equivalent.

\begin{minipage}{\textwidth}
	Confidence intervals for the stability experiments were obtained by the separate paired bootstrap
	described above, and degradation was tested by a one-sided permutation test at $\alpha=0.05$, with
	Benjamini--Hochberg correction across the 27 cells separately for each model. Reporting follows
	TRIPOD+AI where applicable \citep{collins2024tripodai}.
\end{minipage}

\textbf{Use of generative artificial intelligence.} During preparation of this manuscript, the
authors used OpenAI Codex for language editing, structural revision, consistency checking and audit
support. No confidential patient information, patient-level clinical text or other identifiable
clinical data were entered into an unapproved artificial-intelligence service. All outputs were
reviewed and verified by the authors, who remain fully responsible for the manuscript's content.

\hypertarget{data-availability}{%
	\subsection{Data availability}\label{data-availability}}

Materials that contain no real patient data are openly available: the Dutch and English synthetic
training corpora and benchmarks, annotation guideline and worked examples can be accessed through
the MedDeID collection on Hugging Face
(\url{https://huggingface.co/collections/stighellemans/meddeid}) and are archived in versioned Zenodo
records \citep{hellemans2026meddeiddata,hellemans2026meddeidenglishdata}. The public Dutch resources
contain 6,793 notes (6,493 training; 300 benchmark), and the English resources contain 6,700
training notes and a 300-note benchmark. Technetium-I and ASQ-PHI remain available from their
original repositories \citep{temlm2026technetium,weatherhead2026asqphi}. By contrast, the real-EHR
materials---300 hospital benchmark notes, 4,470 hospital training notes, 100 primary-care validation
notes, their annotations and the hospital-trained model weights---cannot be shared publicly or on
request because of patient privacy, GDPR, ethics and institutional-governance restrictions
\citep{gdpr2016}.

\hypertarget{code-availability}{%
	\subsection{Code availability}\label{code-availability}}

MedDeID suite version 0.3.0, including \texttt{meddeid} version 0.4.0, is available at
\url{https://github.com/stighellemans/meddeid-suite}. This release provides the software for local
inference, post-processing and pseudonymisation, as well as for training, synthetic-data generation,
evaluation and language profiling \citep{hellemans2026meddeidsoftware}.
The study-specific benchmarking and comparator orchestration code is maintained separately in
\texttt{deid-battery} (version 0.1.0;
\url{https://github.com/stighellemans/deid-battery}) \citep{hellemans2026deidbattery}. The Belgian
DEDUCE comparator version used in this study is retained as
\texttt{meddeid-study-v1.0.0} at \url{https://github.com/stighellemans/belgian-deduce}
\citep{menger2026belgiandeducemeddeid}. The synthetic-trained Dutch and English model weights
(version 1.0.0) are available from Hugging Face at
\url{https://huggingface.co/stighellemans/meddeid-dutch-synth} and
\url{https://huggingface.co/stighellemans/meddeid-english-synth}, respectively
\citep{hellemans2026meddeiddutchmodel,hellemans2026meddeidenglishmodel}.

\renewcommand{\bibsection}{
	\subsection*{References}}
\bibliography{references}

\clearpage
\hypertarget{acknowledgements}{%
	\subsection{Acknowledgements}\label{acknowledgements}}

S.H. is supported by the Research Foundation Flanders (FWO) under Grant No.1SA3226N. This work was
supported by the Flemish Government (Flanders AI Research Program). The funders had no role in
study design, data collection, analysis, interpretation or preparation of the manuscript. We thank
the participating general practice and Co-Medic for enabling the primary-care validation and the
UZA Data Innovation Lab for the secured research environment.

\hypertarget{author-contributions}{%
	\subsection{Author contributions}\label{author-contributions}}

S.H. conceived the study; developed the MedDeID software, models, datasets and annotation
guideline; performed the primary annotation, experiments, analyses and visualisation; and drafted
the manuscript. T.S. performed the second hospital-benchmark annotation, contributed to the
annotation guideline and reviewed and edited the manuscript. E.S. facilitated the technical setup
at the UZA Data Innovation Lab, contributed methodological input and reviewed the final manuscript.
P.J. supported the ethics-approval process and enabled the study at UZA, and reviewed and edited
the final manuscript. P.M. and K.L. contributed to the methodology, interpretation and supervision,
and reviewed and edited the manuscript. All authors approved the final manuscript.

\hypertarget{competing-interests}{%
	\subsection{Competing interests}\label{competing-interests}}

The authors declare no financial or non-financial competing interests.

\clearpage

\begin{table}[p]
	\hypertarget{tables-and-figures}{%
		\subsection{Tables and figures}\label{tables-and-figures}}

	\centering
	\phantomsection\label{table-1}
	\caption*{\textbf{Table 1 \textbar{} Dataset provenance, construction and review}}
	\fontsize{6.5}{7.6}\selectfont
	\setlength{\tabcolsep}{2.2pt}
	\renewcommand{\arraystretch}{1.22}
	\begin{tabular}{@{}>{\raggedright\arraybackslash}m{0.115\linewidth}
			>{\centering\arraybackslash}m{0.100\linewidth}
			>{\centering\arraybackslash}m{0.121\linewidth}
			>{\centering\arraybackslash}m{0.140\linewidth}
			>{\centering\arraybackslash}m{0.097\linewidth}
			>{\centering\arraybackslash}m{0.117\linewidth}
			>{\centering\arraybackslash}m{0.130\linewidth}
			>{\centering\arraybackslash}m{0.090\linewidth}@{}}
		\toprule
		\textbf{Dataset} & \textbf{Language}
		                 & \textbf{Training}
		                 & \textbf{Benchmark}
		                 & \shortstack{\textbf{Real or}\\\textbf{synthetic}}
		                 & \shortstack{\textbf{Sub-}\\\textbf{annotation}\\\textbf{layer}}
		                 & \shortstack{\textbf{Injected}\\\textbf{metadata}}
		                 & \textbf{Access}                                                                                                                                                                                                                                    \\
		\midrule
		\textbf{Dutch hospital}
		                 & Dutch
		                 & \shortstack{\textbf{4,470}\\\textbf{notes}\\\textcolor{gray}{1 reviewer}}
		                 & \shortstack{\textbf{300 notes}\\\textcolor{gray}{2 reviewers}\\\textcolor{gray}{Adjudicated}}
		                 & Real                                                                                            & Yes                                                             & \shortstack{Patient and\\caregiver\\names} & Governed                          \\
		\addlinespace[3pt]
		\textbf{Dutch primary care}
		                 & Dutch                                                                                           & ---
		                 & \shortstack{\textbf{100 notes}\\\textcolor{gray}{1 reviewer}}
		                 & Real                                                                                            & Yes                                                             & \shortstack{Patient and\\caregiver\\names} & Governed                          \\
		\addlinespace[3pt]
		\textbf{Dutch synthetic}
		                 & Dutch                                                                                           & \shortstack{\textbf{6,493 notes}\\\textcolor{gray}{Unreviewed}}
		                 & \shortstack{\textbf{300 notes}\\\textcolor{gray}{1 reviewer}}
		                 & Synthetic                                                                                       & Yes                                                             & \shortstack{Patient and\\caregiver\\names} & \textbf{Open\textsuperscript{*}}  \\
		\addlinespace[3pt]
		\textbf{English synthetic}
		                 & English                                                                                         & \shortstack{\textbf{6,700 notes}\\\textcolor{gray}{Unreviewed}}
		                 & \shortstack{\textbf{300 notes}\\\textcolor{gray}{1 reviewer}}
		                 & Synthetic                                                                                       & Yes                                                             & \shortstack{Patient and\\caregiver\\names} & \textbf{Open\textsuperscript{*}}  \\
		\addlinespace[3pt]
		Technetium-I
		                 & English                                                                                         & ---
		                 & \shortstack{\textbf{74,700 notes}\\\textcolor{gray}{Unreviewed}}
		                 & Synthetic                                                                                       & No                                                              & None                                       & Open\citep{temlm2026technetium}   \\
		\addlinespace[3pt]
		ASQ-PHI
		                 & English                                                                                         & ---
		                 & \shortstack{\textbf{1,051 queries}\\\textcolor{gray}{3 reviewers\textsuperscript{\textdagger}}}
		                 & Synthetic                                                                                       & No                                                              & None                                       & Open\citep{weatherhead2026asqphi} \\
		\bottomrule
	\end{tabular}
	\begin{minipage}{\linewidth}
		\vspace{4pt}\footnotesize
		The study links governed clinical data with openly shareable synthetic resources. The real Dutch
		hospital and primary-care benchmarks test performance in distinct care settings, while the Dutch
		and English synthetic resources make training and evaluation reproducible beyond those settings.
		Character-level sub-annotations and patient or caregiver metadata are available for the MedDeID
		benchmarks, but not for the two external English benchmarks.
		\par
		\hangindent=1.2em\hangafter=1\noindent\textsuperscript{*}\hspace{0.25em}MedDeID datasets created for this study.
		\par
		\hangindent=1.2em\hangafter=1\noindent\textsuperscript{\textdagger}\hspace{0.25em}In the original ASQ-PHI study, three clinicians or domain experts each audited 100 records
		(300 of 1,051); this audit was not conducted by the present authors, and the complete dataset was not reviewed.
	\end{minipage}
\end{table}

\clearpage
\begin{figure}[!t]
	% Each submission figure is isolated on a forced float page.  The clear-page
	% boundaries below fix the sequence and keep every caption with its artwork.
	\centering
	\noindent\includegraphics[width=\linewidth,height=0.45\textheight,keepaspectratio]{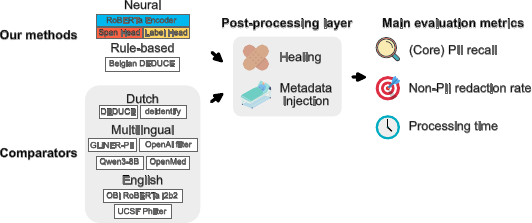}\par
	\smallskip
	\noindent\begin{minipage}{\linewidth}
		\captionof*{figure}{\textbf{Fig. 1 \textbar{} De-identification systems, shared post-processing and evaluation metrics.}
			Our compact neural de-identifiers
			combine a transformer encoder with separate heads for token-level span boundaries and span labels;
			the Dutch models use RobBERT-2023, whereas the separately trained English model uses RoBERTa-base.
			Belgian DEDUCE instead adapts the rule-based Dutch DEDUCE system to the Belgian context. We compared
			these approaches with published Dutch systems (DEDUCE and \texttt{deidentify}), zero-shot or
			general-purpose multilingual detectors (GLiNER-PII, OpenAI Privacy Filter and OpenMed), a locally
			deployed generative model (Qwen3-8B), and English systems (OBI RoBERTa i2b2 and UCSF Philter).
			Although these systems detect PII in different ways, each produces candidate character spans that
			follow the same MedDeID post-processing path. Healing joins or extends detections using deterministic
			rules, while metadata injection adds detections for known patient and caregiver names without
			inserting metadata into the text. This common path allows the evaluation to answer three practical
			questions: how much PII is found, how much non-PII text is removed, and how quickly the notes are
			processed. Dutch benchmarks use core PII recall; English benchmarks use annotation-character recall
			because the external English datasets lack sub-annotations.}
	\end{minipage}\par
\end{figure}

\clearpage
\begin{table}[htbp]
	\centering
	\phantomsection\label{table-2}
	\caption*{\textbf{Table 2 \textbar{} De-identification performance on three annotated test sets}}
	\scriptsize
	\setlength{\tabcolsep}{2.3pt}
	\renewcommand{\arraystretch}{1.20}
	\begin{tabular}{@{}>{\raggedright\arraybackslash}m{0.336\linewidth}*{6}{>{\centering\arraybackslash}m{0.095\linewidth}}@{}}
		\toprule
		                                                         & \multicolumn{3}{c}{\shortstack{\textbf{Core PII recall}\\($\uparrow$ higher is better)}} &
		\multicolumn{3}{c}{\shortstack{\textbf{Non-PII redaction rate}\\($\downarrow$ lower is better)}}                                                                                                                                                                                                  \\
		\cmidrule(lr){2-4}\cmidrule(l){5-7}
		\textbf{Method}                                          & \textbf{Hospital}                                                                        & \textbf{Synth.}                              & \shortstack{\textbf{Primary}\\\textbf{care}} &
		\textbf{Hospital}                                        & \textbf{Synth.}                                                                          & \shortstack{\textbf{Primary}\\\textbf{care}}                                                                                                \\
		\midrule
		\textbf{meddeid-dutch-uza (ours)}                        & \textbf{98.9}                                                                            & 96.6                                         & 87.0                                         & \textbf{0.24} & 0.39          & \textbf{0.29} \\
		\addlinespace[1pt]
		\textbf{meddeid-dutch-synth (ours)}                      & 96.1                                                                                     & \textbf{99.8}                                & \textbf{90.3}                                & 0.83          & \textbf{0.28} & 1.22          \\
		\addlinespace[1pt]
		\textbf{Belgian DEDUCE (ours)}                           & 88.0                                                                                     & 70.2                                         & 74.9                                         & 0.63          & 1.70          & 0.90          \\
		\addlinespace[1pt]
		Qwen3-8B (Yang et al.)\citep{yang2025qwen3}              & 84.2                                                                                     & 90.9                                         & 60.8                                         & 0.88          & 1.28          & 0.31          \\
		\addlinespace[1pt]
		deidentify (Trienes et al.)\citep{trienes2020comparing}  & 86.4                                                                                     & 76.7                                         & 68.6                                         & 0.56          & 0.37          & 5.73          \\
		\addlinespace[1pt]
		GLiNER-PII (Zaratiana et al.)\citep{zaratiana2024gliner} & 76.6                                                                                     & 83.0                                         & 70.4                                         & 2.11          & 1.58          & 3.42          \\
		\addlinespace[1pt]
		DEDUCE (Menger et al.)\citep{menger2018deduce}           & 75.9                                                                                     & 45.9                                         & 57.2                                         & 0.50          & 0.58          & 0.54          \\
		\addlinespace[1pt]
		OpenAI privacy filter                                    & 72.1                                                                                     & 58.9                                         & 66.9                                         & 0.61          & 0.52          & 11.46         \\
		\addlinespace[1pt]
		OpenMed multilingual filter                              & 49.1                                                                                     & 53.0                                         & 37.1                                         & 0.96          & 1.01          & 8.25          \\
		\midrule
		Annotator 1                                              & 98.8                                                                                     & ---                                          & ---                                          & 0.17          & ---           & ---           \\
		\addlinespace[1pt]
		Annotator 2                                              & 98.5                                                                                     & ---                                          & ---                                          & 0.36          & ---           & ---           \\
		\bottomrule
	\end{tabular}
	\begin{minipage}{\linewidth}
		\vspace{4pt}\footnotesize
		The hospital-trained model has the highest recall on the hospital benchmark, whereas the
		synthetic-trained model has the highest recall on the synthetic and primary-care benchmarks. Both
		models retain low non-PII redaction, so their recall is not achieved by indiscriminately removing
		ordinary text. Values are percentages. Bold indicates the best-performing model in each column; annotators are
		excluded from this comparison. Results use metadata; dashes indicate that a system was not
		evaluated. Corresponding 95\% document-clustered bootstrap confidence intervals are reported in
		Supplementary Table \SuppTabDutchUncertainty. Metadata-free results and the non-PII redaction decomposition are reported
		in Supplementary Tables \SuppTabDutchPointRange.
	\end{minipage}
\end{table}

\clearpage
\begin{table}[htbp]
	\centering
	\phantomsection\label{table-3}
	\caption*{\textbf{Table 3 \textbar{} External synthetic English portability benchmarks}}
	\scriptsize
	\setlength{\tabcolsep}{1.5pt}
	\renewcommand{\arraystretch}{1.20}
	\begin{tabular}{@{}>{\raggedright\arraybackslash}m{0.37\linewidth}*{4}{>{\centering\arraybackslash}m{0.145\linewidth}}@{}}
		\toprule
		                                                                 & \multicolumn{2}{c}{\textbf{Technetium-I}}  & \multicolumn{2}{c}{\textbf{ASQ-PHI}}                                            \\
		\cmidrule(lr){2-3}\cmidrule(l){4-5}
		\textbf{Method}                                                  & \shortstack{\textbf{Recall}\\($\uparrow$)} &
			\shortstack{\textbf{Non-PII}\\\textbf{redaction} ($\downarrow$)} &
			\shortstack{\textbf{Recall}\\($\uparrow$)}                       &
			\shortstack{\textbf{Non-PII}\\\textbf{redaction}\textsuperscript{*}\\($\downarrow$)}                                                                                                                                \\
			\midrule
			\textbf{meddeid-english-synth (ours)}                            & \textbf{99.730}                            & 1.606                                & \textbf{98.90} & 6.21                    \\
		\addlinespace[1pt]
		GLiNER Multilingual PII                                          & 97.773                                     & 3.616                                & 96.32          & 11.52                   \\
		\addlinespace[1pt]
		OBI RoBERTa i2b2                                                 & 91.595                                     & 0.868                                & 95.45          & 2.01                    \\
		\addlinespace[1pt]
		OpenAI Privacy Filter                                            & 96.328                                     & 0.870                                & 63.48          & \textbf{1.19}           \\
		\addlinespace[1pt]
		OpenMed Multilingual Privacy Filter                              & 92.032                                     & 1.086                                & 75.36          & 4.39                    \\
		\addlinespace[1pt]
		OpenMed SuperClinical 434M                                       & 94.828                                     & 1.723                                & 61.67          & 4.85                    \\
		\addlinespace[1pt]
		UCSF Philter                                                     & 88.295                                     & \textbf{0.700}                       & 70.89          & 1.80                    \\
		\bottomrule
	\end{tabular}
	\begin{minipage}{\linewidth}
		\vspace{4pt}\footnotesize
		The synthetic-trained MedDeID English model has the highest recall on both benchmarks. Systems with
		lower non-PII redaction generally miss more annotated PII. \textsuperscript{*}Raw non-PII redaction
		rates are shown. ASQ-PHI leaves explicit ages below 90 unannotated, so redacting these ages increases
		this rate. To compare the systems fairly, we recalculated the rate for every system after
		excluding the same age characters. MedDeID's rate after excluding these ages was 0.89\%
		(0.75--1.03\%), second to OBI RoBERTa i2b2 at 0.55\% (0.45--0.66\%); the full comparison is
		reported in Supplementary Table \SuppTabASQ{}.
		The comparison therefore supports
		portability to synthetic English data while showing why recall and benchmark taxonomy must be read
		together; it does not establish performance on real English clinical text. Values are percentages.
		Recall is label-agnostic annotation-character recall. Bold indicates the
		best result in each column. Corresponding 95\% document-clustered bootstrap confidence intervals
		are reported in Supplementary Tables \SuppTabEnglishExternalRange. Full taxonomy-aligned sensitivity results are reported
		in Supplementary Section S10.
	\end{minipage}
\end{table}

\clearpage
\begin{figure}[p]
	\centering
	\noindent\includegraphics[width=1.0\linewidth]{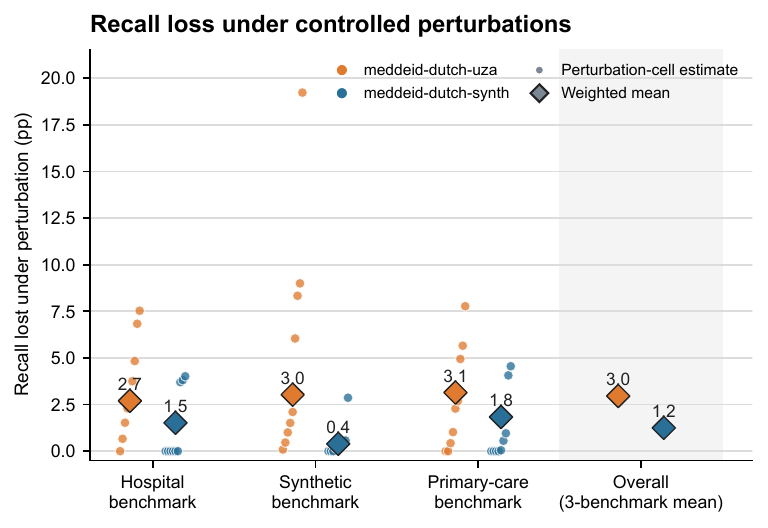}\par
	\smallskip
	\noindent\begin{minipage}{\linewidth}
		\captionof*{figure}{\textbf{Fig. 2 \textbar{} Stability across evaluation datasets.}
			Orange and blue circles show recall loss in each eligible perturbation cell for
			\texttt{meddeid-dutch-uza} and \texttt{meddeid-dutch-synth}, respectively; improvements are plotted
			at zero. Diamonds show the average loss. Within each benchmark, experiments containing more target
			spans contribute more to this average. The overall average gives equal weight to each of the three
			benchmarks. The synthetic-trained model has the smaller mean loss in
			each benchmark and overall (1.2 versus 3.0 percentage points), whereas the hospital-trained model
			shows both a larger average loss and the most extreme perturbation response. Thus, the aggregate
			advantage reflects consistently greater stability rather than a single benchmark. Lower values
			indicate greater stability. Cell-level
			effects, 95\% confidence intervals and adjusted tests are reported in Supplementary Fig.
			\SuppFigStabilityCells{} and Tables \SuppTabStabilityDetailRange.}
	\end{minipage}\par
\end{figure}

\clearpage
\begin{figure}[p]
	\centering
	\noindent\includegraphics[width=1.0\linewidth]{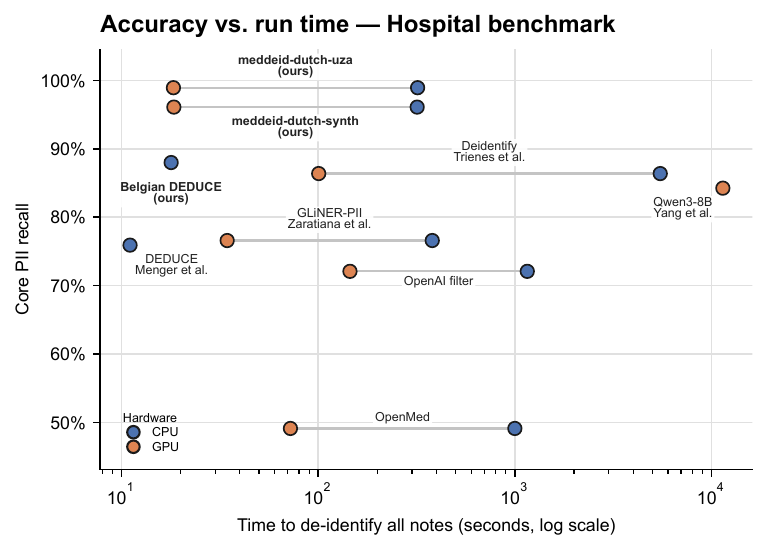}\par
	\smallskip
	\noindent\begin{minipage}{\linewidth}
		\captionof*{figure}{\textbf{Fig. 3 \textbar{} Recall and processing time on the hospital benchmark.}
			Core PII recall versus the time required to de-identify all 300 hospital notes; time is shown on
			a logarithmic scale. Blue and orange markers denote CPU and GPU runs, respectively. Measurements
			for the same system are connected. The compact MedDeID neural models occupy the high-recall,
			low-runtime region, with \texttt{meddeid-dutch-uza} providing the strongest combination. GPU use
			shortens processing for several systems, but the locally deployed Qwen3-8B remains both slower and
			less accurate than the compact models.}
	\end{minipage}\par
\end{figure}

\clearpage
\begin{figure}[!t]
	\centering
	% The exported figure now has a tight page box; no additional crop is needed.
	\noindent\includegraphics[width=\linewidth,height=0.64\textheight,keepaspectratio]{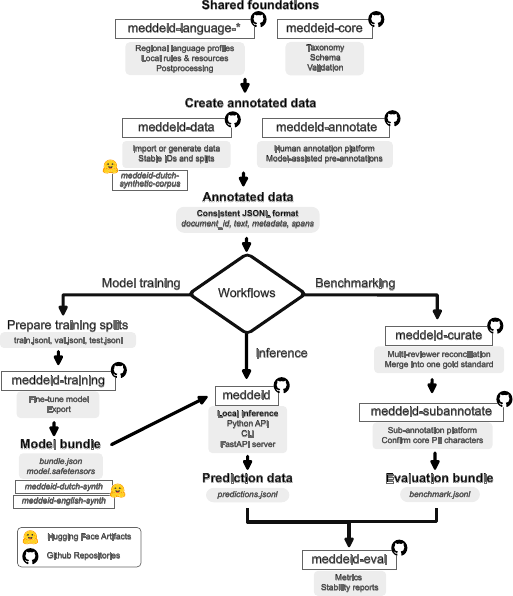}\par
	\smallskip
	\noindent\begin{minipage}{\linewidth}
		\captionof*{figure}{\textbf{Fig. 4 \textbar{} MedDeID software and data workflow.}
			Shared schemas and language profiles connect data import or generation, annotation, training,
			inference, benchmark construction and evaluation through consistent JSONL and versioned model
			artefacts. Annotated data can therefore support either model development or benchmark creation;
			trained models feed local inference, and their predictions return to the same evaluation contract.
			This shared foundation turns otherwise separate tools into a repeatable workflow in which language
			profiles, datasets or models can change without rebuilding every stage. GitHub and Hugging Face symbols indicate public code and model or data artefacts,
			respectively.}
	\end{minipage}\par
\end{figure}

\end{document}

\hypertarget{supplementary-information}{%
	\section{Supplementary Information}\label{supplementary-information}}

\textbf{\manuscripttitle}

Hellemans et al.

This Supplementary Information provides extended dataset summaries, benchmark tables, additional
analyses and figures, and the complete English annotation guideline in its original styled layout.
Both the Dutch- and English-language versions of the annotation guideline are available in the
Zenodo archive \citep{hellemans2026meddeiddata,hellemans2026meddeidenglishdata}. All study methods
are reported in the main manuscript. Personally identifiable information is abbreviated as PII
throughout; Antwerp University Hospital is abbreviated as UZA.

\textbf{Contents}

\begin{itemize}
	\tightlist
	\item
	      S1. Dataset summary --- Fig. \SuppFigDatasetSummary, Table \SuppTabDatasetDepartments
	\item
	      S2. Annotation scheme --- Tables \SuppTabAnnotationRange
	\item
	      S3. Benchmark results --- Tables \SuppTabDutchBenchmarkRange
	\item
	      S4. Per-label performance --- Table \SuppTabLabelRecall, Figs \SuppFigLabelRange
	\item
	      S5. Label fidelity --- Table \SuppTabLabelFidelity, Fig. \SuppFigLabelConfusion
	\item
	      S6. Perturbation stability --- Tables \SuppTabStabilityRange, Figs \SuppFigStabilityRange
	\item
	      S7. Non-PII redaction --- Fig. \SuppFigNonPIIRedaction
	\item
	      S8. Model and timing specifications --- Tables \SuppTabModelSettings{} and \SuppTabTiming
	\item
	      S9. Pseudonymisation validation --- Tables \SuppTabPseudonymRange, Fig. \SuppFigPseudonymFailureModes
	\item
	      S10. English portability --- Tables \SuppTabEnglishRange
	\item
	      Appendix A. Annotation guidelines
\end{itemize}

\begin{center}\rule{0.5\linewidth}{0.5pt}\end{center}

\hypertarget{s1.-dataset-and-gold-standard-summary}{%
	\subsection{S1. Dataset summary}\label{s1.-dataset-and-gold-standard-summary}}

We compared annotation volume and mapped label composition across all six evaluation datasets (Fig.
\SuppFigDatasetSummary). The clinical department or specialty composition of the hospital benchmark
and development corpus is reported in Table \SuppTabDatasetDepartments. Annotation density divides
the number of gold spans by all documents or queries, including the 219 ASQ-PHI hard-negative
queries without annotations. The composition analysis maps the canonical labels to nine broad
categories solely to support comparison across taxonomies. It is descriptive: differences can
reflect clinical setting, document type, synthetic-generation design and source-label coverage, and
should not be interpreted as estimates of identifier prevalence in clinical practice.

\textbf{Hospital benchmark composition.} Of the 579,920 characters in the 300-note hospital
benchmark, 55,379 (9.55\%) fall within a gold annotation and 42,093 contribute to the core PII
denominator; the remaining 524,541 characters form the denominator for the non-PII redaction rate.
For comparison, the open synthetic benchmark contains 419,578 characters, including 37,443 core
PII characters.

\clearpage
\begin{center}
	\begin{minipage}{\linewidth}
		\centering
		\includegraphics[width=\linewidth]{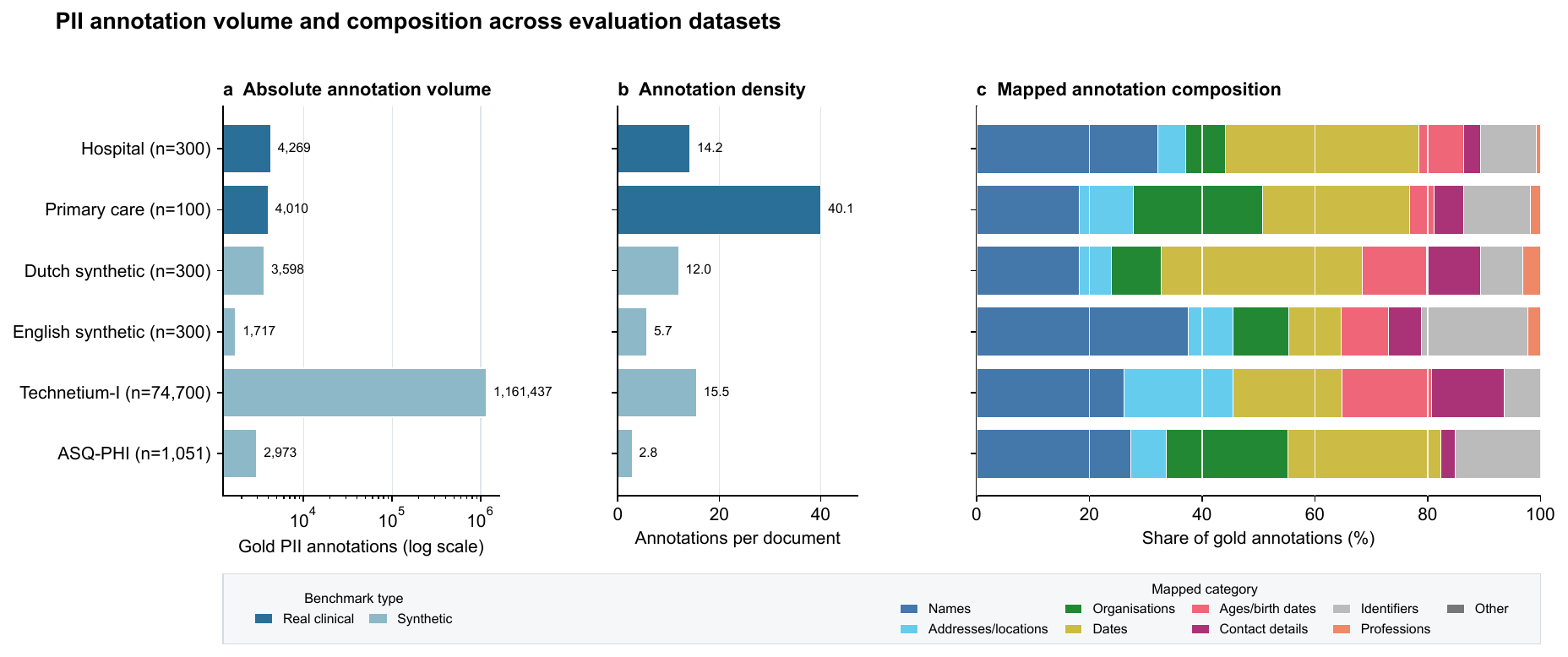}
		\suppfigcaption{\textbf{Fig. \SuppFigDatasetSummary{} \textbar{} PII annotation volume and composition across evaluation datasets.}
			\textbf{a}, Absolute numbers of mapped gold PII spans on a logarithmic scale.
			\textbf{b}, Gold spans per document or query.
			\textbf{c}, Percentage composition after mapping canonical labels to nine broad categories.
			Although the primary-care benchmark contains 100 notes, compared with 300 in the hospital,
			Dutch synthetic and English synthetic benchmarks, its longer notes yield the highest annotation
			density (40.1 gold annotations per note).
			The benchmarks differ by several orders of magnitude and by identifier mix, showing why performance
			must be tested across datasets rather than inferred from a single benchmark. Counts refer to
			annotation records; overlapping source annotations are retained. In the dataset labels, $n$ denotes
			the number of evaluated documents or queries.}
	\end{minipage}
\end{center}

\hypertarget{supp-table-dataset-departments}{%
	\subsubsection{Table \SuppTabDatasetDepartments{} \textbar{} Clinical specialty distribution in the hospital datasets}\label{supp-table-dataset-departments}}

\begin{longtable}[]{@{}>{\raggedright\arraybackslash}p{0.62\linewidth}>{\centering\arraybackslash}p{0.14\linewidth}>{\centering\arraybackslash}p{0.14\linewidth}@{}}
	\toprule
	\textbf{Clinical department or specialty}       & \textbf{Hospital benchmark, n} & \textbf{Development corpus, n} \\
	\midrule
	\endhead
	Not available                                   & 42                             & 357                            \\
	Cardiology                                      & 20                             & 318                            \\
	Otorhinolaryngology (ear, nose and throat)      & 19                             & 218                            \\
	Paediatrics                                     & 17                             & 210                            \\
	Orthopaedics                                    & 17                             & 187                            \\
	Ophthalmology                                   & 16                             & 118                            \\
	Emergency admissions                            & 15                             & 209                            \\
	Gastroenterology and hepatology                 & 14                             & 161                            \\
	Neurology                                       & 11                             & 122                            \\
	Pulmonology                                     & 9                              & 125                            \\
	Thoracic and vascular surgery                   & 9                              & 109                            \\
	Urology                                         & 9                              & 78                             \\
	Neurosurgery                                    & 8                              & 77                             \\
	Magnetic resonance imaging (MRI)                & 8                              & 46                             \\
	Oncology                                        & 7                              & 141                            \\
	Gynaecology                                     & 7                              & 93                             \\
	Haematology                                     & 6                              & 100                            \\
	Intensive care                                  & 5                              & 291                            \\
	Abdominal, paediatric and plastic surgery       & 5                              & 82                             \\
	Cardiac surgery                                 & 5                              & 75                             \\
	Hepatobiliary, transplant and endocrine surgery & 5                              & 57                             \\
	Endocrinology, diabetology and metabolism       & 4                              & 103                            \\
	Physical and rehabilitation medicine            & 4                              & 37                             \\
	Dermatology                                     & 3                              & 86                             \\
	Immunology, allergology and rheumatology        & 3                              & 56                             \\
	Radiology                                       & 3                              & 55                             \\
	Gynaecological oncology                         & 3                              & 41                             \\
	Nuclear medicine                                & 3                              & 36                             \\
	Pain centre                                     & 3                              & 35                             \\
	Oral and maxillofacial surgery                  & 3                              & 18                             \\
	Fertility medicine                              & 3                              & 12                             \\
	General internal medicine                       & 2                              & 275                            \\
	Neonatology                                     & 2                              & 26                             \\
	Central phlebotomy                              & 2                              & 22                             \\
	Dentistry                                       & 2                              & 6                              \\
	Anaesthesiology                                 & 1                              & 68                             \\
	Nephrology outpatient clinic                    & 1                              & 39                             \\
	Sleep centre                                    & 1                              & 29                             \\
	Obstetrics                                      & 1                              & 20                             \\
	Geriatrics                                      & 1                              & 13                             \\
	Cardiac rehabilitation                          & 1                              & 9                              \\
	Psychiatry                                      & 0                              & 240                            \\
	Thoracic oncology                               & 0                              & 32                             \\
	Medical genetics                                & 0                              & 14                             \\
	Chronic dialysis centre                         & 0                              & 7                              \\
	Tropical medicine                               & 0                              & 5                              \\
	Multidisciplinary sports medicine centre        & 0                              & 3                              \\
	Clinical biology laboratory                     & 0                              & 2                              \\
	Hearing and speech rehabilitation               & 0                              & 2                              \\
	Stomatology and maxillofacial surgery           & 0                              & 2                              \\
	Sexual Assault Care Centre                      & 0                              & 2                              \\
	Medical check-up                                & 0                              & 1                              \\
	\bottomrule
\end{longtable}

	{\footnotesize
		The clinical department or specialty was available for 258 of 300 notes (86.0\%) in the hospital
		benchmark and 4,113 of 4,470 notes (92.0\%) in the hospital development corpus. The benchmark
		included 40 distinct mapped departments or specialties, compared with 51 in the development corpus.
		Counts use English translations of the source labels; missing or unmapped information is retained as
		``Not available''. Psychiatry and general internal medicine were deliberately enriched in the
		development corpus with small additional note sets. These counts therefore describe the study corpus
		and not the underlying distribution of hospital documentation.\par}

\clearpage
\hypertarget{s2.-label-scheme-summary}{%
	\subsection{S2. Annotation scheme}\label{s2.-label-scheme-summary}}

The annotation scheme has two levels. A span-level label records what type of identifier was found
and, where relevant, the entity to whom it refers (Table \SuppTabAnnotationLabels). A
character-level sub-annotation then records the role of each part of that span (Table
\SuppTabSubannotations). This second layer separates the characters that carry identifying
information from punctuation, titles and clinically useful context captured inside the same span.

The scheme draws on the NIH/NLM Scrubber clinical-text annotation guidelines
\citep{kayaalp2016guidelines,kayaalp2014challenges}, the HIPAA 18-identifier set
\citep{hhs2012deid}, and the GraSCCo/GeMTeX framework \citep{lohr2024grascco,lohr2024grascco_phi},
adapted to the Belgian and Dutch context. The complete rulebook with worked examples is Appendix A.
Use of \texttt{Anonymize\_Other} was deliberately minimised and treated as a signal that a span
required further consideration rather than as a destination label.

The character-level sub-annotation scheme used to define the core PII denominator is illustrated
below. These sub-annotations are applied within the span-level annotation labels described in
Appendix A.

\begin{center}
	\centering
	\begin{minipage}{0.98\linewidth}
		\centering
		{\small\bfseries How sub-annotations define core PII content\par}
		\vspace{5pt}
		\begin{tabular}{@{}>{\raggedright\arraybackslash}p{0.12\linewidth}>{\raggedright\arraybackslash}p{0.59\linewidth}>{\raggedright\arraybackslash}p{0.22\linewidth}@{}}
			\textbf{Example}                                                                                                                                                                       & \textbf{Full annotated span}                                                                                   & \textbf{Characters counted} \\
			\addlinespace[4pt]
			Name                                                                                                                                                                                   &
			\excludedbox{Dr.}\boxsep\spacebox\boxsep\corebox{Emma}\boxsep\spacebox\boxsep\corebox{Peeters}                                                                                         &
			\resultbox{EmmaPeeters}                                                                                                                                                                                                                                                                                                               \\
			                                                                                                                                                                                       & \multicolumn{2}{l}{\scriptsize Excluded: title and both whitespace characters.}                                                              \\
			\addlinespace[6pt]
			Telephone                                                                                                                                                                              &
			\excludedbox{+32}\boxsep\spacebox\boxsep\excludedbox{(0)}\boxsep\corebox{493}\boxsep\spacebox\boxsep\corebox{23}\boxsep\spacebox\boxsep\corebox{20}\boxsep\spacebox\boxsep\corebox{82} &
			\resultbox{493232082}                                                                                                                                                                                                                                                                                                                 \\
			                                                                                                                                                                                       & \multicolumn{2}{l}{\scriptsize Excluded: country code, trunk-prefix notation, punctuation and all whitespace.}                               \\
			\addlinespace[6pt]
			Organisation                                                                                                                                                                           &
			\excludedbox{General}\boxsep\spacebox\boxsep\excludedbox{practice}\boxsep\spacebox\boxsep\corebox{De}\boxsep\spacebox\boxsep\corebox{Linde}                                            &
			\resultbox{DeLinde}                                                                                                                                                                                                                                                                                                                   \\
			                                                                                                                                                                                       & \multicolumn{2}{l}{\scriptsize Excluded: descriptive context and every whitespace character.}                                                \\
		\end{tabular}
		\vspace{-2pt}

		{\footnotesize
			\corebox{Core}\quad included in core PII recall
			\qquad
			\excludedbox{Excluded}\quad omitted from core PII recall\par}
	\end{minipage}
	\par\vspace{1pt}
	\begin{minipage}{0.98\linewidth}
		\normalsize\textbf{Sub-annotation schematic.} The symbol \(\sqcup\) marks a literal space, which
		remains part of the full span but is excluded from the core PII denominator. Dark boxes are included
		and light boxes are excluded. The right-hand strings concatenate included characters for illustration;
		evaluation uses their original offsets.
	\end{minipage}
\end{center}

\hypertarget{supp-table-annotation-labels}{%
	\subsubsection{Table \SuppTabAnnotationLabels{} \textbar{} Annotation labels and definitions}\label{supp-table-annotation-labels}}

\begingroup
\renewcommand{\arraystretch}{0.92}
\begin{longtable}[]{@{}>{\raggedright\arraybackslash}p{0.25\linewidth}>{\raggedright\arraybackslash}p{0.67\linewidth}@{}}
	\toprule
	\textbf{Annotation label}            & \textbf{Definition}                                                                                                                                               \\
	\midrule
	\endhead
	\texttt{Name:Patient}                & Given names, family names or initials referring to the patient whose record is being annotated.                                                                   \\
	\texttt{Name:Caregiver}              & Names or initials of clinicians and other healthcare or social-care workers involved in care. A directly adjacent professional title may be included in the span. \\
	\texttt{Name:Other}                  & Names or initials of relatives, friends, other patients and external people who are not care providers.                                                           \\
	\addlinespace
	\texttt{Address\_Location:Patient}   & Postal addresses and other locations officially linked to the patient, including home address and place of birth.                                                 \\
	\texttt{Address\_Location:Caregiver} & Addresses or locations linked to a caregiver or care institution when they are not part of the organisation name.                                                 \\
	\texttt{Address\_Location:Other}     & Addresses and locations linked to relatives, external bodies, events, accidents or travel.                                                                        \\
	\addlinespace
	\texttt{Organization:Healthcare}     & Named healthcare or social-care organisations, insurers, departments, units and institution-specific locations relevant to the patient.                           \\
	\texttt{Organization:Other}          & Named non-healthcare organisations such as employers, companies, schools, clubs and associations.                                                                 \\
	\addlinespace
	\texttt{Date}                        & Calendar dates or date fragments, including public holidays and weekdays when directly attached to a date.                                                        \\
	\texttt{Age\_Birthdate}              & A date of birth or an expressed age, including the accompanying age unit.                                                                                         \\
	\texttt{Profession}                  & Occupation, professional status, education, voluntary role or hobby of the patient or a relative; treating-caregiver specialisms are excluded.                    \\
	\texttt{Contactdetails}              & Communication details such as e-mail addresses, telephone or fax numbers, pagers and URLs; postal addresses are labelled as \texttt{Address\_Location}.           \\
	\texttt{ID:Patient}                  & Identifiers traceable to the patient, including patient or national-register numbers, record identifiers, study identifiers and patient-specific access links.    \\
	\texttt{ID:Caregiver}                & Identifiers assigned to a professional, such as a registration, licence or institutional staff number.                                                            \\
	\texttt{Anonymize\_Other}            & Exceptional identifying information that materially raises re-identification risk but does not fit another label.                                                 \\
	\bottomrule
\end{longtable}
\endgroup

{\footnotesize Concise definitions are shown here; the complete annotation rules and worked examples
	are provided in Appendix A.\par}

\hypertarget{supp-table-subannotation-categories}{%
	\subsubsection{Table \SuppTabSubannotations{} \textbar{} Sub-annotations included in core PII recall}\label{supp-table-subannotation-categories}}

\begingroup
\setlength{\defaultaddspace}{2pt}
\begin{longtable}[]{@{}>{\raggedright\arraybackslash}p{0.17\linewidth}>{\raggedright\arraybackslash}p{0.22\linewidth}>{\raggedright\arraybackslash}p{0.48\linewidth}c@{}}
	\toprule
	\textbf{Sub-annotation}     & \textbf{Parent annotation(s)}   & \textbf{Definition}                                                                                                                     & \textbf{Core PII?} \\
	\midrule
	\endhead
	\multicolumn{4}{@{}l}{\textit{Context and span structure}}                                                                                                                                                                   \\
	\addlinespace
	\texttt{formatting}         & Any                             & Non-identifying separators and syntax, including whitespace, punctuation, brackets and telephone-formatting or country-code characters. & No                 \\
	\texttt{additional\_info}   & Any                             & Non-identifying descriptive or contextual words captured within the annotation boundary.                                                & No                 \\
	\texttt{medical\_info}      & Any                             & Clinical information included inside a broader annotated span but not identifying on its own.                                           & No                 \\
	\addlinespace
	\multicolumn{4}{@{}l}{\textit{Person-name components}}                                                                                                                                                                       \\
	\addlinespace
	\texttt{given}              & Name; Contactdetails            & Given-name component of a person name or contact identifier.                                                                            & Yes                \\
	\texttt{family}             & Name; Contactdetails            & Family-name component of a person name or contact identifier.                                                                           & Yes                \\
	\texttt{initials}           & Name; Contactdetails            & Initials representing a person's given or family names.                                                                                 & Yes                \\
	\texttt{title}              & Name                            & Honorific or professional title adjacent to a name, such as ``Dr'' or ``Prof''.                                                         & No                 \\
	\addlinespace
	\multicolumn{4}{@{}l}{\textit{Professional and social context}}                                                                                                                                                              \\
	\addlinespace
	\texttt{hobby}              & Profession                      & Recreational, educational or voluntary activity used as a personal descriptor.                                                          & Yes                \\
	\texttt{profession}         & Profession                      & Occupation, job title, educational field or named professional role.                                                                    & Yes                \\
	\texttt{employment\_state}  & Profession                      & Employment-status descriptor such as retired, unemployed or self-employed.                                                              & Yes                \\
	\addlinespace
	\multicolumn{4}{@{}l}{\textit{Organisation components}}                                                                                                                                                                      \\
	\addlinespace
	\texttt{company}            & Organization                    & Named commercial company or employer.                                                                                                   & Yes                \\
	\texttt{institution}        & Organization; Contactdetails    & Named (healthcare) institution, including an institution-identifying component of an e-mail address or URL.                             & Yes                \\
	\texttt{hospital\_location} & Organization                    & Institution-specific campus, department, ward, unit or room/location code.                                                              & Yes                \\
	\addlinespace
	\multicolumn{4}{@{}l}{\textit{Address and geographic components}}                                                                                                                                                            \\
	\addlinespace
	\texttt{country}            & Address\_Location               & Country component when it occurs within an annotated address or location.                                                               & Yes                \\
	\texttt{province}           & Address\_Location; Organization & Province or equivalent first-level administrative area.                                                                                 & Yes                \\
	\texttt{region}             & Address\_Location; Organization & Named region or other subnational area distinct from a province.                                                                        & Yes                \\
	\texttt{municipality}       & Address\_Location; Organization & City, town, village or municipality.                                                                                                    & Yes                \\
	\texttt{postal\_code}       & Address\_Location               & Postal or ZIP code.                                                                                                                     & Yes                \\
	\texttt{street}             & Address\_Location               & Street or road name.                                                                                                                    & Yes                \\
	\texttt{house\_number}      & Address\_Location               & Building or house number.                                                                                                               & Yes                \\
	\texttt{bus\_number}        & Address\_Location               & Apartment, unit or Belgian ``bus'' number.                                                                                              & Yes                \\
	\texttt{postal\_office}     & Address\_Location               & Post-office, delivery-office or locality suffix forming part of an address.                                                             & Yes                \\
	\addlinespace
	\multicolumn{4}{@{}l}{\textit{Contact components}}                                                                                                                                                                           \\
	\addlinespace
	\texttt{internal\_phone}    & Contactdetails                  & Internal telephone extension or institution-only telephone number.                                                                      & Yes                \\
	\texttt{public\_phone}      & Contactdetails                  & Publicly dialable telephone or mobile number.                                                                                           & Yes                \\
	\texttt{fax\_number}        & Contactdetails                  & Fax number.                                                                                                                             & Yes                \\
	\addlinespace
	\multicolumn{4}{@{}l}{\textit{Identifier components}}                                                                                                                                                                        \\
	\addlinespace
	\texttt{public\_id}         & ID                              & Externally recognised person-specific identifier, such as a national-register or provider number.                                       & Yes                \\
	\texttt{internal\_id}       & ID                              & Locally assigned record, patient, admission, study or staff identifier.                                                                 & Yes                \\
	\addlinespace
	\multicolumn{4}{@{}l}{\textit{Calendar, time and age components}}                                                                                                                                                            \\
	\addlinespace
	\texttt{day}                & Date; Age\_Birthdate; ID        & Day component in a date, birth date or identifier.                                                                                      & Yes                \\
	\texttt{week}               & Date; Age\_Birthdate; ID        & Calendar-week component in a date, birth date or identifier.                                                                            & Yes                \\
	\texttt{month}              & Date; Age\_Birthdate; ID        & Month component in a date, birth date or identifier.                                                                                    & Yes                \\
	\texttt{year}               & Date; Age\_Birthdate; ID        & Year component in a date, birth date or identifier.                                                                                     & Yes                \\
	\texttt{weekday}            & Date; Age\_Birthdate            & Named weekday when included with a calendar date or date of birth.                                                                      & Yes                \\
	\texttt{time}               & Date; Age\_Birthdate            & Clock time occurring within a broader date or birth-date span.                                                                          & No                 \\
	\texttt{season}             & Date                            & Named season in a season--year expression used as an imprecise calendar reference.                                                      & Yes                \\
	\texttt{age\_type}          & Age\_Birthdate                  & Lexical age unit or marker, such as ``years'', ``months'' or ``y''.                                                                     & Yes                \\
	\texttt{age\_year}          & Age\_Birthdate                  & Age component expressed in years.                                                                                                       & Yes                \\
	\texttt{age\_month}         & Age\_Birthdate                  & Age component expressed in months.                                                                                                      & Yes                \\
	\texttt{age\_week}          & Age\_Birthdate                  & Age component expressed in weeks.                                                                                                       & Yes                \\
	\texttt{age\_day}           & Age\_Birthdate                  & Age component expressed in days.                                                                                                        & Yes                \\
	\bottomrule
\end{longtable}
\endgroup

{\footnotesize
	All sub-annotation categories are included in core PII recall unless marked ``No''. The five excluded
	categories are \texttt{formatting}, \texttt{additional\_info}, \texttt{medical\_info},
	\texttt{title} and \texttt{time}. Parent annotations indicate where each sub-annotation was observed
	in this study and are not exhaustive.\par}

\hypertarget{supp-section-dutch-benchmark-results}{%
	\subsection{S3. Full Dutch benchmark results}\label{supp-section-dutch-benchmark-results}}

Tables \SuppTabDutchPointRange{} report character-level performance and Table
\SuppTabDutchUncertainty{} reports uncertainty for the primary Dutch benchmark outcomes; all values
are percentages. Tables \SuppTabDutchPointRange{} give point estimates for the full metric
decomposition and both metadata configurations. Table \SuppTabDutchUncertainty{} gives intervals
for core PII recall and non-PII redaction in the metadata-enabled configuration used in the main
analysis. For the hospital and primary-care benchmarks, intervals are reported for every system.
The Dutch synthetic benchmark tests the complete workflow in-domain; the uncertainty analysis
therefore covered the headline \texttt{meddeid-dutch-synth} result. The narrower English tables
retain their intervals inline. Core PII recall is the percentage of identifying information hidden.
Its denominator excludes the five non-identifying or clinically meaningful sub-annotation
categories listed in Table \SuppTabSubannotations. Overall recall instead covers all characters in
the complete annotated PII spans, including those five categories. The non-PII redaction rate is
the percentage of characters outside annotated PII spans that are redacted. It is decomposed into
characters redacted by \emph{false-positive spans}, which do not overlap any annotated PII span,
and \emph{PII boundary extensions}, the excess characters redacted by predicted spans that overlap
an annotated PII span but extend beyond its boundary. These two components sum to the total non-PII
redaction rate. Metadata-enabled rows are the deployed, main-text configurations in which patient
and caregiver names from note metadata are injected into the detection layer; dashes indicate not
applicable. The non-PII denominator includes every evaluated document, including hard-negative
notes without gold PII.

\hypertarget{supp-table-dutch-hospital}{%
	\subsubsection{Table \SuppTabDutchHospital{} \textbar{} Hospital benchmark (300 notes)}\label{supp-table-dutch-hospital}}

\begingroup\scriptsize\setlength{\tabcolsep}{3pt}
\begin{longtable}[]{@{}p{0.245\linewidth}>{\centering\arraybackslash}p{0.065\linewidth}*{6}{>{\centering\arraybackslash}p{0.083\linewidth}}@{}}
	\toprule
	Method                        & Meta. & Core PII recall (\%) & Overall recall (\%) & Non-PII redaction rate (\%) & False positives (\%) & Boundary extensions (\%)\tabularnewline
	\midrule
	\endhead
	meddeid-dutch-uza (ours)      & no    & 98.8                 & 98.4                & 0.226                       & 0.067                & 0.159\tabularnewline
	meddeid-dutch-uza (ours)      & yes   & 98.9                 & 98.5                & 0.236                       & 0.067                & 0.168\tabularnewline
	meddeid-dutch-synth (ours)    & no    & 95.3                 & 94.5                & 0.815                       & 0.630                & 0.185\tabularnewline
	meddeid-dutch-synth (ours)    & yes   & 96.1                 & 95.2                & 0.825                       & 0.630                & 0.195\tabularnewline
	Belgian DEDUCE (ours)         & no    & 78.5                 & 77.5                & 0.594                       & 0.401                & 0.193\tabularnewline
	Belgian DEDUCE (ours)         & yes   & 88.0                 & 85.7                & 0.634                       & 0.423                & 0.211\tabularnewline
	Qwen3-8B (Yang et al.)        & no    & 81.8                 & 81.4                & 0.872                       & 0.451                & 0.420\tabularnewline
	Qwen3-8B (Yang et al.)        & yes   & 84.2                 & 83.6                & 0.877                       & 0.451                & 0.426\tabularnewline
	deidentify (Trienes et al.)   & no    & 82.8                 & 74.6                & 0.546                       & 0.539                & 0.007\tabularnewline
	deidentify (Trienes et al.)   & yes   & 86.4                 & 78.8                & 0.558                       & 0.539                & 0.019\tabularnewline
	GLiNER-PII (Zaratiana et al.) & no    & 75.0                 & 75.4                & 2.108                       & 1.796                & 0.313\tabularnewline
	GLiNER-PII (Zaratiana et al.) & yes   & 76.6                 & 76.8                & 2.113                       & 1.796                & 0.317\tabularnewline
	DEDUCE (Menger et al.)        & no    & 65.5                 & 65.8                & 0.465                       & 0.317                & 0.148\tabularnewline
	DEDUCE (Menger et al.)        & yes   & 75.9                 & 74.9                & 0.502                       & 0.332                & 0.169\tabularnewline
	OpenAI privacy filter         & no    & 66.2                 & 63.3                & 0.607                       & 0.527                & 0.080\tabularnewline
	OpenAI privacy filter         & yes   & 72.1                 & 68.5                & 0.611                       & 0.527                & 0.084\tabularnewline
	OpenMed multilingual filter   & no    & 37.6                 & 33.5                & 0.946                       & 0.927                & 0.019\tabularnewline
	OpenMed multilingual filter   & yes   & 49.1                 & 44.1                & 0.955                       & 0.927                & 0.028\tabularnewline
	Annotator 1                   & --    & 98.8                 & 98.5                & 0.174                       & 0.007                & 0.167\tabularnewline
	Annotator 2                   & --    & 98.5                 & 98.5                & 0.359                       & 0.123                & 0.236\tabularnewline
	\bottomrule
\end{longtable}
\endgroup

{\footnotesize Values are percentages. Meta., metadata injection; non-PII redaction is the sum of
	false positives and boundary extensions. Annotator rows are descriptive.\par}

\hypertarget{supp-table-dutch-synthetic}{%
	% Keep this title with the table; otherwise it is stranded at the foot of
	% the preceding page while the column headings begin on the next page.
	\pagebreak
	\subsubsection{Table \SuppTabDutchSynthetic{} \textbar{} Synthetic benchmark (300 notes, openly released)}\label{supp-table-dutch-synthetic}}

\begingroup\scriptsize\setlength{\tabcolsep}{3pt}
\begin{longtable}[]{@{}p{0.245\linewidth}>{\centering\arraybackslash}p{0.065\linewidth}*{6}{>{\centering\arraybackslash}p{0.083\linewidth}}@{}}
	\toprule
	Method                        & Meta. & Core PII recall (\%) & Overall recall (\%) & Non-PII redaction rate (\%) & False positives (\%) & Boundary extensions (\%)\tabularnewline
	\midrule
	\endhead
	meddeid-dutch-uza (ours)      & no    & 96.2                 & 95.4                & 0.395                       & 0.184                & 0.210\tabularnewline
	meddeid-dutch-uza (ours)      & yes   & 96.6                 & 95.8                & 0.395                       & 0.184                & 0.210\tabularnewline
	meddeid-dutch-synth (ours)    & no    & 99.8                 & 99.7                & 0.279                       & 0.046                & 0.233\tabularnewline
	meddeid-dutch-synth (ours)    & yes   & 99.8                 & 99.7                & 0.279                       & 0.046                & 0.233\tabularnewline
	Belgian DEDUCE (ours)         & no    & 65.7                 & 65.4                & 0.791                       & 0.309                & 0.482\tabularnewline
	Belgian DEDUCE (ours)         & yes   & 70.2                 & 69.6                & 1.696                       & 1.089                & 0.607\tabularnewline
	Qwen3-8B (Yang et al.)        & no    & 90.4                 & 90.4                & 1.284                       & 0.495                & 0.790\tabularnewline
	Qwen3-8B (Yang et al.)        & yes   & 90.9                 & 90.9                & 1.284                       & 0.495                & 0.790\tabularnewline
	deidentify (Trienes et al.)   & no    & 74.9                 & 71.8                & 0.371                       & 0.306                & 0.065\tabularnewline
	deidentify (Trienes et al.)   & yes   & 76.7                 & 74.5                & 0.371                       & 0.306                & 0.065\tabularnewline
	GLiNER-PII (Zaratiana et al.) & no    & 82.2                 & 81.9                & 1.580                       & 1.024                & 0.555\tabularnewline
	GLiNER-PII (Zaratiana et al.) & yes   & 83.0                 & 83.1                & 1.580                       & 1.024                & 0.555\tabularnewline
	DEDUCE (Menger et al.)        & no    & 35.6                 & 36.3                & 0.405                       & 0.273                & 0.132\tabularnewline
	DEDUCE (Menger et al.)        & yes   & 45.9                 & 46.1                & 0.576                       & 0.408                & 0.168\tabularnewline
	OpenAI privacy filter         & no    & 53.3                 & 53.3                & 0.517                       & 0.389                & 0.128\tabularnewline
	OpenAI privacy filter         & yes   & 58.9                 & 58.9                & 0.517                       & 0.389                & 0.128\tabularnewline
	OpenMed multilingual filter   & no    & 48.8                 & 46.3                & 1.012                       & 0.964                & 0.048\tabularnewline
	OpenMed multilingual filter   & yes   & 53.0                 & 50.6                & 1.012                       & 0.964                & 0.048\tabularnewline
	\bottomrule
\end{longtable}
\endgroup

{\footnotesize Values are percentages. Meta., metadata injection; non-PII redaction is the sum of
	false positives and boundary extensions.\par}

\hypertarget{supp-table-dutch-primary-care}{%
	\subsubsection{Table \SuppTabDutchPrimaryCare{} \textbar{} Primary-care external validation (100 notes)}\label{supp-table-dutch-primary-care}}

\begingroup\scriptsize\setlength{\tabcolsep}{3pt}
\begin{longtable}[]{@{}p{0.245\linewidth}>{\centering\arraybackslash}p{0.065\linewidth}*{6}{>{\centering\arraybackslash}p{0.083\linewidth}}@{}}
	\toprule
	Method                        & Meta. & Core PII recall (\%) & Overall recall (\%) & Non-PII redaction rate (\%) & False positives (\%) & Boundary extensions (\%)\tabularnewline
	\midrule
	\endhead
	meddeid-dutch-uza (ours)      & no    & 86.7                 & 84.8                & 0.285                       & 0.225                & 0.060\tabularnewline
	meddeid-dutch-uza (ours)      & yes   & 87.0                 & 85.0                & 0.288                       & 0.227                & 0.061\tabularnewline
	meddeid-dutch-synth (ours)    & no    & 89.9                 & 87.0                & 1.219                       & 1.144                & 0.075\tabularnewline
	meddeid-dutch-synth (ours)    & yes   & 90.3                 & 87.3                & 1.224                       & 1.149                & 0.075\tabularnewline
	Belgian DEDUCE (ours)         & no    & 73.0                 & 69.3                & 0.881                       & 0.829                & 0.052\tabularnewline
	Belgian DEDUCE (ours)         & yes   & 74.9                 & 71.2                & 0.905                       & 0.855                & 0.050\tabularnewline
	Qwen3-8B (Yang et al.)        & no    & 58.4                 & 58.0                & 0.306                       & 0.151                & 0.155\tabularnewline
	Qwen3-8B (Yang et al.)        & yes   & 60.8                 & 60.0                & 0.312                       & 0.156                & 0.156\tabularnewline
	deidentify (Trienes et al.)   & no    & 67.9                 & 62.4                & 5.727                       & 5.650                & 0.077\tabularnewline
	deidentify (Trienes et al.)   & yes   & 68.6                 & 62.9                & 5.728                       & 5.651                & 0.077\tabularnewline
	GLiNER-PII (Zaratiana et al.) & no    & 70.1                 & 69.1                & 3.423                       & 3.327                & 0.095\tabularnewline
	GLiNER-PII (Zaratiana et al.) & yes   & 70.4                 & 69.4                & 3.424                       & 3.328                & 0.095\tabularnewline
	DEDUCE (Menger et al.)        & no    & 54.4                 & 52.4                & 0.522                       & 0.478                & 0.044\tabularnewline
	DEDUCE (Menger et al.)        & yes   & 57.2                 & 54.9                & 0.543                       & 0.496                & 0.047\tabularnewline
	OpenAI privacy filter         & no    & 64.5                 & 61.9                & 11.461                      & 11.296               & 0.165\tabularnewline
	OpenAI privacy filter         & yes   & 66.9                 & 63.9                & 11.464                      & 11.299               & 0.165\tabularnewline
	OpenMed multilingual filter   & no    & 34.1                 & 31.4                & 8.244                       & 8.236                & 0.008\tabularnewline
	OpenMed multilingual filter   & yes   & 37.1                 & 33.9                & 8.247                       & 8.238                & 0.009\tabularnewline
	\bottomrule
\end{longtable}
\endgroup

{\footnotesize Values are percentages. Meta., metadata injection; non-PII redaction is the sum of
	false positives and boundary extensions.\par}

\hypertarget{supp-table-dutch-uncertainty}{%
	\subsubsection{Table \SuppTabDutchUncertainty{} \textbar{} Confidence intervals for Dutch benchmark outcomes}\label{supp-table-dutch-uncertainty}}

\begingroup\scriptsize\setlength{\tabcolsep}{4pt}
\begin{longtable}[]{@{}p{0.15\linewidth}p{0.30\linewidth}>{\centering\arraybackslash}p{0.23\linewidth}>{\centering\arraybackslash}p{0.23\linewidth}@{}}
	\toprule
	Benchmark    & Method                        & Core PII recall, \% (95\% CI) & Non-PII redaction, \% (95\% CI)\tabularnewline
	\midrule
	\endhead
	Hospital     & meddeid-dutch-uza (ours)      & 98.9 (98.5--99.3)             & 0.24 (0.16--0.32)\tabularnewline
	Hospital     & meddeid-dutch-synth (ours)    & 96.1 (95.3--96.9)             & 0.83 (0.64--1.03)\tabularnewline
	Hospital     & Belgian DEDUCE (ours)         & 88.0 (86.6--89.4)             & 0.63 (0.46--0.84)\tabularnewline
	Hospital     & Qwen3-8B (Yang et al.)        & 84.2 (82.0--86.4)             & 0.88 (0.71--1.09)\tabularnewline
	Hospital     & deidentify (Trienes et al.)   & 86.4 (85.0--87.7)             & 0.56 (0.37--0.80)\tabularnewline
	Hospital     & GLiNER-PII (Zaratiana et al.) & 76.6 (74.8--78.4)             & 2.11 (1.87--2.38)\tabularnewline
	Hospital     & DEDUCE (Menger et al.)        & 75.9 (74.0--77.9)             & 0.50 (0.34--0.70)\tabularnewline
	Hospital     & OpenAI privacy filter         & 72.1 (69.5--74.6)             & 0.61 (0.41--0.87)\tabularnewline
	Hospital     & OpenMed multilingual filter   & 49.1 (47.2--51.0)             & 0.96 (0.85--1.06)\tabularnewline
	\addlinespace[3pt]
	Synthetic    & meddeid-dutch-synth (ours)    & 99.8 (99.5--100.0)            & 0.28 (0.21--0.35)\tabularnewline
	\addlinespace[3pt]
	Primary care & meddeid-dutch-uza (ours)      & 87.0 (83.9--90.0)             & 0.29 (0.23--0.38)\tabularnewline
	Primary care & meddeid-dutch-synth (ours)    & 90.3 (88.4--92.3)             & 1.22 (0.88--1.71)\tabularnewline
	Primary care & Belgian DEDUCE (ours)         & 74.9 (71.4--78.4)             & 0.90 (0.67--1.20)\tabularnewline
	Primary care & Qwen3-8B (Yang et al.)        & 60.8 (56.1--65.5)             & 0.31 (0.18--0.51)\tabularnewline
	Primary care & deidentify (Trienes et al.)   & 68.6 (62.6--74.1)             & 5.73 (2.13--9.54)\tabularnewline
	Primary care & GLiNER-PII (Zaratiana et al.) & 70.4 (67.2--73.3)             & 3.42 (2.57--4.37)\tabularnewline
	Primary care & DEDUCE (Menger et al.)        & 57.2 (51.7--62.6)             & 0.54 (0.40--0.67)\tabularnewline
	Primary care & OpenAI privacy filter         & 66.9 (62.7--71.2)             & 11.46 (3.03--20.33)\tabularnewline
	Primary care & OpenMed multilingual filter   & 37.1 (34.5--39.8)             & 8.25 (3.64--13.22)\tabularnewline
	\bottomrule
\end{longtable}
\endgroup

{\footnotesize
	Values are point estimates with 95\% percentile-bootstrap confidence intervals in parentheses.
	Complete documents were sampled with replacement in 10,000 replicates, using the same document
	multiplicities for every system within each analysis; rates were recalculated from summed character
	counts. Metadata-enabled configurations are shown. Hospital and primary-care analyses include every
	system; the Dutch synthetic analysis includes the headline \texttt{meddeid-dutch-synth} result only.
	The annotator comparison is descriptive.\par}

\hypertarget{supp-section-per-label-performance}{%
	\subsection{S4. Per-label performance}\label{supp-section-per-label-performance}}

\hypertarget{supp-table-label-recall}{%
	\subsubsection{Table \SuppTabLabelRecall{} \textbar{} Core PII recall by gold label in hospital notes}\label{supp-table-label-recall}}

\begingroup\fontsize{6.5}{7.3}\selectfont\setlength{\tabcolsep}{3.5pt}
\begin{longtable}[]{@{}p{0.232\linewidth}>{\centering\arraybackslash}p{0.060\linewidth}*{7}{>{\centering\arraybackslash}p{0.066\linewidth}}@{}}
	\toprule
	Gold label                  & Spans & UZA   & Synth. & Bel. DEDUCE & Qwen3-8B & OpenAI & Ann. 1 & Ann. 2\tabularnewline
	\midrule
	\endhead
	Date                        & 1463  & 99.9  & 99.7   & 87.5        & 77.8     & 82.5   & 99.3   & 98.8\tabularnewline
	Name:Caregiver              & 1023  & 99.4  & 95.3   & 92.9        & 87.0     & 84.7   & 99.2   & 99.4\tabularnewline
	Name:Patient                & 337   & 100.0 & 98.2   & 96.8        & 98.5     & 97.4   & 100.0  & 100.0\tabularnewline
	Age\_Birthdate              & 335   & 99.4  & 98.6   & 87.3        & 89.8     & 77.7   & 99.1   & 96.7\tabularnewline
	ID:Patient                  & 314   & 96.8  & 94.1   & 79.4        & 78.0     & 67.9   & 98.8   & 98.9\tabularnewline
	Organization:Healthcare     & 296   & 94.7  & 77.0   & 67.3        & 62.3     & 12.6   & 93.5   & 96.4\tabularnewline
	Contactdetails              & 132   & 99.8  & 100.0  & 75.6        & 92.8     & 46.4   & 100.0  & 99.7\tabularnewline
	Address\_Location:Patient   & 127   & 100.0 & 99.7   & 94.2        & 89.4     & 36.0   & 100.0  & 100.0\tabularnewline
	ID:Caregiver                & 107   & 100.0 & 97.5   & 73.8        & 87.4     & 76.7   & 99.3   & 97.0\tabularnewline
	Address\_Location:Caregiver & 81    & 100.0 & 99.3   & 97.2        & 90.7     & 45.1   & 99.8   & 99.8\tabularnewline
	Profession                  & 31    & 72.7  & 47.8   & 0.0         & 28.8     & 0.0    & 73.8   & 47.8\tabularnewline
	Name:Other                  & 12    & 100.0 & 90.3   & 100.0       & 91.0     & 73.1   & 91.0   & 100.0\tabularnewline
	Organization:Other          & 7     & 21.4  & 66.1   & 7.1         & 17.9     & 0.0    & 71.4   & 94.6\tabularnewline
	Address\_Location:Other     & 4     & 100.0 & 82.6   & 100.0       & 100.0    & 69.6   & 100.0  & 100.0\tabularnewline
	\bottomrule
\end{longtable}
\endgroup

{\footnotesize Values are percentages from metadata-enabled configurations. Gold span counts are
	shown because several categories are small and should not be over-interpreted. Abbreviations: UZA,
	\texttt{meddeid-dutch-uza}; Synth., \texttt{meddeid-dutch-synth}; Bel. DEDUCE, Belgian DEDUCE;
	Ann., annotator.\par}

For gold-label categories containing more than 100 spans, \texttt{meddeid-dutch-uza} achieved
recall of at least 96.0\%, except for Organization:Healthcare (94.7\%; 296 spans). Recall was
100.0\% for patient names, caregiver identifiers, and patient and caregiver addresses. Lower recall
was observed for Profession (72.7\%; 31 spans) and Organization:Other (21.4\%; 7 spans). These
estimates are based on small numbers of spans and should be interpreted cautiously. The two
annotators also differed for these small categories. For Profession, they identified 73.8\% and
47.8\% of the annotated text; for Organization:Other, they identified 71.4\% and 94.6\%.

The general-purpose OpenAI neural PII detector achieved 12.6\% recall for healthcare organisations
and 0.0\% for professions; Belgian DEDUCE also achieved 0.0\% for professions. These categories are
not represented in the corresponding general-purpose or rule-based detector taxonomies.

\begin{figure}[!htbp]
	\centering
	\includegraphics[width=\linewidth]{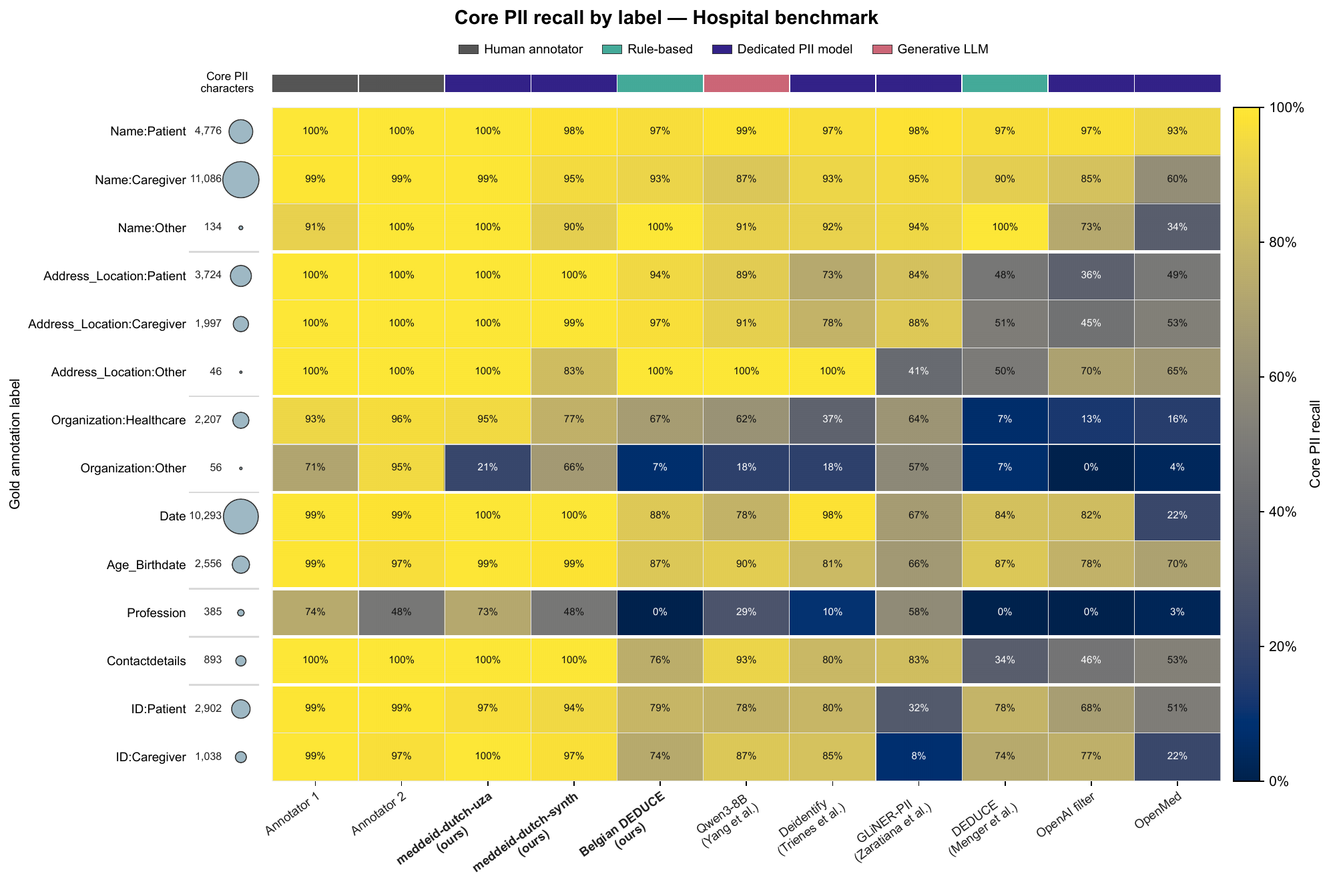}
	\caption*{\textbf{Fig. \SuppFigLabelRecall{} \textbar{} Recall by gold label, hospital benchmark.}
		Core PII recall for each system and gold-label category. Cell values are percentages;
		circle area at left represents the character-level core PII denominator, with the exact
		character count shown beside it. The strongest systems perform consistently on common names, dates
		and identifiers, but differences widen for rarer or clinically specific categories. Overall recall
		therefore conceals category-specific gaps. Rows with few gold spans should be interpreted cautiously.}
\end{figure}

\begin{figure}[!htbp]
	\centering
	\includegraphics[height=0.69\textheight,keepaspectratio]{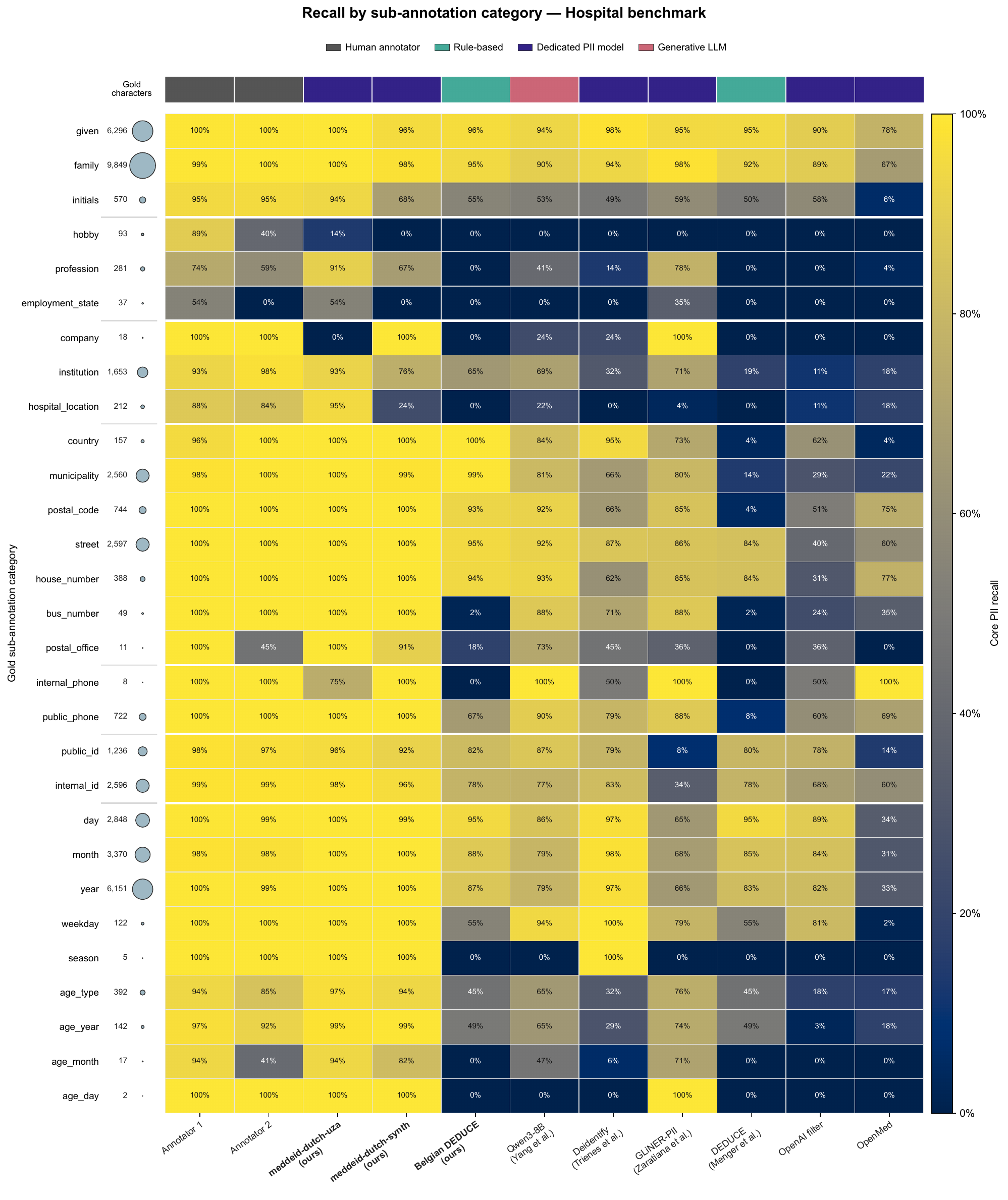}
	\caption*{\textbf{Fig. \SuppFigSubannotationRecall{} \textbar{} Recall by sub-annotation category, hospital benchmark.}
		Character recall by gold sub-annotation category. Cell values are percentages. The two leftmost
		columns show the physician annotators. Circle area represents the number of gold sub-annotation
		characters, with the exact count printed beside each category.}
\end{figure}

\clearpage
\hypertarget{s5-label-fidelity}{%
	\subsection{S5. Span detection and label fidelity}\label{s5-label-fidelity}}

Core PII recall measures the fraction of core identifying characters redacted, irrespective of the
predicted label. To analyse span-level detection and label assignment separately, predictions and
annotations were paired one-to-one when their core-PII character positions overlapped. Label
fidelity is the proportion of detected spans assigned the correct identifier type and role. The
confusion matrices include only detected spans; missed annotations are reflected in the span
detection-recall values in Table \SuppTabLabelFidelity. Only gold spans containing at least one
core-PII character entered this analysis. Table \SuppTabLabelFidelity{} therefore includes 4,263 of
the 4,269 hospital annotation records and 4,008 of the 4,010 primary-care records; the remaining
six and two records, respectively, contained only sub-annotations excluded from core PII recall.

\subsubsection{Table \SuppTabLabelFidelity{} \textbar{} Span detection and label fidelity}\label{supp-table-label-fidelity}
\begingroup\fontsize{7.8}{9.2}\selectfont\setlength{\tabcolsep}{4pt}
\begin{tabular}{@{}>{\raggedright\arraybackslash}p{0.13\linewidth}>{\raggedright\arraybackslash}p{0.24\linewidth}>{\raggedright\arraybackslash}p{0.17\linewidth}>{\raggedright\arraybackslash}p{0.20\linewidth}>{\raggedright\arraybackslash}p{0.10\linewidth}@{}}
	\toprule
	Benchmark    & Model                        & Detected / annotated (span detection recall) & Correct label / detected (label accuracy) & Incorrectly labelled\tabularnewline
	\midrule
	Hospital     & \texttt{meddeid-dutch-uza}   & 4,188 / 4,263 (98.2\%)                       & 4,143 / 4,188 (98.9\%)                    & 45\tabularnewline
	Hospital     & \texttt{meddeid-dutch-synth} & 3,956 / 4,263 (92.8\%)                       & 3,663 / 3,956 (92.6\%)                    & 293\tabularnewline
	Primary care & \texttt{meddeid-dutch-uza}   & 3,102 / 4,008 (77.4\%)                       & 2,787 / 3,102 (89.8\%)                    & 315\tabularnewline
	Primary care & \texttt{meddeid-dutch-synth} & 3,218 / 4,008 (80.3\%)                       & 2,552 / 3,218 (79.3\%)                    & 666\tabularnewline
	\bottomrule
\end{tabular}
\endgroup

{\footnotesize All values use the metadata-enabled configuration and the matching current evaluator
	export for each benchmark.\par}

\begin{figure}[!htbp]
	\centering
	\includegraphics[width=0.96\linewidth,height=0.48\textheight,keepaspectratio]{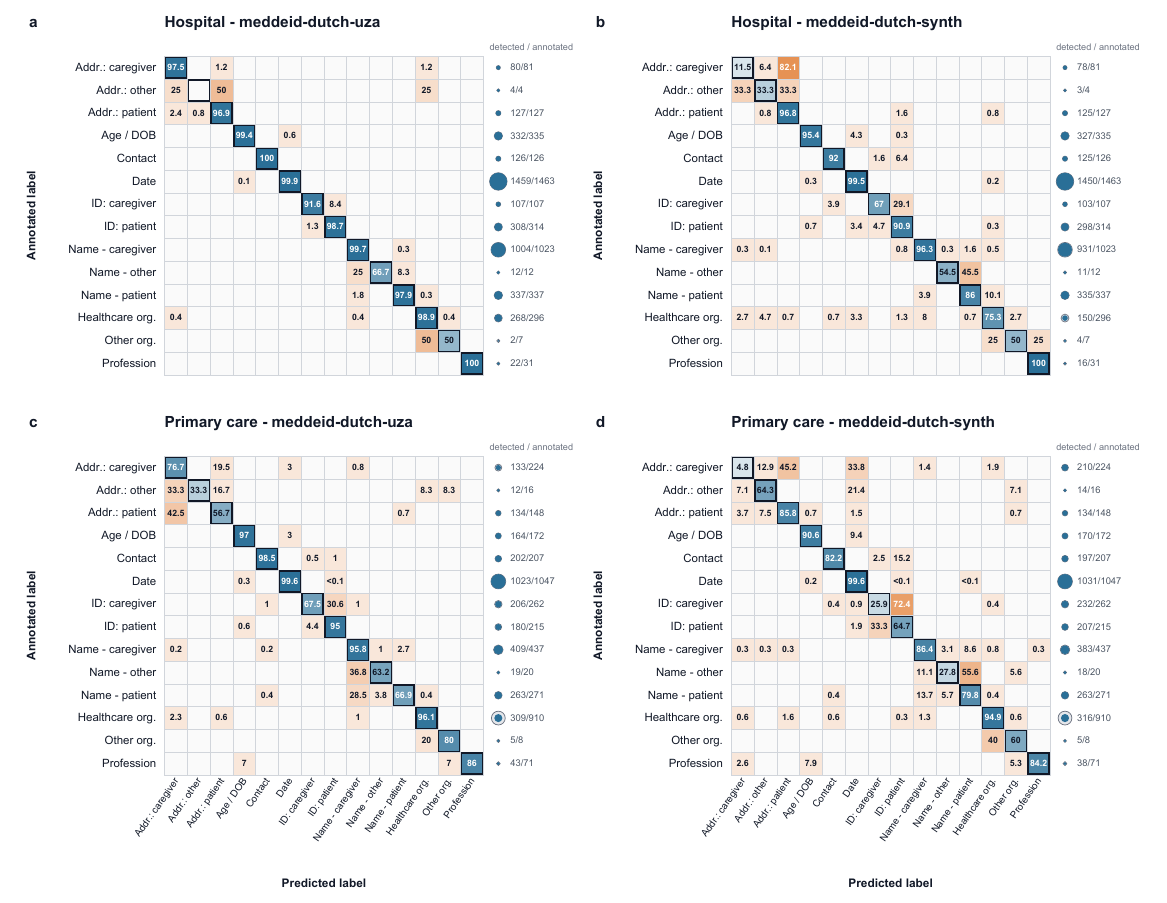}
	\caption*{\textbf{Fig. \SuppFigLabelConfusion{} \textbar{} Exact-label confusion among detected PII spans.}
		Columns compare the hospital-trained \texttt{meddeid-dutch-uza} and synthetic-trained
		\texttt{meddeid-dutch-synth}. \textbf{a,b}, Hospital benchmark; \textbf{c,d}, primary-care
		benchmark. Rows are reference labels and columns are predicted labels. Percentages are normalised
		within detected spans; blue diagonal cells indicate exact-label assignment and orange off-diagonal
		cells indicate misclassification. Circle area shows the number of annotated spans and blue fill the
		detected fraction; exact detected / annotated counts are printed alongside.}
\end{figure}

\clearpage
\hypertarget{supp-section-stability}{%
	\subsection{S6. Stability under input perturbations}\label{s6-stability}}

Coverage-selected subsets of up to 100 notes per scope were perturbed along name source, name
capitalisation, name format, date format, date value and age format. Asterisks below indicate
statistically significant degradation; interval construction and hypothesis testing are described
in Methods.

\textbf{Sign convention.} Tables~\SuppTabStabilityAggregateRange{} report recall loss as baseline minus perturbed recall:
positive values therefore indicate worse performance. \textit{Losses only} and \textit{worst drop}
are non-negative loss magnitudes. Table~\SuppTabStabilityCells{} instead reports recall change as perturbed minus
baseline recall: negative values indicate worse performance and positive values indicate an
improvement.

\subsubsection{Table \SuppTabStabilitySamples{} \textbar{} Perturbation-analysis sample sizes}\label{supp-table-stability-samples}
\begingroup\fontsize{8.4}{9.8}\selectfont\setlength{\tabcolsep}{3.5pt}
\begin{longtable}[]{@{}>{\raggedright\arraybackslash}p{0.13\linewidth}>{\raggedright\arraybackslash}p{0.20\linewidth}>{\raggedright\arraybackslash}p{0.33\linewidth}>{\raggedright\arraybackslash}p{0.25\linewidth}@{}}
	\toprule
	Test set     & Notes selected for testing ($n$) & Baseline--perturbation span pairs per analysis ($n$, range) & Notes represented per analysis ($n$, range)\tabularnewline
	\midrule
	\endhead
	Hospital     & 100                              & 27--519                                                     & 19--98\tabularnewline
	Synthetic    & 100                              & 30--508                                                     & 23--97\tabularnewline
	Primary care & 100                              & 87--1127                                                    & 28--91\tabularnewline
	\bottomrule
\end{longtable}
\endgroup

{\footnotesize For each test set, 100 notes were selected to cover the six perturbation dimensions.
	An analysis included only notes containing the relevant target identifier. Ranges give the smallest
	and largest sample sizes across the nine analyses listed for that test set in Table
	\SuppTabStabilityCells. A span pair is the same identifier in the original and perturbed note.\par}

\subsubsection{Table \SuppTabStabilityAggregate{} \textbar{} Aggregate stability across test sets}\label{supp-table-stability-aggregate}
\begin{longtable}[]{@{}llllll@{}}
	\toprule
	Model                      & Baseline recall (\%) & Net loss (pp) & Losses only (pp) & Worst drop (pp) & Significant cells\tabularnewline
	\midrule
	\endhead
	meddeid-dutch-uza (ours)   & 97.8                 & 2.87          & 2.96             & 19.23           & 7 / 27\tabularnewline
	meddeid-dutch-synth (ours) & 96.9                 & 0.60          & 1.25             & 4.56            & 5 / 27\tabularnewline
	\bottomrule
\end{longtable}

	{\footnotesize Each model contributes 27 cells. Positive loss denotes worse performance.
		\textit{Net loss} is the signed, target-span-weighted mean of baseline minus perturbed recall;
		\textit{losses only} uses the same weighting but replaces gains with zero. Metrics are calculated
		within each test set and then averaged equally across the three test sets. Significant cells have
		Benjamini--Hochberg-adjusted $q<0.05$. For \texttt{meddeid-dutch-synth}, perturbations caused 1.25 pp
		of average loss, while gains in other analyses reduced the net loss to 0.60 pp.\par}

\subsubsection{Table \SuppTabStabilityByScope{} \textbar{} Aggregate stability by test set}\label{supp-table-stability-by-scope}
\begin{longtable}[]{@{}llllll@{}}
	\toprule
	Scope        & Model                      & Baseline (\%) & Net loss (pp) & Worst drop (pp) & Significant cells\tabularnewline
	\midrule
	\endhead
	Hospital     & meddeid-dutch-uza (ours)   & 99.1          & 2.66          & 7.53            & 2 / 9\tabularnewline
	Hospital     & meddeid-dutch-synth (ours) & 95.2          & 0.45          & 4.02            & 2 / 9\tabularnewline
	Synthetic    & meddeid-dutch-uza (ours)   & 98.2          & 3.03          & 19.23           & 4 / 9\tabularnewline
	Synthetic    & meddeid-dutch-synth (ours) & 99.9          & 0.32          & 2.86            & 1 / 9\tabularnewline
	Primary care & meddeid-dutch-uza (ours)   & 96.3          & 2.93          & 7.77            & 1 / 9\tabularnewline
	Primary care & meddeid-dutch-synth (ours) & 95.5          & 1.03          & 4.56            & 2 / 9\tabularnewline
	\bottomrule
\end{longtable}

	{\footnotesize Positive loss denotes worse performance. Each test set contributes nine cells per
		model. Significant cells have Benjamini--Hochberg-adjusted $q<0.05$.\par}

\subsubsection{Table \SuppTabStabilityCells{} \textbar{} Cell-level recall changes under perturbation}\label{supp-table-stability-cells}
\begingroup\scriptsize\setlength{\tabcolsep}{3pt}
\begin{longtable}[]{@{}llllrll@{}}
	\toprule
	Scope                 & Dimension                   & Role               & $n$ pairs & $n$ notes & \shortstack{Hospital-trained model\\$\Delta$ recall, pp ($q$)} & \shortstack{Synthetic-trained model\\$\Delta$ recall, pp ($q$)}\tabularnewline
	\midrule
	\endhead
	Hospital              & age\_format                 & age                & 27        & 19        & -7.53 (0.126)                                                  & -3.70 (0.982)\tabularnewline
	\textbf{Hospital}     & \textbf{capitalization}     & \textbf{caregiver} & 164       & 66        & -1.52 (0.086)                                                  & \textbf{-3.81* (0.044)}\tabularnewline
	Hospital              & capitalization              & patient            & 40        & 32        & -3.75 (0.342)                                                  & +4.17 (1.000)\tabularnewline
	Hospital              & date\_format                & date               & 519       & 98        & -0.66 (0.099)                                                  & +0.25 (1.000)\tabularnewline
	\textbf{Hospital}     & \textbf{date\_value\_shift} & \textbf{date}      & 411       & 96        & \textbf{-4.83* (0.001)}                                        & \textbf{-4.02* (0.001)}\tabularnewline
	\textbf{Hospital}     & \textbf{format}             & \textbf{caregiver} & 161       & 66        & \textbf{-6.83* (0.033)}                                        & +2.91 (1.000)\tabularnewline
	Hospital              & format                      & patient            & 40        & 32        & -2.50 (0.126)                                                  & +14.50 (1.000)\tabularnewline
	Hospital              & name\_source                & caregiver          & 164       & 66        & +0.40 (0.735)                                                  & +1.34 (1.000)\tabularnewline
	Hospital              & name\_source                & patient            & 40        & 32        & -2.30 (0.086)                                                  & +2.80 (1.000)\tabularnewline
	Synthetic             & age\_format                 & age                & 87        & 84        & -1.51 (0.333)                                                  & -0.57 (0.982)\tabularnewline
	Synthetic             & capitalization              & caregiver          & 32        & 25        & -8.33 (0.086)                                                  & -2.86 (0.839)\tabularnewline
	\textbf{Synthetic}    & \textbf{capitalization}     & \textbf{patient}   & 91        & 89        & \textbf{-6.04* (0.008)}                                        & -0.55 (0.982)\tabularnewline
	\textbf{Synthetic}    & \textbf{date\_format}       & \textbf{date}      & 508       & 97        & -0.46 (0.099)                                                  & \textbf{-0.50* (0.001)}\tabularnewline
	\textbf{Synthetic}    & \textbf{date\_value\_shift} & \textbf{date}      & 357       & 86        & \textbf{-2.10* (0.001)}                                        & -0.16 (0.063)\tabularnewline
	\textbf{Synthetic}    & \textbf{format}             & \textbf{caregiver} & 30        & 23        & \textbf{-9.00* (0.040)}                                        & +3.33 (1.000)\tabularnewline
	\textbf{Synthetic}    & \textbf{format}             & \textbf{patient}   & 91        & 89        & \textbf{-19.23* (0.001)}                                       & -0.00 (1.000)\tabularnewline
	Synthetic             & name\_source                & caregiver          & 32        & 25        & -1.01 (0.413)                                                  & -0.50 (0.982)\tabularnewline
	Synthetic             & name\_source                & patient            & 91        & 89        & -0.09 (0.550)                                                  & -0.00 (1.000)\tabularnewline
	Primary care          & age\_format                 & age                & 87        & 28        & -4.94 (0.333)                                                  & -0.56 (0.982)\tabularnewline
	Primary care          & capitalization              & caregiver          & 236       & 53        & -2.68 (0.213)                                                  & -4.56 (0.487)\tabularnewline
	Primary care          & capitalization              & patient            & 163       & 43        & -1.02 (0.333)                                                  & +2.51 (1.000)\tabularnewline
	\textbf{Primary care} & \textbf{date\_format}       & \textbf{date}      & 1127      & 91        & +0.64 (0.785)                                                  & \textbf{-0.96* (0.008)}\tabularnewline
	\textbf{Primary care} & \textbf{date\_value\_shift} & \textbf{date}      & 984       & 87        & \textbf{-7.77* (0.001)}                                        & \textbf{-4.06* (0.001)}\tabularnewline
	Primary care          & format                      & caregiver          & 229       & 53        & -5.66 (0.291)                                                  & +2.42 (1.000)\tabularnewline
	Primary care          & format                      & patient            & 163       & 43        & +0.03 (0.550)                                                  & +7.61 (1.000)\tabularnewline
	Primary care          & name\_source                & caregiver          & 236       & 53        & -0.43 (0.413)                                                  & +2.16 (1.000)\tabularnewline
	Primary care          & name\_source                & patient            & 163       & 43        & -2.28 (0.414)                                                  & -0.05 (0.982)\tabularnewline
	\bottomrule
\end{longtable}
\endgroup

{\footnotesize Recall change is perturbed minus baseline recall; negative values denote worse
	performance. Parentheses contain Benjamini--Hochberg-adjusted $q$ values; asterisks denote $q<0.05$.
	Rows containing at least one significant degradation are bold.\par}

Substituting one set of names for another did not significantly reduce recall after
multiple-testing correction. The main vulnerabilities were changes to name formatting and shifted
date values, especially for the hospital-trained model. The synthetic-trained model was more stable
overall, although it was not better under every individual perturbation.

\begin{figure}[!htbp]
	\centering
	\begin{minipage}[t]{0.49\linewidth}\textbf{a}\par
		\includegraphics[width=\linewidth]{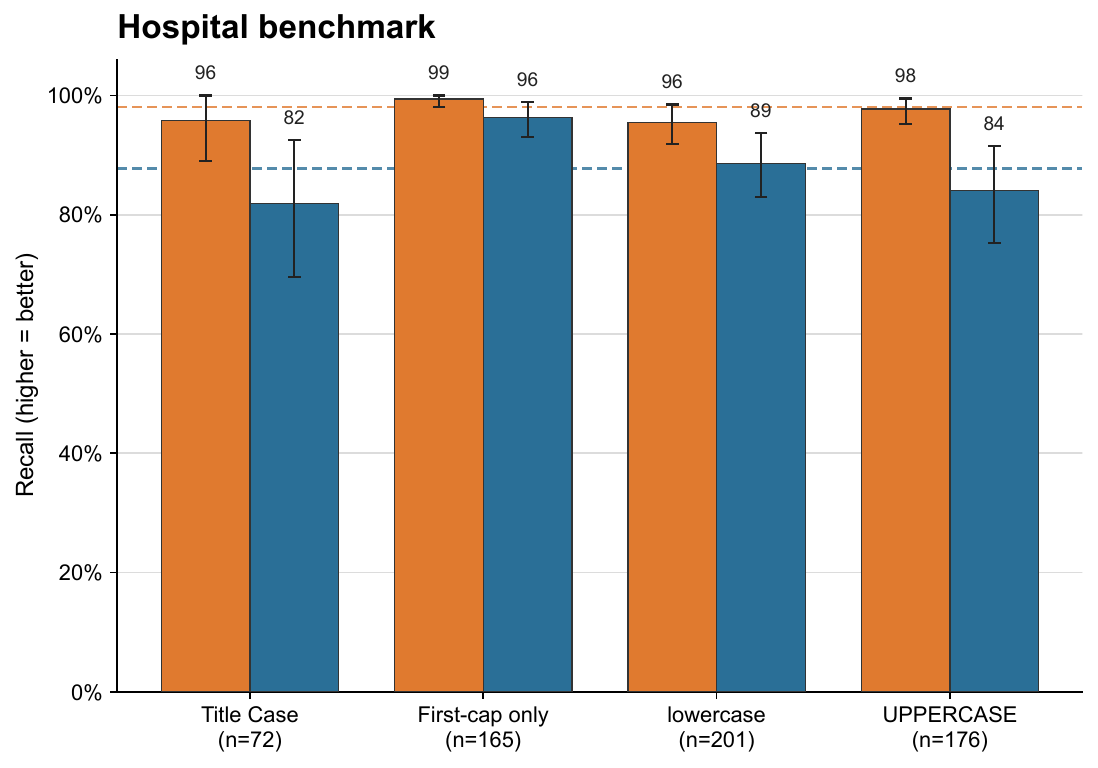}
	\end{minipage}\hfill
	\begin{minipage}[t]{0.49\linewidth}\textbf{b}\par
		\includegraphics[width=\linewidth]{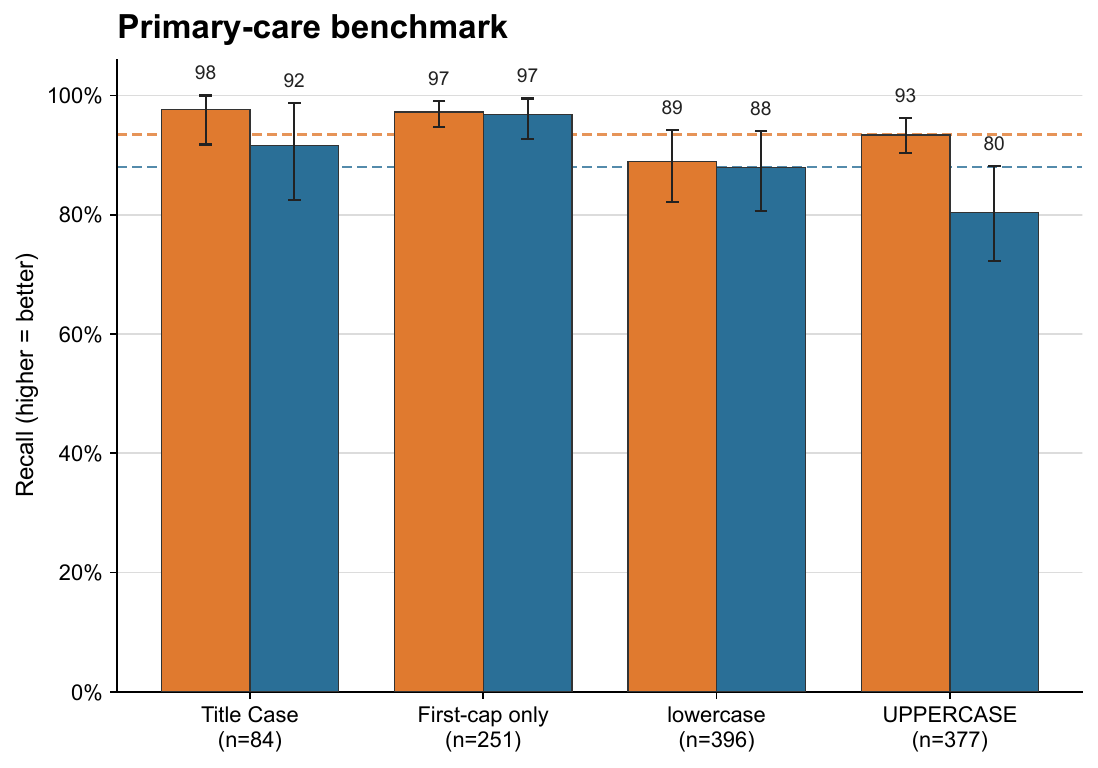}
	\end{minipage}
	\par\smallskip
	\includegraphics[width=0.78\linewidth]{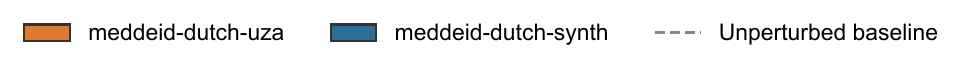}
	\caption*{\textbf{Fig. \SuppFigNameCapitalisation{} \textbar{} Name recall by capitalisation.}
		\textbf{a}, Hospital benchmark. \textbf{b}, Primary-care benchmark. Error bars show
		95\% note-level cluster-bootstrap confidence intervals. Higher recall indicates better
		performance. Descriptively, the largest reduction shown is for the synthetic-trained model with
		all-uppercase names in primary care; other capitalisation effects vary by model and benchmark.}
\end{figure}

\clearpage

\begin{center}
	\centering
	\begin{minipage}[t]{0.49\linewidth}\textbf{a}\par
		\includegraphics[width=\linewidth]{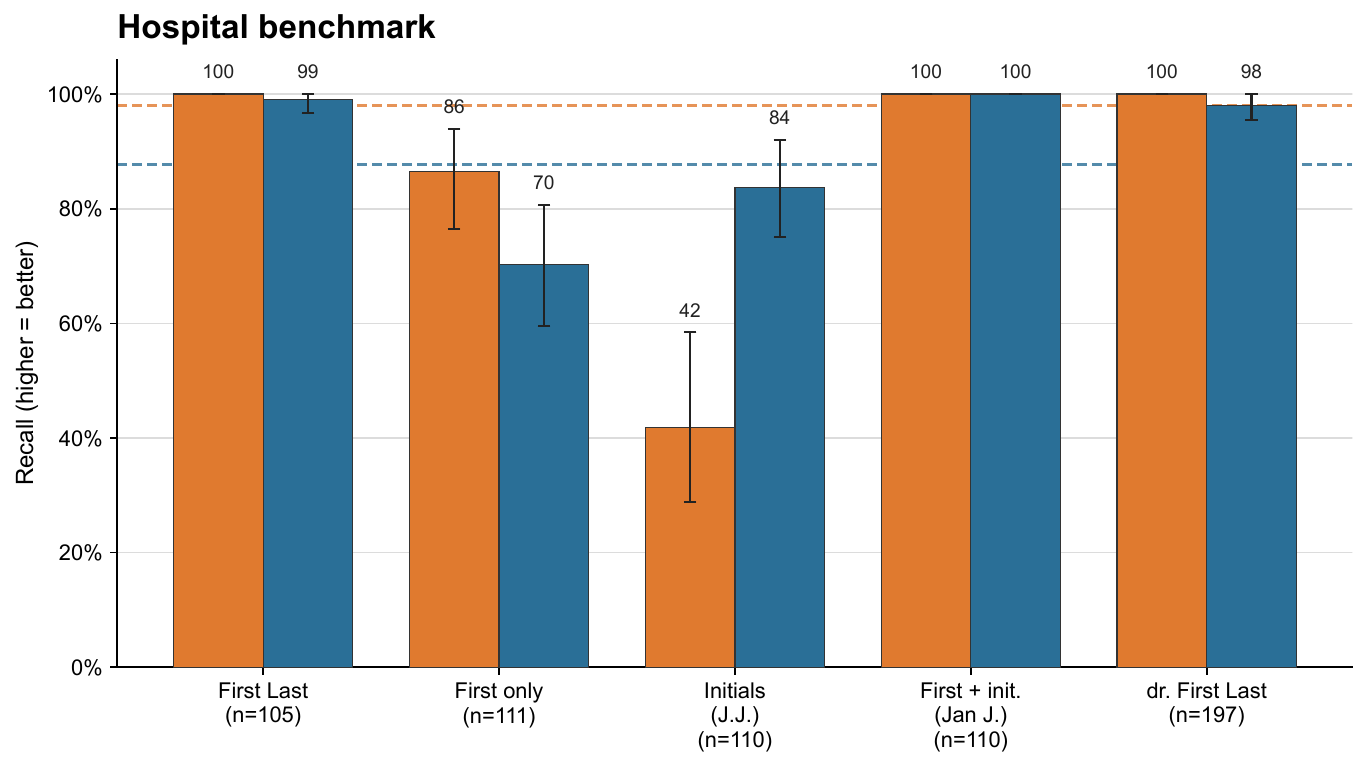}
	\end{minipage}\hfill
	\begin{minipage}[t]{0.49\linewidth}\textbf{b}\par
		\includegraphics[width=\linewidth]{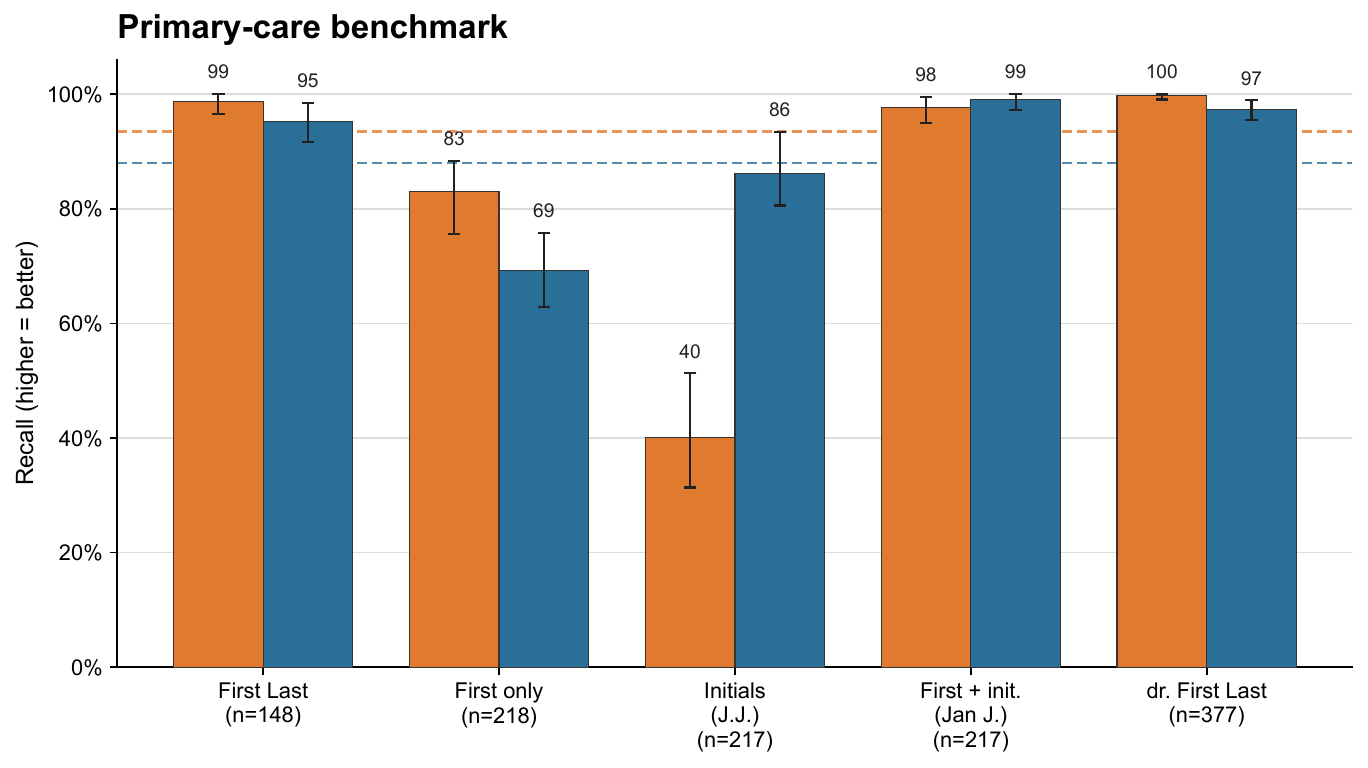}
	\end{minipage}
	\par\smallskip
	\includegraphics[width=0.78\linewidth]{FigS_shared_bar_legend.pdf}
	\suppfigcaption{\textbf{Fig. \SuppFigNameFormat{} \textbar{} Name recall by written format.}
		The models fail in different ways: initials cause the largest drop for the
		hospital-trained model, whereas first-name-only forms reduce recall more for the synthetic-trained
		model. \textbf{a}, Hospital benchmark. \textbf{b}, Primary-care benchmark. Error bars show
		95\% note-level cluster-bootstrap confidence intervals.}
\end{center}

\begin{center}
	\centering
	\makebox[0.49\textwidth][l]{\normalsize\textbf{a}}\hfill
	\makebox[0.49\textwidth][l]{\normalsize\textbf{b}}\par\vspace{-2pt}
	\begin{tabular}{@{}p{0.49\textwidth}@{\hfill}p{0.49\textwidth}@{}}
		\includegraphics[width=\linewidth]{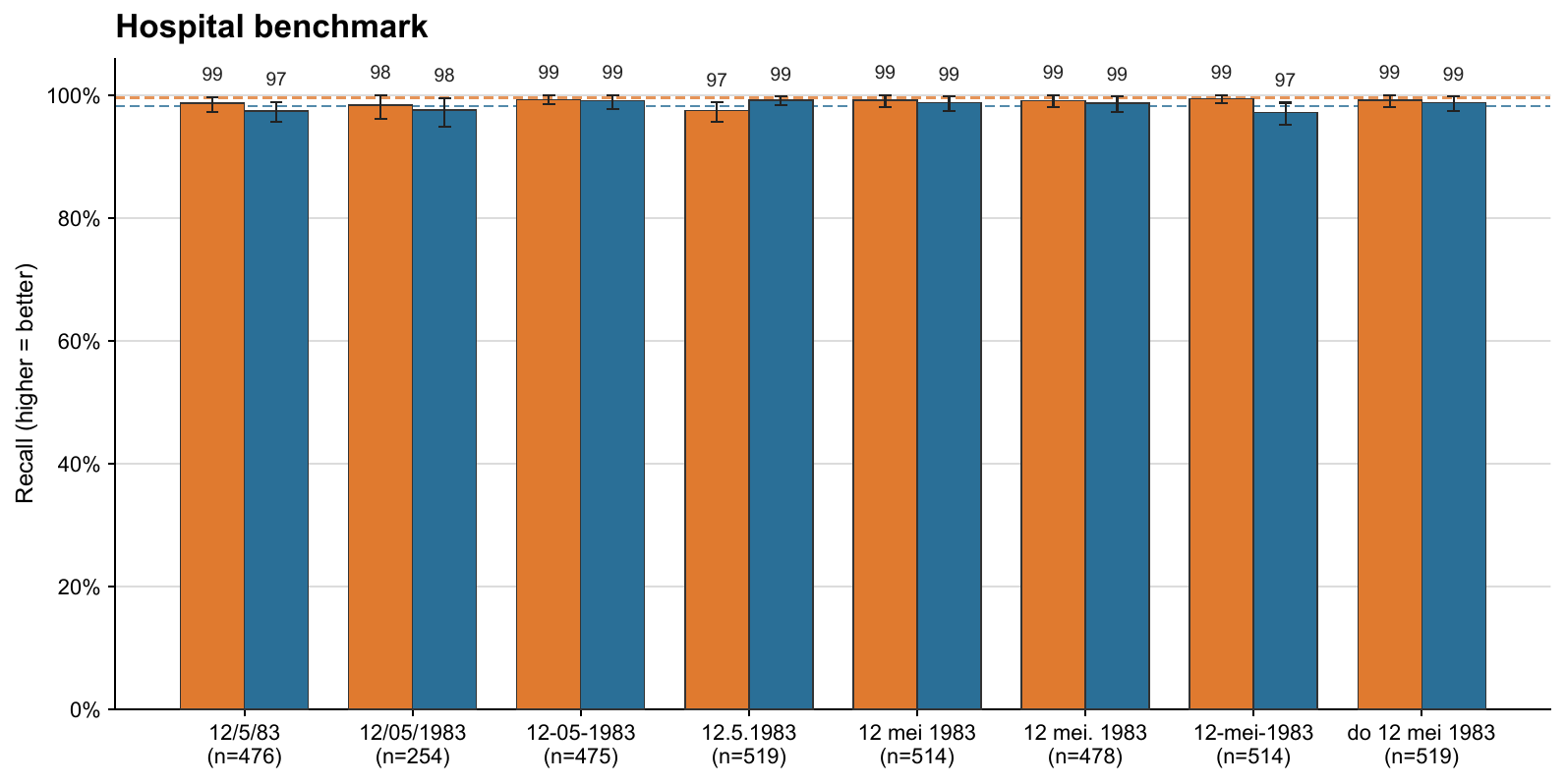} &
		\includegraphics[width=\linewidth]{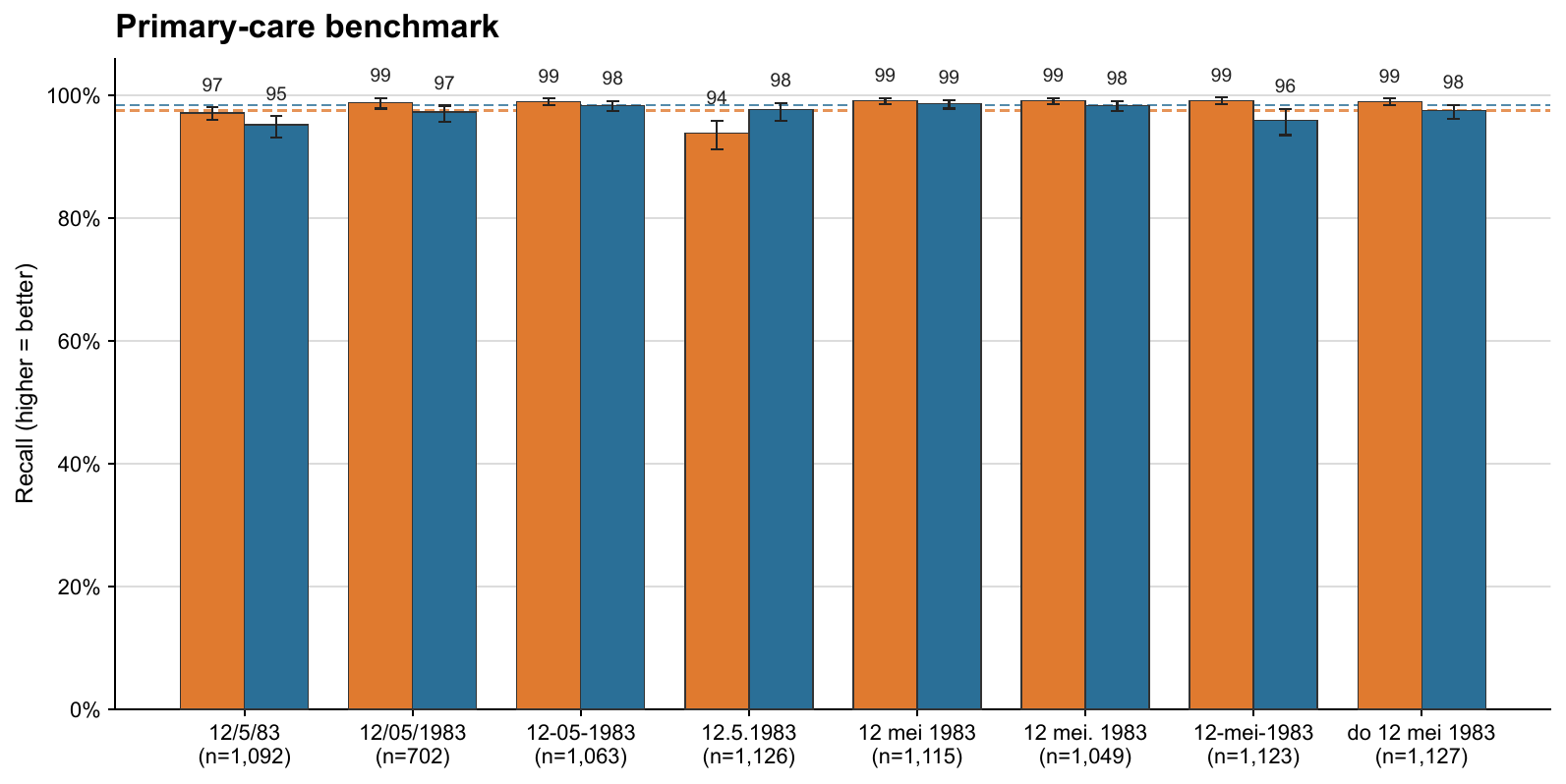}
	\end{tabular}
	\par\smallskip
	\includegraphics[width=0.78\linewidth]{FigS_shared_bar_legend.pdf}
	\suppfigcaption{\textbf{Fig. \SuppFigDateFormat{} \textbar{} Date recall by written format.}
		Both models remain close to their unperturbed baselines across most written formats, so formatting
		alone explains little of the larger date-stability differences. \textbf{a}, Hospital benchmark.
		\textbf{b}, Primary-care benchmark. Error bars show 95\% note-level cluster-bootstrap confidence intervals.}
\end{center}

\begin{figure}[!htbp]
	\centering
	\begin{minipage}[t]{0.62\linewidth}\textbf{a}\par
		\includegraphics[width=\linewidth]{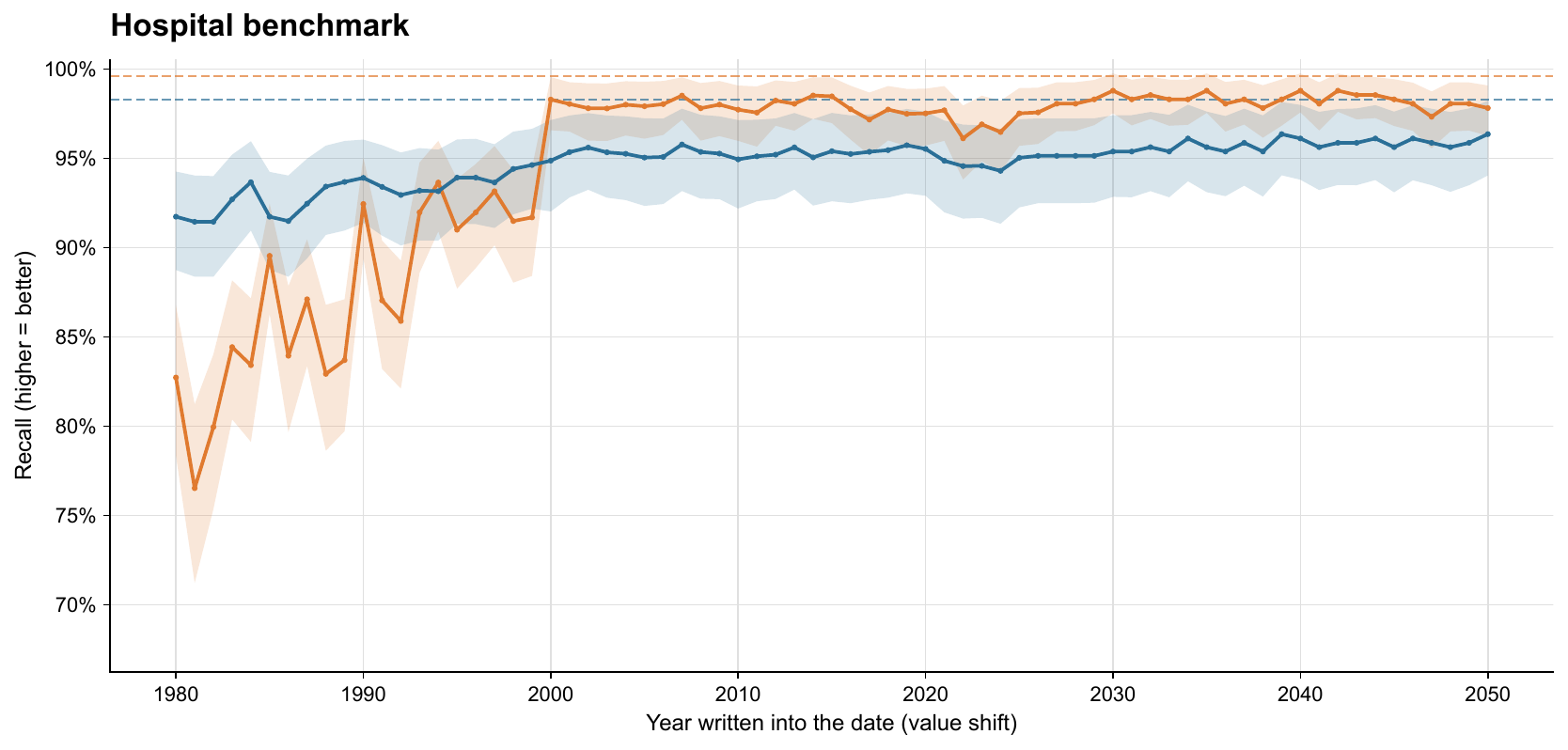}
	\end{minipage}
	\par\smallskip
	\begin{minipage}[t]{0.62\linewidth}\textbf{b}\par
		\includegraphics[width=\linewidth]{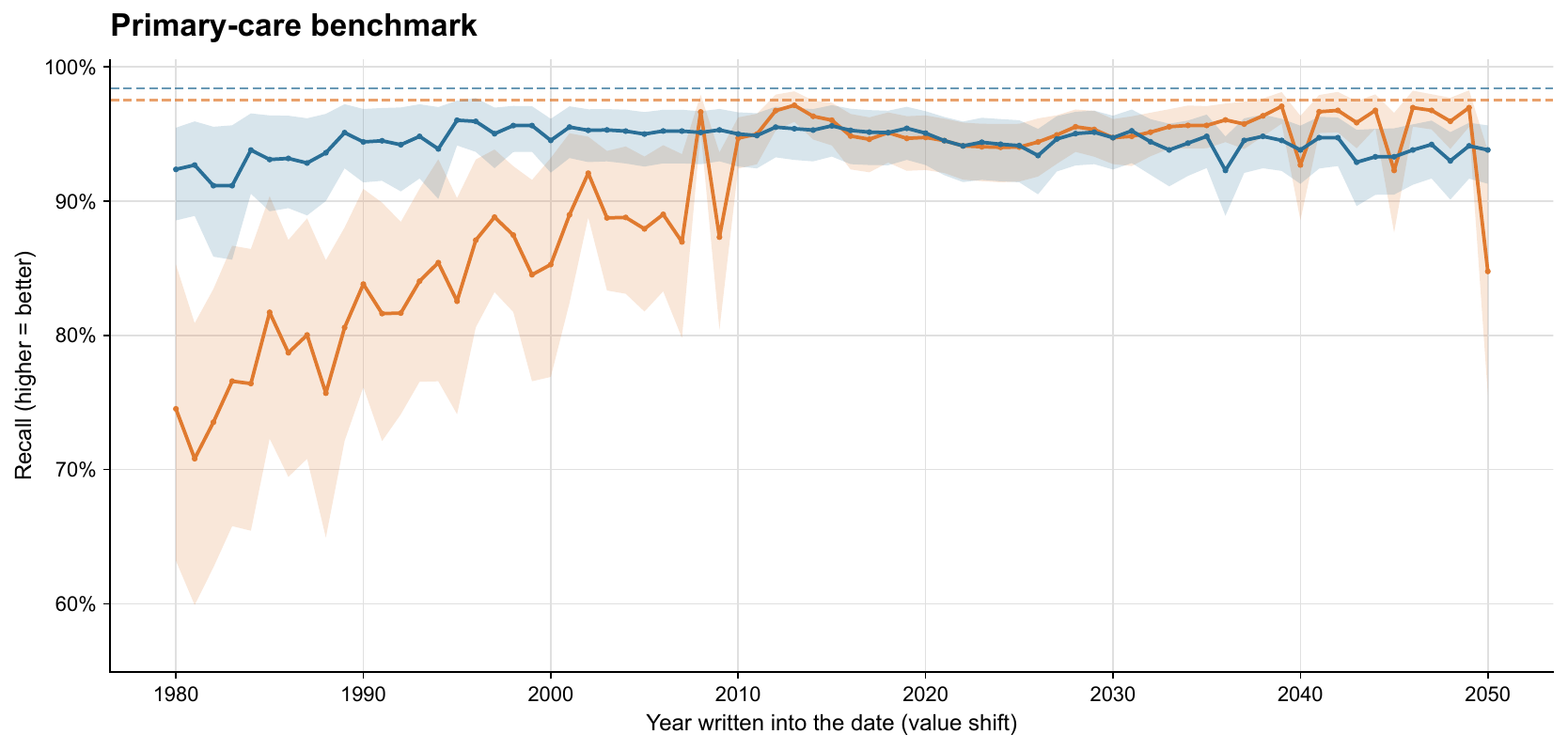}
	\end{minipage}
	\par\smallskip
	\includegraphics[width=0.78\linewidth]{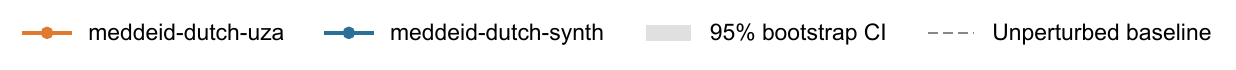}
	\caption*{\textbf{Fig. \SuppFigDateShift{} \textbar{} Date recall under year shifts.}
		Recall for the hospital-trained model falls as dates are shifted towards earlier years, while the
		synthetic-trained model remains comparatively stable. This identifies date value, rather than merely
		formatting, as a source of sensitivity. \textbf{a}, Hospital benchmark. \textbf{b}, Primary-care benchmark.
		Shaded bands show 95\% note-level cluster-bootstrap confidence intervals.}
\end{figure}

\begin{figure}[!htbp]
	\centering
	\begin{minipage}[t]{0.62\linewidth}\textbf{a}\par
		\includegraphics[width=\linewidth]{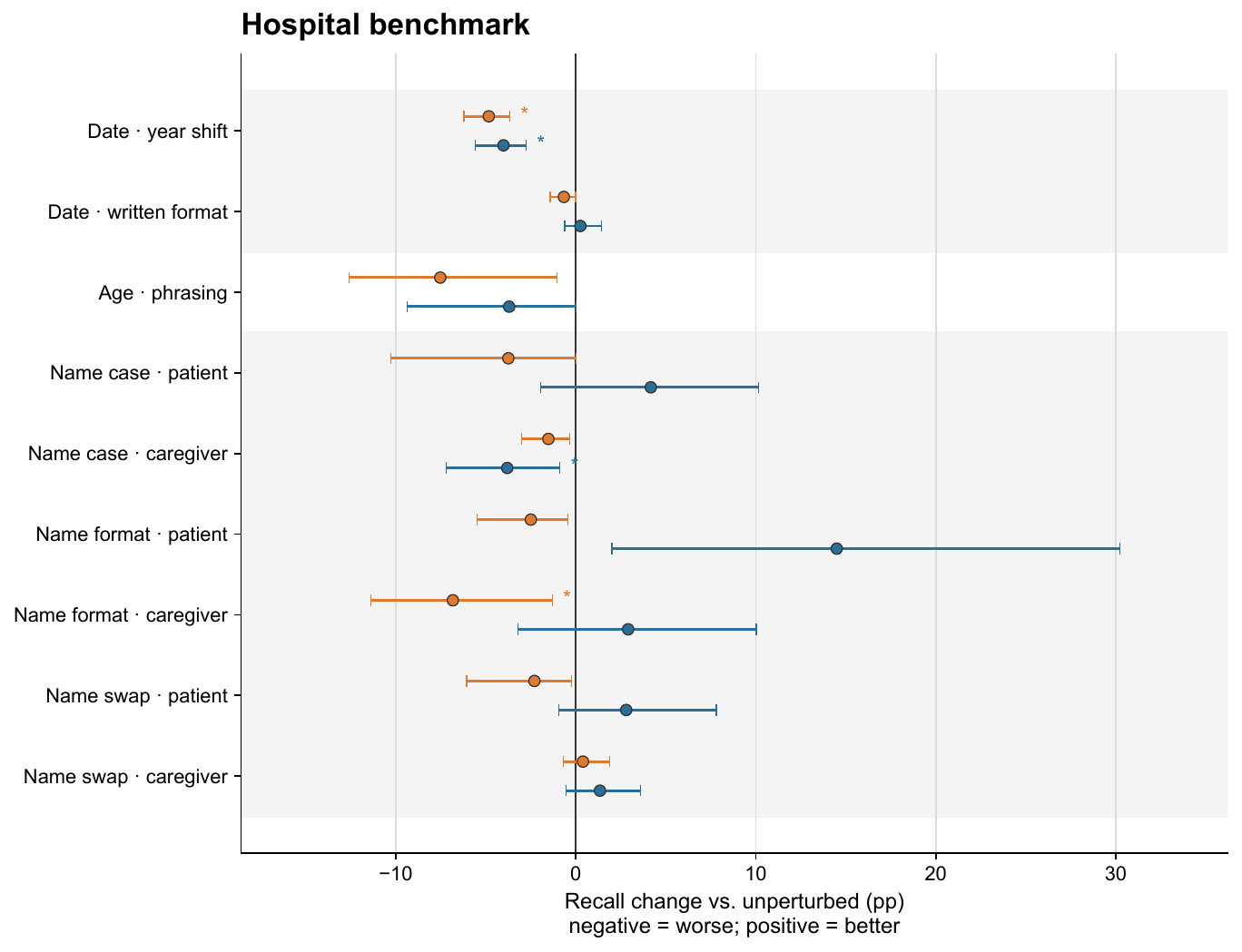}
	\end{minipage}
	\par\smallskip
	\begin{minipage}[t]{0.62\linewidth}\textbf{b}\par
		\includegraphics[width=\linewidth]{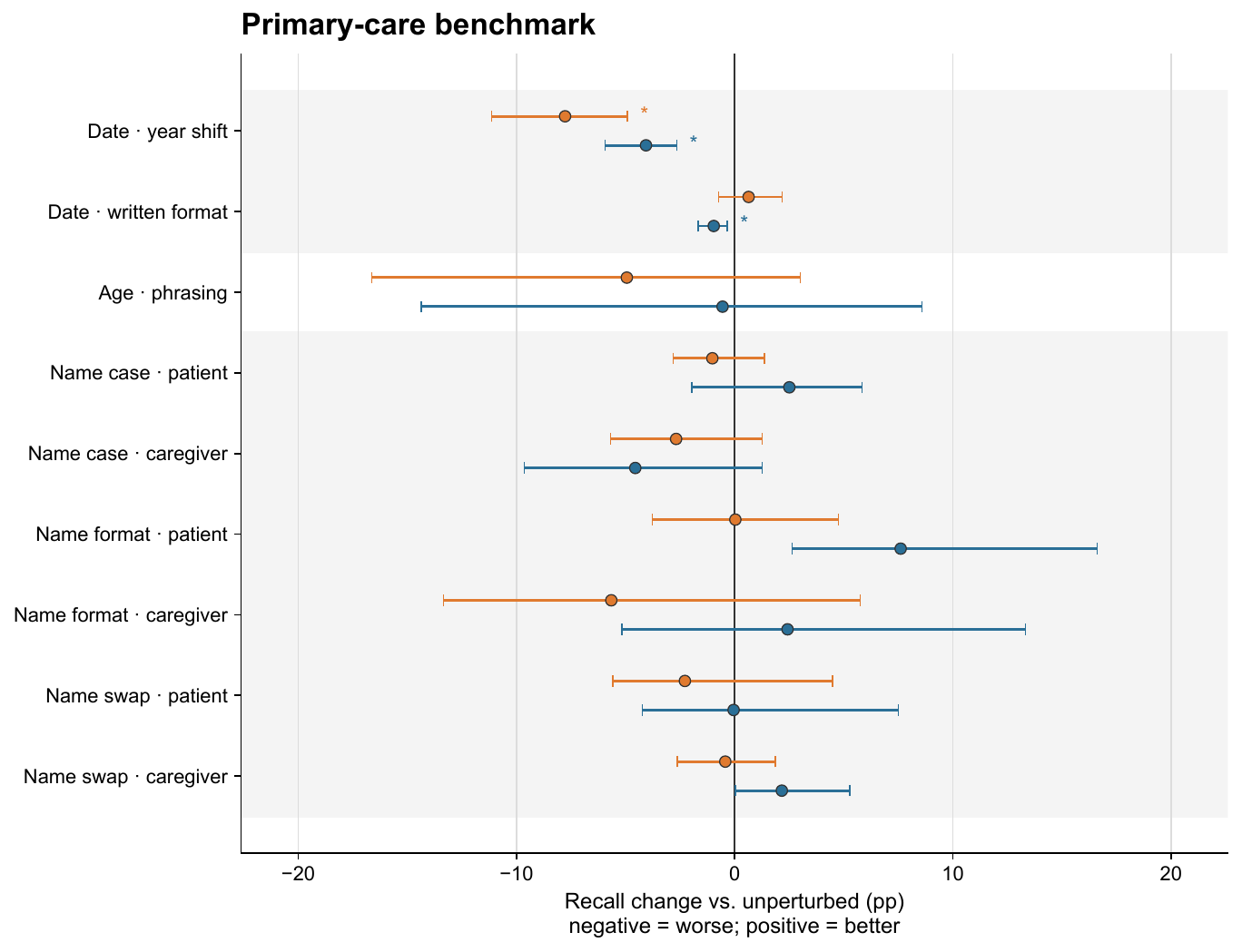}
	\end{minipage}
	\par\smallskip
	\includegraphics[width=0.78\linewidth]{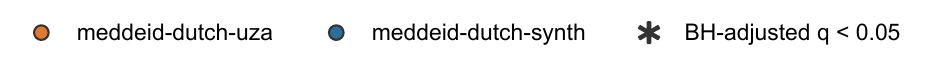}
	\suppfigcaption{\textbf{Fig. \SuppFigStabilityCells{} \textbar{} Cell-level recall change under controlled perturbation.}
		Year-shifted dates produce the clearest consistent degradation, particularly for the hospital-trained
		model; most name effects include little or no change. \textbf{a}, Hospital benchmark. \textbf{b},
		Primary-care benchmark. Points and horizontal lines show mean paired changes and 95\% note-level
		cluster-bootstrap confidence intervals. Negative values indicate worse performance. Asterisks denote
		Benjamini--Hochberg-adjusted $q<0.05$ in the 27-cell cross-scope family for each model.}
\end{figure}

\clearpage

\hypertarget{supp-section-non-pii-redaction}{%
	\subsection{S7. Non-PII redaction analysis}\label{supp-section-non-pii-redaction}}

\begin{center}
	\centering
	\includegraphics[width=\linewidth]{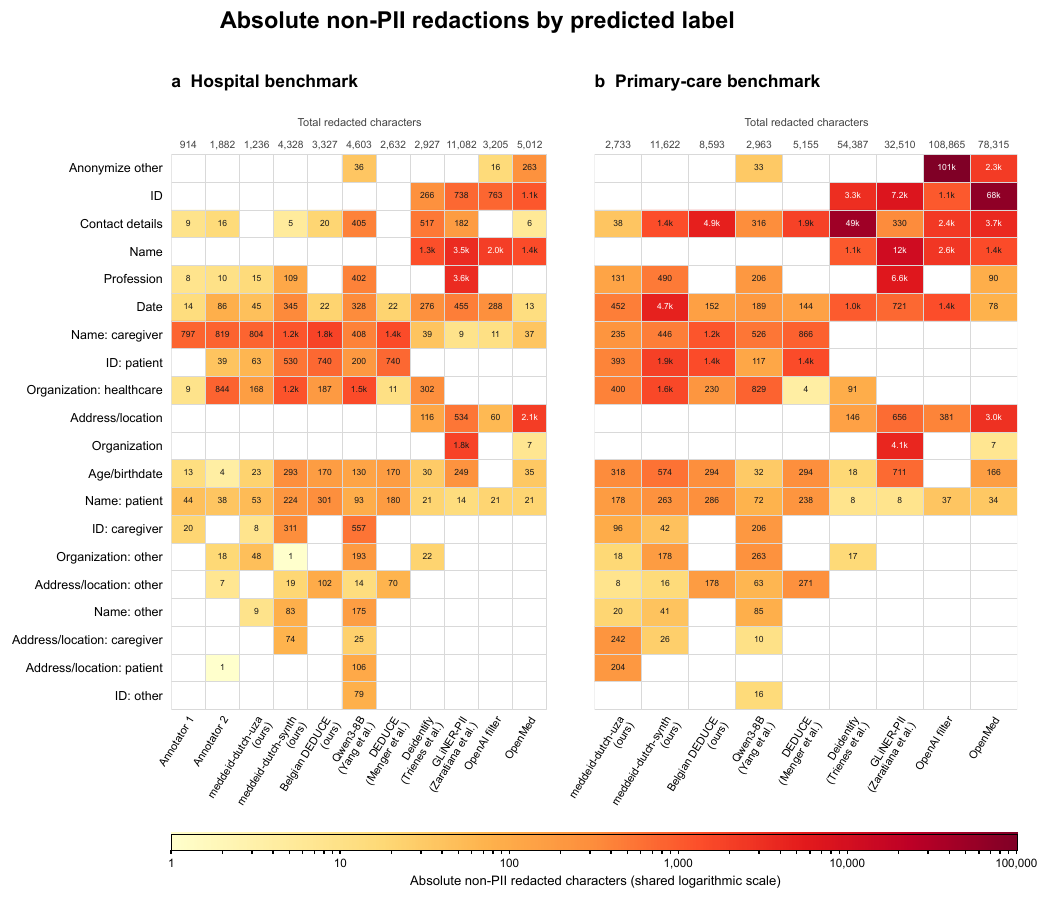}
	\suppfigcaption{\textbf{Fig. \SuppFigNonPIIRedaction{} \textbar{} Absolute non-PII redactions by predicted label in the hospital and primary-care benchmarks.}
		\textbf{a}, Hospital benchmark, including the two human annotators. \textbf{b}, Primary-care
		benchmark; separate human-annotator outputs were unavailable. Cell colour indicates the absolute
		number of non-PII characters redacted on a logarithmic scale shared by both panels; compact values
		are printed in non-zero cells. Empty cells represent zero. Values above the columns show the exact
		total number of non-PII characters redacted by each system or annotator. Non-PII redaction includes
		both false-positive spans and extensions of predicted PII spans beyond the annotated boundaries.}
\end{center}
On primary-care text, the OpenAI neural PII detector redacted 11.46\% of non-PII characters and the
OpenMed detector 8.25\%, compared with 0.29\% for \texttt{meddeid-dutch-uza} (Table
\SuppTabDutchPrimaryCare). Most of this excess came from false-positive spans concentrated in a few
predicted categories; boundary extensions made only a small contribution (Tables
\SuppTabDutchPointRange{} and Fig. \SuppFigNonPIIRedaction{}).

\clearpage
\begin{landscape}

	\hypertarget{supp-section-model-and-timing-specifications}{%
		\subsection{S8. Model and timing specifications}\label{supp-section-model-and-timing-specifications}}

	\hypertarget{supp-table-model-settings}{%
		\subsubsection{Table \SuppTabModelSettings{} \textbar{} MedDeID training and model-selection settings}\label{supp-table-model-settings}}

	\begingroup
	\fontsize{6.25}{6.65}\selectfont
	\setlength{\tabcolsep}{3pt}
	\renewcommand{\arraystretch}{0.94}
	\begin{tabular}{@{}>{\raggedright\arraybackslash}p{0.17\linewidth}>{\raggedright\arraybackslash}p{0.26\linewidth}>{\raggedright\arraybackslash}p{0.26\linewidth}>{\raggedright\arraybackslash}p{0.26\linewidth}@{}}
		\toprule
		Setting                                   & \texttt{meddeid-dutch-uza}                                                                                                   & \texttt{meddeid-dutch-synth} & \texttt{meddeid-english-synth}\tabularnewline
		\midrule
		\multicolumn{4}{@{}l}{\textbf{Training data, architecture and input}}\tabularnewline
		Training dataset
		                                          & 4,470 annotated real Dutch UZA hospital development notes
		                                          & 6,493 synthetic Dutch clinical notes; no real patient text
		                                          & 6,700 synthetic English clinical documents (3,350 \texttt{en-GB}, 3,350 \texttt{en-US}); no real patient text\tabularnewline
		Language / profiles
		                                          & Dutch
		                                          & Dutch
		                                          & English; \texttt{en-GB} and \texttt{en-US}\tabularnewline
		Encoder backbone
		                                          & \path{DTAI-KULeuven/robbert-2023-dutch-base} \citep{delobelle2024robbert2023}
		                                          & \path{DTAI-KULeuven/robbert-2023-dutch-base} \citep{delobelle2024robbert2023}
		                                          & \path{FacebookAI/roberta-base} \citep{liu2019roberta}\tabularnewline
		Prediction heads
		                                          & Token-level BIO head (3 classes) and entity-type head (14 classes)
		                                          & Same
		                                          & Same\tabularnewline
		Maximum sequence length                   & 512 tokens                                                                                                                   & Same                         & Same\tabularnewline
		Window overlap                            & 64 tokens                                                                                                                    & Same                         & Same\tabularnewline
		Random seed                               & 42                                                                                                                           & Same                         & Same\tabularnewline
		\addlinespace
		\multicolumn{4}{@{}l}{\textbf{Optimisation}}\tabularnewline
		Optimiser                                 & AdamW                                                                                                                        & Same                         & Same\tabularnewline
		Encoder learning rate                     & $2\times10^{-5}$                                                                                                             & Same                         & Same\tabularnewline
		Classification-head learning rate         & $1\times10^{-4}$                                                                                                             & Same                         & Same\tabularnewline
		Weight decay                              & 0.01 on all trainable parameters                                                                                             & Same                         & Same\tabularnewline
		Learning-rate schedule
		                                          & Linear decay; warm-up ratio 0.10; restarted after head warm-up
		                                          & Linear decay; warm-up ratio 0.10; restarted after head warm-up
		                                          & Linear decay; warm-up ratio 0.01\tabularnewline
		Head-only warm-up                         & 1 epoch                                                                                                                      & 1 epoch                      & None\tabularnewline
		Gradient checkpointing                    & Disabled                                                                                                                     & Same                         & Same\tabularnewline
		\addlinespace
		\multicolumn{4}{@{}l}{\textbf{Batching and runtime}}\tabularnewline
		Numerical precision                       & fp16                                                                                                                         & fp32                         & fp32\tabularnewline
		Training microbatch size                  & 16                                                                                                                           & 2                            & 8\tabularnewline
		Gradient-accumulation steps               & 1                                                                                                                            & 8                            & 2\tabularnewline
		Effective training batch size             & 16                                                                                                                           & Same                         & Same\tabularnewline
		Evaluation batch size                     & 16                                                                                                                           & 8                            & 16\tabularnewline
		Data-loader workers / prefetch factor     & 4 / 4                                                                                                                        & 0 / not applicable           & 0 / not applicable\tabularnewline
		Compute device                            & NVIDIA T4 (CUDA)                                                                                                             & Apple M4 Pro GPU (MPS)       & Apple M4 Pro GPU (MPS)\tabularnewline
		\addlinespace
		\multicolumn{4}{@{}l}{\textbf{Epoch selection and final refit}}\tabularnewline
		Selection training / validation documents & 4,023 / 447                                                                                                                  & 5,844 / 649                  & 6,030 / 670\tabularnewline
		Selection metric                          & Validation entity-level F1                                                                                                   & Same                         & Same\tabularnewline
		Maximum selection epochs                  & 50                                                                                                                           & 50                           & 30\tabularnewline
		Early stopping
		                                          & Patience 3; minimum improvement 0.001; active from epoch 3
		                                          & Patience 4; minimum improvement 0.001; active from epoch 4
		                                          & Patience 3; minimum improvement 0.001; active from epoch 3\tabularnewline
		Selection epochs completed                & 20                                                                                                                           & 21                           & 7\tabularnewline
		Selected epoch count                      & 17                                                                                                                           & 17                           & 4\tabularnewline
		Final refit corpus
		                                          & All 4,470 hospital development notes
		                                          & All 6,493 synthetic Dutch training notes
		                                          & All 6,700 synthetic English development documents\tabularnewline
		Final refit
		                                          & Reinitialised from the base encoder; 17 fixed epochs
		                                          & Reinitialised from the base encoder; 17 fixed epochs
		                                          & Reinitialised from the base encoder; 4 fixed epochs\tabularnewline
		Benchmark use during selection
		                                          & Benchmark withheld during selection and evaluated once after final refit
		                                          & Same
		                                          & Same\tabularnewline
		\bottomrule
	\end{tabular}
	\endgroup

	\vspace{2pt}
	{\fontsize{6.5}{7.2}\selectfont ``Same'' repeats the first model-column value. Zero data-loader
		workers made prefetching inapplicable. Selected epoch counts determined final-refit duration;
		benchmarks were withheld until final evaluation.\par}

\end{landscape}

\hypertarget{supp-table-timing}{%
	\subsubsection{Table \SuppTabTiming{} \textbar{} Time to de-identify 300 hospital notes}\label{supp-table-timing}}

\begingroup
\renewcommand{\arraystretch}{0.85}
\begin{longtable}[]{@{}llll@{}}
	\toprule
	System                        & Device & Time (s)  & Throughput (notes/s)\tabularnewline
	\midrule
	\endhead
	meddeid-dutch-uza (ours)      & CPU    & 320.31    & 0.937\tabularnewline
	meddeid-dutch-uza (ours)      & GPU    & 18.38     & 16.322\tabularnewline
	meddeid-dutch-synth (ours)    & CPU    & 318.74    & 0.941\tabularnewline
	meddeid-dutch-synth (ours)    & GPU    & 18.47     & 16.243\tabularnewline
	Belgian DEDUCE (ours)         & CPU    & 17.90     & 16.760\tabularnewline
	Qwen3-8B (Yang et al.)        & GPU    & 11,411.46 & 0.026\tabularnewline
	deidentify (Trienes et al.)   & GPU    & 100.66    & 2.980\tabularnewline
	deidentify (Trienes et al.)   & CPU    & 5,483.29  & 0.055\tabularnewline
	GLiNER-PII (Zaratiana et al.) & CPU    & 380.25    & 0.789\tabularnewline
	GLiNER-PII (Zaratiana et al.) & GPU    & 34.48     & 8.700\tabularnewline
	DEDUCE (Menger et al.)        & CPU    & 11.06     & 27.125\tabularnewline
	OpenAI privacy filter         & CPU    & 1,156.35  & 0.259\tabularnewline
	OpenAI privacy filter         & GPU    & 145.13    & 2.067\tabularnewline
	OpenMed multilingual filter   & CPU    & 999.85    & 0.300\tabularnewline
	OpenMed multilingual filter   & GPU    & 72.31     & 4.149\tabularnewline
	\bottomrule
\end{longtable}
\endgroup

{\footnotesize Warm end-to-end timings using the definition and computing environments described in
	Methods.\par}

\hypertarget{s9.-pseudonymisation-validation-and-age-dependent-transformation}{%
	\subsection{S9. Pseudonymisation validation}\label{s9-pseudonymisation}}

The gold-span evaluation supplied every Date and Age\_Birthdate span directly to the transformation
layer, independently of detector recall. The predicted-span evaluation instead used the
metadata-enabled \texttt{meddeid-dutch-synth} predictions. Both evaluations used a fixed creation
date (15 January 2025), a $+371$-day shift and birthdate-to-age replacement.

\subsubsection{Table \SuppTabPseudonymValidity{} \textbar{} Gold-span transformation and predicted-span end-to-end failures}\label{supp-table-pseudonym-validity}
\begingroup
\footnotesize
\begin{longtable}[]{@{}llrrrr@{}}
	\toprule
	Dataset      & Target        & Gold spans & \shortstack{Gold-transform\\failed, n (\%)} & \shortstack{Predicted end-to-end\\failed, n (\%)} & \shortstack{Gold spans with\\unredacted\\characters, n (\%)}\tabularnewline
	\midrule
	\endhead
	Synthetic    & Overall       & 1693       & 0 (0.00)                                    & 25 (1.48)                                         & 12 (0.71)\tabularnewline
	             & Date          & 1281       & 0 (0.00)                                    & 22 (1.72)                                         & 9 (0.70)\tabularnewline
	             & Age/birthdate & 412        & 0 (0.00)                                    & 3 (0.73)                                          & 3 (0.73)\tabularnewline
	\addlinespace
	UZA          & Overall       & 1798       & 14 (0.78)                                   & 61 (3.39)                                         & 39 (2.17)\tabularnewline
	             & Date          & 1463       & 13 (0.89)                                   & 33 (2.26)                                         & 26 (1.78)\tabularnewline
	             & Age/birthdate & 335        & 1 (0.30)                                    & 28 (8.36)                                         & 13 (3.88)\tabularnewline
	\addlinespace
	Primary-care & Overall       & 1219       & 36 (2.95)                                   & 73 (5.99)                                         & 29 (2.38)\tabularnewline
	             & Date          & 1047       & 23 (2.20)                                   & 45 (4.30)                                         & 23 (2.20)\tabularnewline
	             & Age/birthdate & 172        & 13 (7.56)                                   & 28 (16.28)                                        & 6 (3.49)\tabularnewline
	\bottomrule
\end{longtable}
\endgroup

{\footnotesize All percentages use gold Date and Age\_Birthdate spans as the denominator.
	Gold-transform failure isolates the deterministic transformation layer by supplying the gold span
	directly. Predicted end-to-end failure counts a gold span unless one predicted span covers it
	completely, has the correct label and produces a protocol-valid transformation. Gold spans with
	unredacted characters are identifiers for which at least one original character remained outside
	the model's predicted redactions. These cases are included among end-to-end failures. Other
	end-to-end failures were fully redacted but failed because the identifier was split across
	predictions, assigned the wrong label or transformed incorrectly. Gold-transform failures comprised
	14 UZA cases and 36 primary-care cases.\par}

\begin{center}
	\begin{minipage}{0.96\linewidth}
		\centering
		\includegraphics[width=\linewidth]{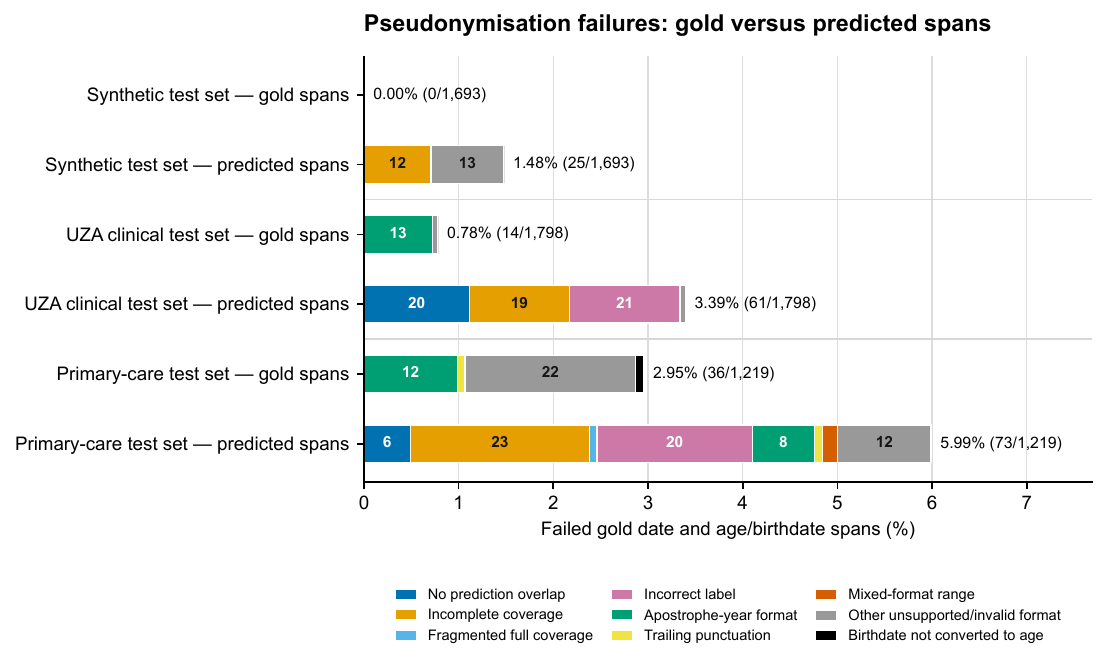}
		\suppfigcaption{\textbf{Fig. \SuppFigPseudonymFailureModes{} \textbar{} Pseudonymisation failures with gold versus model-predicted spans.}
			Paired rows use the same denominator of all gold Date and Age\_Birthdate spans. Gold-span rows
			bypass detection and isolate the deterministic transformation layer; predicted-span rows evaluate
			the full end-to-end pipeline. Stacked widths are failure rates, segment labels are counts and labels
			to the right give the total failure rate and count. A colour-blind-friendly palette encodes failure
			mode. The detailed format categories retain the
			transformation-layer failure analysis in both setups. No prediction overlap and incomplete coverage
			are the two modes that leave gold-span characters unredacted and occur only in the predicted-span
			setup. Fragmented full coverage and incorrect labels can fail end to end without leaving source
			characters unredacted.
			Gold-span failures comprised 13 apostrophe-year dates and one other invalid format in UZA; in
			primary care they comprised 12 apostrophe-year dates, one trailing-punctuation date, 22 other
			invalid formats and one birthdate that was not converted to age.}
	\end{minipage}
\end{center}

\subsubsection{Table \SuppTabAgeGranularity{} \textbar{} Age-dependent retained granularity}\label{supp-table-age-granularity}

\begin{longtable}[]{@{}ll@{}}
	\toprule
	Age                                     & Retained granularity\tabularnewline
	\midrule
	\endhead
	12 years or older                       & Whole years\tabularnewline
	2 to less than 12 years                 & Years and months\tabularnewline
	6 months to less than 2 years           & Months\tabularnewline
	More than 90 days to less than 6 months & Months and weeks\tabularnewline
	More than 28 through 90 days            & Weeks and days\tabularnewline
	28 days or younger                      & Days\tabularnewline
	\bottomrule
\end{longtable}

	{\footnotesize Age and birth-date values are reduced more aggressively as age increases, while finer
		units are retained when clinically important in early childhood. MedDeID Suite allows users to
		customise these groups and retained levels of detail for local clinical or governance requirements.\par}

	\clearpage
	\hypertarget{s10.-english-synthetic-portability-evaluation}{%
	\subsection{S10. English synthetic benchmark results}\label{s10-english-portability}}

Complete benchmark results are reported in Tables \SuppTabEnglishRange.

\subsubsection{Table \SuppTabTechnetium{} \textbar{} Technetium-I benchmark results}\label{supp-table-technetium}

\begin{longtable}[]{@{}>{\raggedright\arraybackslash}p{0.40\linewidth}>{\centering\arraybackslash}p{0.25\linewidth}>{\centering\arraybackslash}p{0.25\linewidth}@{}}
	\toprule
	Method                                         & Annotation-character recall, \%  & Non-PII redaction, \%\tabularnewline
	\midrule
	\endhead
	\textbf{\texttt{meddeid-english-synth} (ours)} & \textbf{99.730 (99.725--99.735)} & 1.606 (1.605--1.608)\tabularnewline
	GLiNER Multilingual PII                        & 97.773 (97.754--97.791)          & 3.616 (3.610--3.622)\tabularnewline
	OpenAI Privacy Filter                          & 96.328 (96.307--96.348)          & 0.870 (0.869--0.872)\tabularnewline
	OpenMed SuperClinical 434M                     & 94.828 (94.817--94.840)          & 1.723 (1.721--1.724)\tabularnewline
	OpenMed Multilingual Privacy Filter            & 92.032 (91.999--92.065)          & 1.086 (1.083--1.089)\tabularnewline
	OBI RoBERTa i2b2                               & 91.595 (91.582--91.608)          & 0.868 (0.865--0.870)\tabularnewline
	UCSF Philter                                   & 88.295 (88.285--88.304)          & \textbf{0.700 (0.699--0.701)}\tabularnewline
	\bottomrule
\end{longtable}

	{\footnotesize Percentages include 95\% document-clustered bootstrap confidence intervals.
		Technetium-I is template-generated with PHI in every document, so it is a reproducibility and scale
		test rather than clinical validation.\par}

\subsubsection{Table \SuppTabASQ{} \textbar{} ASQ-PHI benchmark results}\label{supp-table-asq}

\begingroup
\footnotesize
\begin{longtable}[]{@{}>{\raggedright\arraybackslash}p{0.31\linewidth}>{\centering\arraybackslash}p{0.21\linewidth}>{\centering\arraybackslash}p{0.20\linewidth}>{\centering\arraybackslash}p{0.20\linewidth}@{}}
	\toprule
	Method                                         & \shortstack{Annotation-character\\recall, \%} & \shortstack{Raw non-PII\\redaction, \%} & \shortstack{Non-PII redaction after\\excluding ages, \%}\tabularnewline
	\midrule
	\endhead
	\textbf{\texttt{meddeid-english-synth} (ours)} & \textbf{98.90 (98.59--99.19)}                 & 6.21 (5.99--6.43)                       & 0.89 (0.75--1.03)\tabularnewline
	GLiNER Multilingual PII                        & 96.32 (95.84--96.78)                          & 11.52 (11.11--11.93)                    & 4.29 (3.94--4.65)\tabularnewline
	OBI RoBERTa i2b2                               & 95.45 (95.08--95.80)                          & 2.01 (1.90--2.12)                       & \textbf{0.55 (0.45--0.66)}\tabularnewline
	OpenMed Multilingual Privacy Filter            & 75.36 (74.18--76.55)                          & 4.39 (4.10--4.69)                       & 3.20 (2.92--3.48)\tabularnewline
	UCSF Philter                                   & 70.89 (70.03--71.75)                          & 1.80 (1.59--2.01)                       & 1.80 (1.59--2.01)\tabularnewline
	OpenAI Privacy Filter                          & 63.48 (62.12--64.80)                          & \textbf{1.19 (1.02--1.36)}              & 1.18 (1.01--1.35)\tabularnewline
	OpenMed SuperClinical 434M                     & 61.67 (60.56--62.81)                          & 4.85 (4.66--5.03)                       & 3.43 (3.27--3.59)\tabularnewline
	\bottomrule
\end{longtable}
\endgroup

{\footnotesize Recall and both non-PII redaction columns are percentages with 95\% document-clustered
	bootstrap confidence intervals in parentheses. The last column reports results after the
	same unannotated age expressions were excluded for every system, regardless of its predicted labels.
	These expressions comprised 8,745 non-gold characters across 878 queries and captured explicit numeric ages
	(values 5--88, including two infant ages expressed in months) written as year-old or month-old
	phrases, \texttt{yo} or \texttt{y} shorthand, \texttt{age N}, compact age--sex expressions, or
	over/under thresholds; overlapping matches were merged. None of these characters overlapped a gold
	annotation, and qualitative descriptors such as \textit{elderly} and \textit{adolescents} remained
	included. For each system, redacted characters inside these age expressions were removed from the numerator while
	the original denominator of 119,651 non-PII characters was retained. Numerators after exclusion, in table
	order were 1,064, 5,137, 661, 3,826, 2,151, 1,411 and 4,101 characters.\par}

\subsubsection{Table \SuppTabEnglishHeldout{} \textbar{} Held-out English benchmark with metadata injection}\label{supp-table-english-heldout}

\begin{longtable}[]{@{}>{\raggedright\arraybackslash}p{0.28\linewidth}>{\centering\arraybackslash}p{0.23\linewidth}>{\centering\arraybackslash}p{0.23\linewidth}>{\centering\arraybackslash}p{0.18\linewidth}@{}}
	\toprule
	Method                                         & \shortstack{Recall, \%\\No metadata} & \shortstack{Recall, \%\\Patient/caregiver metadata} & \shortstack{Non-PII redaction, \%\\With metadata}\tabularnewline
	\midrule
	\endhead
	\textbf{\texttt{meddeid-english-synth} (ours)} & \textbf{99.96 (99.89--100.00)}       & \textbf{99.96 (99.89--100.00)}                      & \textbf{0.009 (0.002--0.017)}\tabularnewline
	GLiNER Multilingual PII                        & 90.27 (88.99--91.42)                 & 90.59 (89.33--91.75)                                & 1.831 (1.663--2.001)\tabularnewline
	OBI RoBERTa i2b2                               & 86.64 (85.59--87.70)                 & 86.73 (85.68--87.78)                                & 0.194 (0.154--0.237)\tabularnewline
	OpenMed SuperClinical 434M                     & 75.10 (73.32--76.81)                 & 75.12 (73.34--76.83)                                & 0.480 (0.400--0.563)\tabularnewline
	UCSF Philter                                   & 70.83 (69.45--72.20)                 & 70.94 (69.56--72.30)                                & 0.986 (0.899--1.076)\tabularnewline
	OpenMed Multilingual Privacy Filter            & 64.26 (62.04--66.45)                 & 69.45 (67.41--71.48)                                & 0.347 (0.292--0.405)\tabularnewline
	OpenAI Privacy Filter                          & 64.38 (61.62--67.17)                 & 68.45 (65.89--71.01)                                & 0.051 (0.029--0.076)\tabularnewline
	\bottomrule
\end{longtable}

	{\footnotesize Values are percentages with 95\% document-clustered bootstrap confidence intervals in
		parentheses. Metadata injection did not affect MedDeID recall or non-PII redaction. The largest recall
		gains were 5.19 percentage points for OpenMed Multilingual and 4.06 for OpenAI Privacy Filter. As an
		in-domain synthetic test, this 14-label benchmark does not constitute clinical validation.\par}

\bibliography{references}

% Preserve the guideline's original vector typography, highlights and page design.
% A visible divider makes the manuscript's Appendix A references unambiguous in print.
\clearpage
\phantomsection
\pdfbookmark[1]{Appendix A. Annotation guidelines}{appendix-a-annotation-guidelines}
\thispagestyle{plain}
\begin{center}
	\vspace*{0.23\textheight}
	{\Large\bfseries Appendix A. Annotation guidelines\par}
	\vspace{1.5em}
	{\large Complete English annotation guideline\par}
	\vspace{1em}
	The following pages reproduce the guideline in its original styled layout.
	\vspace{1.5em}

\end{center}


\section*{Supplementary material}

# DEID annotation guidelines (EN)

## ProductionLabels_v1

| ProductionLabels_v1 | |
|---|---|
| Name<br>    Patient<br>    Caregiver<br>    Other<br>Address_Location<br>    Patient<br>    Caregiver<br>    Other<br>Organization<br>    Healthcare<br>    Other | Date<br>Age_Birthdate<br>Profession<br>Contactdetails<br>ID<br>    Patient<br>    Caregiver<br>Anonymize_Other |

**Name**

This label covers all personal names and initials that refer to people. It is subdivided into Name:Patient, Name:Caregiver and Name:Other.

**Name:Patient**

The patient is the person the record belongs to. First names, surnames and initials must all be annotated as Name:Patient.

- Forms of address such as Mr/Mrs must not be annotated.
- Generic terms such as patient must not be annotated.

Examples:

1) *"Jan Jansen was admitted to the ICU"*
2) *"Veerle had chest pain"*
3) *"K. Aerts was seen at the cardiology clinic"*
4) *"Mr De Meyer received chemotherapy"*
5) *"The patient felt well"*
6) *"Pt is dizzy"*
7) *"The lady preferred not to be treated"*
8) *"Jan's mother is worried"*

**Name:Caregiver**

These are the names and initials of care providers, meaning any healthcare worker who comes into contact with the patient. This includes doctors, nurses, students and allied health professionals such as physiotherapists, speech therapists, dietitians, etc., as well as social services staff. External doctors and care providers also belong here.

- Professional titles such as dr./prof. or doctor are annotated together with the name when they appear immediately before or after it. If the title cannot be captured in a single annotation with the name because another word comes between them, do not annotate the title.
- Job descriptions referring to a specialty (e.g. pulmonologist) are not annotated.
- Roles such as head of department, responsible supervisor, … are not annotated.
- Names of care providers that form part of a publication reference are not annotated. This choice was made because publications are publicly available, which creates a route to re-identifying a care provider or a date.

Examples:

1) *"Validated by Dr. Geraert"*
2) *"Head of department Prof. Dr. Iepermans"*
3) *"Aaron inserted the IV line"*
4) *"Radiologist Jan Aerschot, Doctor"*
5) *"PI worked the night shift." (Initials refer to care provider Peter Ijkert)*
6) *"Contact has already been made with Jan from Bauren Care Home" (Bauren Care Home = Organization:Healthcare)*
7) *"Jozef is a doctor" (The professional title doctor is not directly adjacent to the name because other words come between them → do not annotate the title)*
8) *"Student JK saw the patient this morning" (JK = Name:Caregiver)*
9) *"(Friedmans et al. 2012)" (Even though Friedmans may be a doctor, do not annotate. The date is not annotated here either, because of the increased re-identification risk.)*

**Name:Other**

These are names and initials of family members, friends, other patients whose record this is not, and external people who are not care providers. The only exception to this rule is in reproductive medicine, where a patient together with their partner and/or child can both be considered patients within the same record.

Examples:

1) *"Sarah is the patient's mother"*
2) *"Joost had an argument with a fellow patient, Ronald" (Joost = Name:Patient)*
3) *"Lies is coming in for a NIPT test together with her husband Jacob" (Lies = Name:Patient, Jacob = Name:Patient because this is reproductive medicine)*

**Address_Location**

This label covers all descriptions of addresses, such as street names, postcodes, buildings, cities, municipalities and provinces. Post office district numbers and box (unit) numbers in apartment buildings also belong here. A further distinction is made between Address_Location:Patient, Address_Location:Caregiver and Address_Location:Other.

- Countries, continents and cross-border regions with more than 100,000 inhabitants must not be annotated, UNLESS they form part of a larger address. For the annotator's convenience, anything smaller than a country may be annotated.

**Address_Location:Patient**
These are addresses officially linked to the patient, for example home addresses, places of birth, etc.

Examples:
1) *"Registered address: Veltwijcklaan 77, 2180 Ekeren"*
2) *"The patient lives at Langestraat 128"*
3) *"She is of South African origin." (Country)*
4) *"She is originally from the Kempen region" (because it is smaller than a country)*
5) *"Patient's home address: Gravenlaan 23 box 2003, B-2900 Antwerp 4, Belgium." (Belgium is a country BUT part of a larger address; '4' is the post office district number)*

**Address_Location:Caregiver**
These are addresses of care providers, the care institution, etc.

- Departments and names of care institutions that contain a place name do not belong to this label. They fall under Organization:Healthcare instead.

Examples:
1) *"Out-of-hours GP post in Bornhem"*
2) *"The patient is taken to Middelheim General Hospital" (Middelheim General Hospital = Organization:Healthcare)*
3) *"Private practice: WILRIJK" (Private practice: WILRIJK = Organization:Healthcare)*

**Address_Location:Other**
These are addresses of external bodies, events (e.g. accidents, travel), friends and family.

Examples:
1) *"The accident happened at the junction of Leerstraat and Uifellaan"*
2) *"The patient went on holiday to Morocco." (Country)*
3) *"He is from the Middle East" (because this multi-country region has more than 100,000 inhabitants)*
4) *"His uncle has a holiday home in Nice" (because it is smaller than a country)*

**Organization**
A distinction is made between Organization:Healthcare and Organization:Other because it may later be decided to keep care institutions visible after all (with targeted pseudonymisation). It can be clinically relevant that the patient attended a university hospital or a burn centre, whereas other organisations such as companies, hobby clubs, etc. are of no clinical importance and only add re-identification risk.

**Organization:Healthcare**
All names of care institutions relevant to the patient, such as hospitals, care homes, out-of-hours GP posts, health insurance funds, social services, etc. Annotate the full name of the

organisation. The only exception is national health authorities, because they carry as much information as a country, which is likewise not annotated.

- Department, building and room codes such as D2 or A3 are also annotated.
- Hospital route numbers are also annotated.
- Software applications specific to a hospital are also annotated.
- File names (not URLs) that refer to a care institution also fall under this label.
- Names of medical device manufacturers are not annotated.

Examples:

1) *"The patient was admitted to Antwerp University Hospital (UZA)." (If an Address_Location sits directly next to the Organization, annotate them as a single span)*

2) *"Consultation at UZA – Edegem"*
3) *"The patient was admitted to UZA in Antwerp." (If words come between the Organization and the Address_Location, annotate them separately; Antwerp = Address_Location:Caregiver)*
4) *"The MDT meeting at MOCA takes place on Wednesday" (MOCA = the hospital's oncology centre, traceable to Antwerp University Hospital)*
5) *"There has been consultation with the burn centre at Middelheim." (the generic term burn centre is preserved; only the hospital name makes it identifiable)*
6) *"She was referred to the Pulderbos rehabilitation centre." (the generic term rehabilitation centre is preserved)*
7) *"Contact was made with the CM health insurance fund."*
8) *"An appointment was scheduled at Leuven University Hospital." (the generic term university hospital is preserved)*
9) *"He was referred to the Merksem out-of-hours GP post." (the generic term out-of-hours GP post is preserved)*
10) *"Report forwarded to the national Ministry of Health." (not annotated, because it is a national authority)*
11) *"Reimbursement by the national health insurance institute" (not annotated, because it is a national authority)*
12) *"Vaccination status is available on the national patient portal" (not annotated, because it is a national system)*
13) *"He received support from the Ekeren municipal social welfare service." (social services)*
14) *"The patient is on ward D2 after coming from the ICU." (ward codes are annotated; the word 'ward' itself is not)*
15) *"The ward is located on route 11" (hospital route)*
16) *"Surgery in OR 4" (building code)*
17) *"The patient is in UZA I4:BOX5" (= Organization:Healthcare, because it is a room code)*
18) *"Performed by MOLIS:LABO" (software application used in particular hospitals)*
19) *"Originally from C2M CHS" (software application used in particular hospitals)*
20) *"Internal guideline Pneumonia-UZA-guideline.pdf" (this refers to the care institution)*

21) *"Internal guideline pneumonia-guideline.pdf" (No reference to a care institution OR a patient)*
22) *"A Medtronic Inceptiv™ spinal cord neurostimulation system was implanted" (do not annotate Medtronic, because it is a medical device manufacturer)*

**Organization:Other**
All organisations not relevant to the patient's treatment and follow-up.

Examples:
1) *"He works at ING Bank."*

2) *"The patient is a member of Riviera Korfball Club."*
3) *"His daughter studies at Ghent University."*
4) *"The patient is a member of the pensioners' association."*
5) *"She was in contact with her employer, BASF Antwerp."*
6) *"He volunteers for Red Cross Flanders." (→ no clinical relevance → Other)*
7) *"He sustained a skull fracture during the fire at De Zwaan pub in 2019." (De Zwaan pub = Organization:Other, 2019 = Date; 'fire' is generic enough)*

**Date**
This covers dates in both numeric and fully written-out form: anything representing a day, month and/or year.

- Public holidays such as Christmas also count as a date.
- Seasons and clock times are not annotated.
- Weekdays (e.g. Monday) are only annotated when they appear immediately before or after a date.
- Additional information such as 'early'/'late' [month] is also annotated (e.g. early January).
- Dates are always annotated separately, UNLESS one date carries information that is essential to the other (e.g. 23/09 – 2/10/2023 → annotate together).
- By way of exception, dates are not annotated when they refer to publications, copyright statements, clinical scores or external systems, because they stand apart from the patient and the organisation AND pose a re-identification risk, since this information is publicly available.

Examples:
1) "The operation took place on Monday 12/03/2021." (weekday → annotate, because it sits next to a date)

2) "Consultation scheduled for 15 August 2020."
3) "He was seen on 1 January."
4) "The procedure took place at Christmas."
5) "Rehabilitation started in April 2019."
6) "Discharge took place on New Year's Eve."
7) "He came in for a check-up on Monday." (weekday → do not annotate, because it stands alone)
8) "She felt worse in the winter." (season → do not annotate)

9) "Pt came to the emergency department on 1/03/22 at 11:55" (time → do not annotate)
10) "He presented in early June." ('early' adds information about the time)
11) "Admitted between 23/09 – 2/10" (no shared information → separate)
12) "Admitted between 23/09 – 2/10/2023" (shared information in the final date → together!)
13) "Admitted from 23/09 to 2/10/2023" (shared information in the final date → together!)
14) "Allergy testing during the period April–June" (a period spanning several months)

**Age_Birthdate**

This label combines date of birth (in the same formats as Date) and age.

- For ages, the unit of time (year, months, -year-old) must be annotated along with the number, but additional words such as '... old' must not.
- Gestational age is not treated as age.

Examples:

1) *"The patient is 23 years old."*

2) *"A 65-year-old man was admitted."*
3) *"The baby is 6 months old."*
4) *"She was born on 12/05/1998."*
5) *"Date of birth: 3 October 1975."*
6) *"This concerns an 84-year-old lady."*
7) *"A boy aged 14 years came to the clinic."*
8) *"The patient has lived in Belgium for 12 years." (a duration, but not an age)*
9) *"Term baby (39w) with Apgar 8-9-9" (gestational age ≠ age)*

**Profession**

The occupation or job title must only be annotated for patients or their relatives, not for the treating care providers themselves.

- Education, voluntary roles, religions and hobbies also fall under this label, including a mention of a school year. Occupations and job titles of relatives must also be annotated.
- Annotate the full job title, e.g. retired firefighter.
- When an organisation is clearly named, that part falls under Organization.

Examples:

1) *"The patient works as a plumber."*

2) *"His father is a teacher."*
3) *"She is a hairdresser by profession."*
4) *"The patient is a retired postman." (the full title → 'retired' is part of it)*
5) *"He is in 2nd grade at Regenboog Primary School" (2nd grade = Profession, Regenboog Primary School = Organization:Other)*

6) *"She does break-dancing and art school" (break-dancing = Profession, art school = Profession)*
7) *"The patient is an engineer at BASF." (engineer = Profession, BASF = Organization:Other)*
8) *"His brother works as a chef at De Gouden Lepel restaurant." (chef = Profession, De Gouden Lepel restaurant = Organization:Other)*
9) *"The wife is a nurse." (because she is a relative of the patient, not a care provider in the record)*
10) *"The patient worked as a doctor in Congo for many years." (the patient's occupation, but not the country)*
11) *"Seen by dr. Peeters in the emergency department." (care provider in the record; dr. Peeters = Name:Caregiver)*
12) *"The report was drawn up by the radiologist." (care provider, so not annotated)*
13) *"Discussed with the head of department." (role of a care provider, so not annotated)*
14) *"The patient is the only imam of the mosque in Merksem." (imam of the mosque = Profession, Merksem = Address_Location:Patient)*
15) *"She is currently studying communication sciences at AP University College" (communication sciences = Profession, AP University College = Organization:Other)*
16) *"The patient is self-employed" (self-employed = Profession)*
17) *"He is chair of the local pigeon racing club." (= Profession, because no organisation is clearly named)*
18) *"The patient is the son of the mayor of Ekeren." (the mayor = Profession, Ekeren = Address_Location:Other)*
19) *"Religion: Protestantism" (Protestantism = Profession)*

**Contactdetails**

These are contact details used for communication, excluding postal addresses; postal addresses fall under Address_Location. This includes email, telephone, fax, pagers, URLs, …

- URLs containing patient-specific IDs fall under ID:Patient

Examples:

1) *"Telephone number: 03/123.45.67."*

2) *"Email: jan.jansen@hotmail.com."*
3) *"Fax: 03/765.43.21."*
4) *"Reachable on mobile: +32 477 12 34 56."*
5) *"Contact us via www.ziekenhuis-antwerpen.be."*
6) *"DECT: #4521." (internal telephone number)*
7) *"Practice website: http://www.huisarts-peeters.be."*
8) *"The patient lives at Langestraat 128." (Langestraat 128 = Address_Location:Patient)*
9) *"Further information can be found at www.uza.be/cfs-guideline" (a URL is an indirect contact detail)*

**ID**

This covers all codes and identifiers that can be traced back to one unique person or to a group of people (e.g. a protocol). A distinction is made between ID:Patient and ID:Caregiver. There is currently no reason to define ID:Other as well; as long as that does not change, annotate such cases as Anonymize_Other (see below).

Files without a reference to a care institution or a patient must not be annotated. A reference to a care institution falls under Anonymize_Other.

**ID:Patient**

All IDs specific to patients, for example patient numbers, national registry numbers, etc.

- Codes for genes or biomarkers do not fall under this label.
- Codes and links for imaging and pathology do fall under this label.
- Codes for surgical materials, diagnoses and systems do not fall under this label if they cannot be traced back to a patient or an organisation.
- All study protocols are also treated as ID:Patient.
- Files that refer to patients also fall under this label.

Examples:

1) *"Patient number: 12345678."*

2) *"National registry number: 85.07.15-123.45."*
3) *"Record number: UZ-2021-998877."*
4) *"The patient's internal ID is PT-00987."*
5) *"National social security number: 75021512345."*
6) *"Medical imaging can be found at: https://beelden.zna.be/23h3H0" (include the full URL, because it identifies the hospital)*
7) *"OR no. H3038L93"*
8) *"Hospitalisation ID: 308354043"*
9) *"Final report 304332-UZA-MDT.pdf" (this refers to the patient)*
10) *"Lab results can be identified by MI 3023 3920"*
11) *"CFDNA 32038" (CFDNA still carries clinical information = cell-free DNA)*
12) *"The patient is taking part in the BHEALTH study under reference K20214" (BHEALTH = ID:Patient, K20214 = ID:Patient)*
13) *"The ICD D633-22042 is implanted" (an ICD device code is a serial number, a unique number traceable to the patient)*

**ID:Caregiver**

All IDs specific to a care provider, for example a national provider registration number. This can be extended to IDs specific to a care institution.

Examples:

1) *"National provider registration number: 12345678901."*

2) *"Healthcare professional registration number: 19001234501."*
3) *"Doctor ID: DR-998877."*
4) *"The nursing registration number is VRN-223344."*

5) *“The ISO 12452 standard is met, with national accreditation under MED-245” (the ISO number is not annotated, because it cannot be traced back to a care institution)*
6) *“The material has LOT number 23532”*

**Anonymize_Other**

This can be seen as the catch-all bin for personally identifiable information about patients or care providers that does not immediately fit another label. Its main purpose is to give annotators the opportunity to flag gaps in the current ProductionLabels. Annotators are always asked to explain in a free-text field why that particular entity was annotated. Always consider whether the piece of information disproportionately increases the chance of re-identification.

Examples:

1) *“She was admitted after the bus accident in Sierre.” (Sierre = Address_Location:Patient; ‘bus accident’ is generic enough that removing ‘Sierre’ keeps the re-identification risk from increasing disproportionately)*



%% file: supplementary_identifiers.tex
\newcommand{\SuppTabDatasetDepartments}{S1}
\newcommand{\SuppTabAnnotationLabels}{S2}
\newcommand{\SuppTabSubannotations}{S3}
\newcommand{\SuppTabDutchHospital}{S4a}

\newcommand{\SuppTabDutchUncertainty}{S4d}
\newcommand{\SuppTabLabelRecall}{S5}
\newcommand{\SuppTabLabelFidelity}{S6}
\newcommand{\SuppTabStabilitySamples}{S7a}
\newcommand{\SuppTabStabilityAggregate}{S7b}

\newcommand{\SuppTabStabilityCells}{S7d}
\newcommand{\SuppTabModelSettings}{S8}
\newcommand{\SuppTabTiming}{S9}
\newcommand{\SuppTabPseudonymValidity}{S10a}
\newcommand{\SuppTabAgeGranularity}{S10b}
\newcommand{\SuppTabTechnetium}{S11}
\newcommand{\SuppTabASQ}{S12}
\newcommand{\SuppTabEnglishHeldout}{S13}

\newcommand{\SuppTabDutchPointRange}{S4a--c}

\newcommand{\SuppTabStabilityDetailRange}{S7b--d}
\newcommand{\SuppTabModelTimingRange}{\SuppTabModelSettings--\SuppTabTiming}

\newcommand{\SuppTabEnglishExternalRange}{\SuppTabTechnetium--\SuppTabASQ}
\newcommand{\SuppTabEnglishRange}{\SuppTabTechnetium--\SuppTabEnglishHeldout}

\newcommand{\SuppFigDatasetSummary}{S1}
\newcommand{\SuppFigLabelRecall}{S2}
\newcommand{\SuppFigSubannotationRecall}{S3}
\newcommand{\SuppFigLabelConfusion}{S4}
\newcommand{\SuppFigNameCapitalisation}{S5}

\newcommand{\SuppFigStabilityCells}{S9}
\newcommand{\SuppFigNonPIIRedaction}{S10}
\newcommand{\SuppFigPseudonymFailureModes}{S11}

\newcommand{\SuppFigLabelRange}{\SuppFigLabelRecall--\SuppFigSubannotationRecall}
\newcommand{\SuppFigStabilityRange}{\SuppFigNameCapitalisation--\SuppFigStabilityCells}